\documentclass[twoside]{article}

\usepackage[log]{snapshot}
\usepackage[preprint]{aistats2022preprint}

\usepackage[utf8]{inputenc} %
\usepackage[T1]{fontenc}    %
\usepackage{hyperref}       %
\usepackage{url}            %
\usepackage{booktabs}       %
\usepackage{amsfonts}       %
\usepackage{nicefrac}       %
\usepackage{microtype}      %
\usepackage{floatrow}

\usepackage{lmodern}

\usepackage{times}

\usepackage{tikz} %
\usetikzlibrary{shapes,arrows} %
\usetikzlibrary{positioning}
\usetikzlibrary{bayesnet}
\usetikzlibrary{patterns}
\usetikzlibrary{backgrounds}

\usepackage{booktabs}
\usepackage{comment}

\usepackage{bm}
\usepackage{adjustbox}
\usepackage{multirow}
\usepackage{amsmath, amssymb}
\usepackage{amsbsy}
\usepackage{stackengine}
\usepackage{mathtools}
\usepackage{xcolor}

\usepackage{natbib} %
\usepackage[font=small]{caption}
\usepackage[labelformat=simple]{subcaption}

\usepackage{titlesec}
\titlespacing*{\paragraph}{0pt}{0.0ex plus .2ex}{0.6em}

\newcommand{\y}{\mathbf{y}}
\newcommand{\f}{\mathbf{f}}
\newcommand{\g}{\mathbf{g}}
\newcommand{\z}{\mathbf{z}}
\newcommand{\U}{\mathbf{u}}
\newcommand{\h}{\mathbf{h}}
\newcommand{\X}{\mathbf{X}}
\newcommand{\Y}{\mathbf{Y}}
\newcommand{\F}{\mathbf{F}}
\newcommand{\x}{\mathbf{x}}
\newcommand{\K}{\mathbf{K}}
\newcommand{\0}{\mathbf{0}}
\newcommand{\I}{\mathbf{I}}
\newcommand{\m}{\mathbf{m}}
\DeclareMathOperator{\S_}{\mathbf{S}}
\newcommand{\s}{\mathbf{s}}
\newcommand{\w}{\mathbf{w}}

\newcommand{\Xaux}{\Tilde{\mathbf{X}}}
\newcommand{\Zaux}{\Tilde{\mathbf{Z}}}

\newcommand{\Fstar}{\mathbf{f}^*}
\newcommand{\Xstar}{\mathbf{X}^*}
\newcommand{\Zstar}{\mathbf{Z}^*}

\newcommand{\EX}{\mathbb{E}}
\newcommand{\KL}{\mathsf{KL}}

\newcommand{\given}{\,\vert\,}

\DeclarePairedDelimiterX{\diverge}[2]{\big[}{\big]}{%
  #1\,\delimsize\|\,#2%
}

\newcommand*\dd{\mathop{}\!\mathrm{d}}

\DeclareMathOperator{\sinc}{sinc}

\DeclareMathAlphabet{\mathcal}{OMS}{cmsy}{m}{n}

\newcommand{\MAPAMTGP}{{M-AMTGP}}

\begin{document}

\twocolumn[

\aistatstitle{Aligned Multi-Task Gaussian Process}

\aistatsauthor{Olga~Mikheeva \textsuperscript{1}\And Ieva~Kazlauskaite\textsuperscript{2} \And Adam~Hartshorne\textsuperscript{3} \And Hedvig~Kjellström\textsuperscript{1} \And Carl~Henrik~Ek\textsuperscript{2} \And Neill~D.~F.~Campbell\textsuperscript{3}}

\aistatsaddress{\textsuperscript{1} KTH Royal Institute of Technology,\\ Sweden  \And \textsuperscript{2} University of Cambridge,\\United Kingdom \And \textsuperscript{3} University of Bath,\\United Kingdom } ]

\begin{abstract}

Multi-task learning requires accurate identification of the correlations between tasks.
In real-world time-series, tasks are rarely perfectly temporally aligned; traditional multi-task models do not account for this and subsequent errors in correlation estimation will result in poor predictive performance and uncertainty quantification.
We introduce a method that automatically accounts for temporal misalignment in a unified generative model that improves predictive performance. %
Our method uses Gaussian processes (GPs) to model the correlations both within and between the tasks. %
Building on the previous %
work by~\citet{kazlauskaite2019gaussian}%
, we include a separate monotonic warp of the input data to model temporal misalignment.
In contrast to previous work, we formulate a lower bound that accounts for uncertainty in both the estimates of the warping process and the underlying functions. 
Also, our new take on a monotonic stochastic process, with efficient path-wise sampling for the warp functions, allows us to perform full Bayesian inference in the model rather than MAP estimates. 
Missing data experiments, on synthetic and real time-series, demonstrate the advantages of accounting for misalignments (vs standard unaligned method) as well as modelling the uncertainty in the warping process (vs baseline MAP alignment approach).

%
%
%

%
%
%
%
%
%
%
%
%
%

%
    
%

\end{abstract}

\section{Introduction}

Multivariate datasets gathered across a range of tasks are increasingly prevalent. 
In contrast to the established regression regime, where we aim to learn correlations across a time series within a single data source, we now wish to consider the relationships between different sources of data. 
This is the canonical multi-task learning scenario where we seek to model both the correlations within individual datasets as well as the correlations between datasets. 
If we perform this well, we are able to provide high quality predictions, with appropriate uncertainty quantification, under missing data scenarios; we can use correlations between time-series to fill in the gaps in data for a specific instances.
Success necessitates an accurate decomposition of correlations between these two factors and is inherently ill-posed.

To make progress we must find a principled regularisation that trades-off between the two generating components. 
Current approaches suffer a limiting assumption that all sources of data have perfect temporal alignment. 
Importantly, our terminology refers to the fundamental alignment between the generative process not to the precision of a sampling rate.
For example, two sources of data can share a common ancestral generating process but subsequent activities will introduce delays and phase shifts that result in temporal misalignment irrespective of some measurement clock. 
Failure to account jointly for these effects must lead to incorrect estimates of task correlations; this is particularly noticeable when we seek to account for uncertainty in our predictions.

This problem is also called domain shift; the observed covariates are transformed from some idealised covariates via a  distinct (unknown) per-task transformation~\citep{Quionero-Candela:2009}. 
The idealised covariates are typically unobserved, making such varying shifts challenging to identify. 
The problem is further complicated by the i.i.d.~observation noise.\footnote{Knowledge transfer between tasks is only possible when observations are assumed to include observation noise or correspond to different inputs~\citep{Wackernagel:2003, bonilla2008multi, alvarez2011computationally}. }
Typical examples of knowledge transfer between time-series data in a regression setting include: observing data from multiple tasks and sharing the knowledge between tasks~\citep{bonilla2008multi}; observing multiple trials of the same experiment and inferring missing data in some of the trials using the information from the others~\citep{Alvarez:2012}; multi-fidelity learning using cheap measurements as a proxy for expensive ones~\citep{Liu:2018, wang2020multi}; and clinical bench-marking where the emphasis is on the interpretability of the parameters of the covariates~\citep{durichen2014multitask}.

This paper addresses the temporal misalignment problem in a multi-task setting for time-series data. 
We build our model on Gaussian processes (GPs) to encode the prior knowledge of the inter-task and intra-task structure of the observations.
We use a latent variable construction to infer the inter-task correlations. 
Similarly to the GP-LVA model of~\citet{kazlauskaite2019gaussian}, we introduce a separate warping function for each task to address dataset shift.
We remove systematically the mismatch between tasks and allow the multi-task model to transfer accurately information between them.

\begin{figure*}[t]
\centering
\begin{subfigure}[b]{0.32\textwidth}
     \centering
     \includegraphics[width=\textwidth, height=3cm]{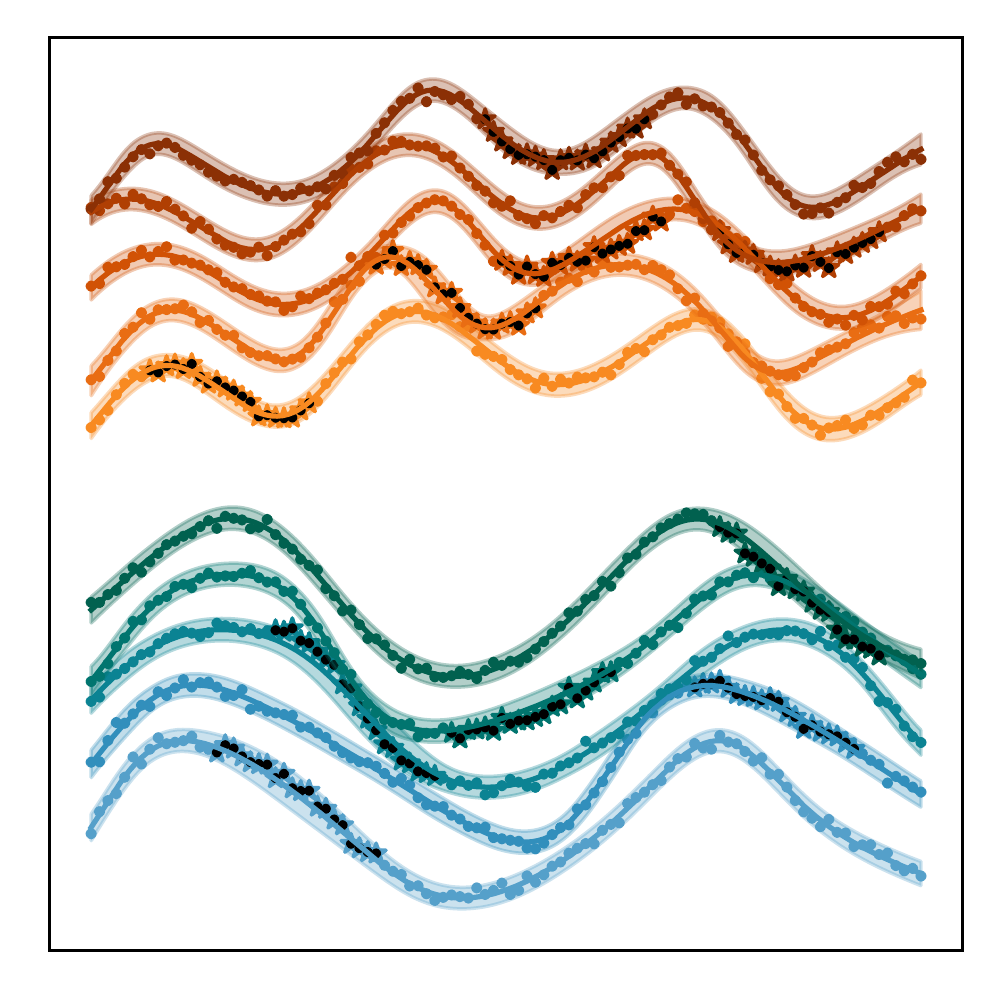}
     \caption{Observations and data fit}
     \label{fig:synthetic_latent_teaser}
 \end{subfigure}
 \hfill
 \begin{subfigure}[b]{0.32\textwidth}
     \centering
     \includegraphics[width=\textwidth, height=3cm]{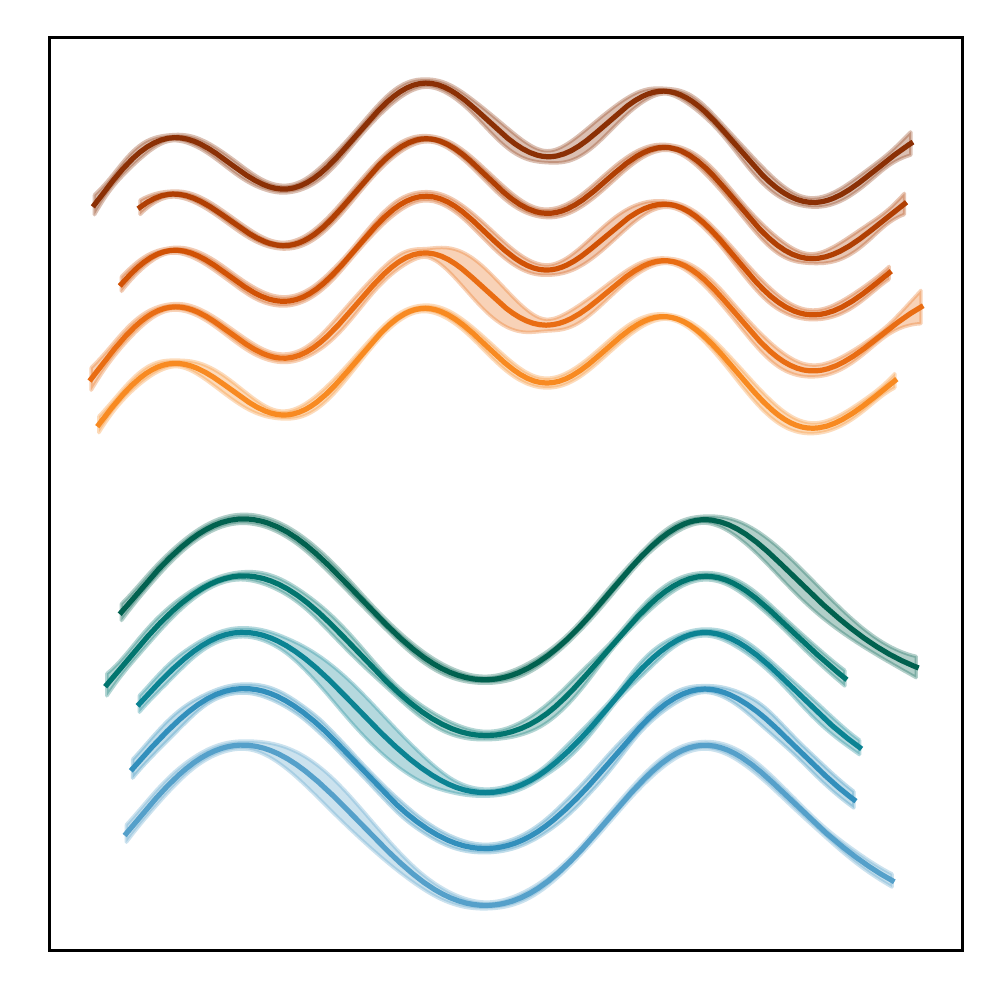}
     \caption{Aligned multi-task GP}
     \label{fig:synthetic_f_post_teaser}
 \end{subfigure}
 \hfill
 \begin{subfigure}[b]{0.32\textwidth}
     \centering
     \includegraphics[width=\textwidth, height=3cm]{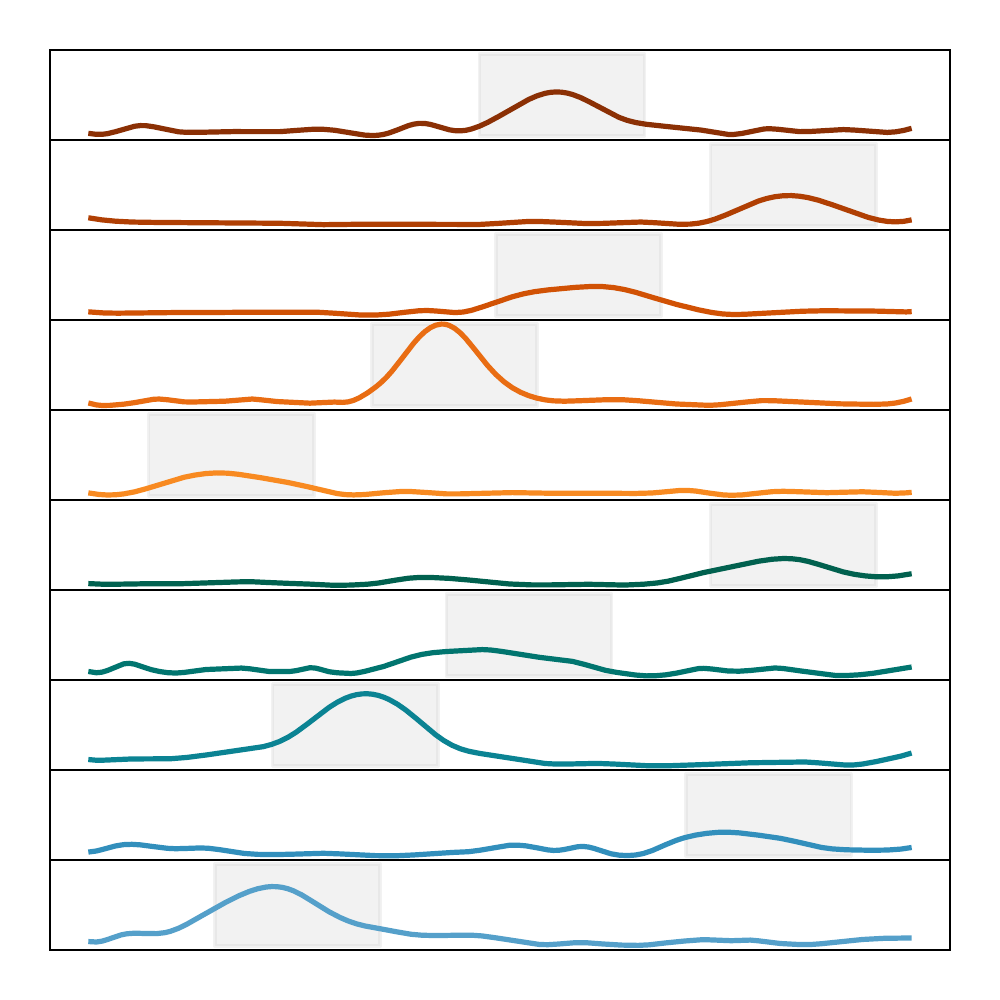}
     \caption{Uncertainty in the warps}
     \label{fig:warps_teaser}
 \end{subfigure}
\caption{Multi-task model for 10 time-series with missing data (shown in black). 
\subref{fig:synthetic_latent_teaser}~Observed data that comes from two different underlying sequences; for clarity, the sequences are coloured and plotted with a vertical offset. 
\subref{fig:synthetic_f_post_teaser}~Fitted aligned multi-task model. 
The model correctly uncovers and describes the two types of sequences despite missing data and dataset shift.
\subref{fig:warps_teaser}~The predictive standard deviation (uncertainty) of the estimated warps is shown to correlate well with the missing data regions (shown in grey) for each task.}
\label{fig:teaser}
\end{figure*}

Fig.~\ref{fig:teaser} illustrates a typical problem of dataset shift, \emph{i.e.}~the inputs to each sequence have been warped by unknown functions, with missing data. 
Our aligned multi-task Gaussian process (AMTGP) model shares the inter-task information to compensate for both the unknown transformation of the inputs and the missing data. The inferred uncertainty in the unknown warps, Fig.~\ref{fig:warps_teaser}, is observed to correlate with the regions of missing data and prevents over confidence.
In our experiments we test the model on synthetic and real time-series. %
We demonstrate that information sharing between tasks is improved by aligning the inputs and quantify the performance of the standard and the aligned MTGP models on missing data problems. We show that our uncertainty estimates are superior to the previous GP-LVA approach.

In summary, the contributions of this work are:
(1) a novel probabilistic approach for information transfer between tasks corrupted by temporal misalignment;
(2) an efficient inference scheme based on sparse stochastic variational inference;
(3) a reformulation of monotonic GP flow~\citep{ustyuzhaninov2020monotonic} with efficient training;
and (4) a model that is a generalisation of the GP-LVA model with a rigorous probabilistic formulation.

\section{Related Work}
\label{sec:related_work}

GPs are a standard Bayesian tool for time-series problems and have been used in Multi-Task (MT) settings such as dependent GPs~\citep{Boyle:2004}, Multi-Output GPs (MOGPs)~\citep{Bilionis:2013, Alvarez:2012} and Multi-Task GPs (MTGPs)~\citep{bonilla2008multi}. %
Historically, the topic has been termed the linear model of coregionalization~\citep{Journel:1978}, kernel methods for vector-valued functions~\citep{Evgeniou:2004} and matrix-variate Gaussian distributions~\citep{Dutilleul:1999}; for a review  please see, \emph{e.g.}~\citep{alvarez2011kernels}.

\paragraph{MTGPs}
\cite{bonilla2008multi} place a GP prior over each sequence (task) and include a free-form covariance matrix (constrained to be positive semi-definite). %
To reduce complexity, inter-task correlations can be modelled using probabilistic principal component analysis (pPCA); 
\cite{Stegle:2011} extend this to a GP-LVM for the covariance. %
\cite{alvarez2011computationally} use convolution processes to impose correlations 
that can be applied in cases where some of the sequences are blurred versions of the others.
More recently, \cite{boustati2019non} used compositional (deep) GPs for MT learning through non-linear mixing of latent processes (shared and individual). %
\cite{zhe2019scalable} propose a MOGP model with latent GPs as covariates and focus on inference efficiency exploiting the grid placement of the observations and the (deep) Kronecker factorisation. 
\cite{hamelijnck2019multi} propose an application of the MT framework to multi-resolution spatio-temporal problems. 

\paragraph{Alignment}
While %
many heuristic methods for the alignment of data have been developed (\emph{e.g.}~dynamic time warping), the work on alignment of data in probabilistic multi-task learning has been limited to simple shift or scale~\citep{durichen2014multitask}. The multi-task models from statistical literature largely come from geostatistics, \emph{i.e.}~spatio-temporal modelling~\citep{sahu2005recent}.  These models are usually application specific and in most cases use linear transformations for alignments (\emph{e.g.}~\citep{forlani2020joint}).
Another line of work is structured covariance estimation \citep{barnard2000modeling}, \citep{spezia2019modelling}, although these methods do not explicitly model misalignments.

\paragraph{Deep GPs \& Alignment} While some of the proposed approaches consider deeps GPs (DGPs) as models for  sequences~\citep{boustati2019non, hamelijnck2019multi}, their motivation is applications where the data are known to be generated by functional composition. Explicit temporal mismatch between sequences is not considered (no monotonic constraint on latent layers).
Importantly, the existing works that do model temporal alignment, \citep{kaiser2018bayesian, duncker2018temporal}, assume the groups of tasks to align are known a-priori. This significantly simplifies the alignment problem and this knowledge is not present in the general MT learning formulation.
The motivation for our work is closer to the GP alignment models~\citep{kazlauskaite2018sequence, kazlauskaite2019gaussian}. 
Contrary to these models, we propose a joint probabilistic approach that is motivated by MT applications rather than a regularised GP regression model that is aimed primarily at an alignment goal.

\paragraph{Sparse Variational GPs} One of the weaknesses of the traditional GP formulation is the poor computational scaling with respect to the number of observations; this is especially apparent in the multi-task case where the computational cost scales as $\mathcal{O}(J^3 N^3)$ for $J$ tasks, each with $N$ observations. Therefore, efficiency issues of such multi-task models have been considered in most papers on the subject, \emph{e.g.}~\citep{Alvarez:2012, hamelijnck2019multi, zhe2019scalable}. 

Bringing the ideas from \cite{kazlauskaite2018sequence} to a multi-task scenario, we propose a method that is able to  model flexibly misalignments in GP multi-task learning.
In this work, we follow the sparse GP approach of~\cite{titsias2009variational} and the subsequent stochastic extension of the variational inference framework~\citep{hensman2013gaussian}. 

\section{Background}\label{sec:background}

\paragraph{Gaussian Processes}
We make use of GPs for the Bayesian modelling of time-series data as they offer a convenient way of defining priors over functions~\citep{williams2006gaussian}. 
We denote a GP functional prior, fully specified by a mean function $m(x)$ (typically assumed to be zero) and a covariance function $k(x, x')$, 
as $f(x) \sim \mathcal{GP}\big(m(x), k(x, x')\big) $.
Thus given a finite set of inputs $x_1, \dots , x_N$, we may draw samples $f(x_1), \dots , f(x_N)$ from the GP prior: $f(x_1), \dots , f(x_N) \sim \mathcal{N}(0,\mathbf{K})$ where $K_{ij} := k(x_i, x_j)$.
The model of the data is $y_i = f(x_i) + \epsilon_i$ where $\epsilon_i \sim \mathcal{N}(0, \sigma^2)$ is Gaussian noise. Learning in exact GP models typically consists of inferring the hyper-parameters of a specified covariance function. This can only be performed in closed form under Gaussian likelihoods and at high computational expense $\mathcal{O}(N^3)$ (due to the inversion of the covariance matrix); approximate inference methods provide more efficient inference and relax the likelihood restrictions.

\paragraph{Multi-task GPs}
In multi-task GPs (MTGPs), we assume that observations of some latent functions~$\F \in \mathbb{R}^{N \times J}$ comprise $J$ sequences (corresponding to $J$ tasks), each of which we model using a GP and, furthermore, there exists some unknown correlation structure among the tasks. 
In one of the most widely used models, intrinsic coregionalization model (ICM,~\cite{goovaerts1997geostatistics}),
the joint model is then defined as $\mathsf{vec}(\F) \sim \mathcal{N}( \mathsf{vec}(\mathbf{M}), \mathbf{K}_\psi \otimes \mathbf{K}_\theta)$ with some mean $\mathbf{M}$ and a covariance structure that adopts a Kronecker product form where the $J \times J$  covariance matrix $\mathbf{K}_\psi$ captures the correlations among the $J$ tasks, while the $N \times N$ matrix $\mathbf{K}_\theta$ models the correlations between the N observations in each of the $J$ sequences.

While we base our work on ICM and its latent variable extension, the idea of temporal alignment in multi-task learning is general and can be applied to other MTGP models. For a review of other MTGP models see \cite{liu2018remarks}.

\paragraph{GP Latent Variable Alignment Model} 
Here we give a brief introduction to the work of \cite{kazlauskaite2019gaussian}, which inspired this work, and highlight the important differences. GP Latent Variable Alignment (GP-LVA)~\citep{kazlauskaite2018sequence, kazlauskaite2019gaussian} is designed with the primary goal of sequence alignment.
The method models aligned functions as pseudo-observations that should have high likelihoods under two separate parts of the model simultaneously. One part models temporal consistency using GPs, the other models inter-sequence relationship using a GP-LVM. While this trick works empirically, it has a number of downsides:
(1)~predictive posterior is conditioned on both data and pseudo-observations, which leads to underestimated uncertainty,
(2)~it is unclear how to generate samples from the model since the dependencies between two parts of the model are induced via the pseudo-observations of aligned functions, while marginalizing out these aligned functions leads to the two parts of the model becoming independent,
and (3)~to reconcile the two parts of the model the authors add heuristic noise terms which reduces the model interpretability. In contrast, our model is a fully generative probabilistic model, cast in terms of a standard GP framework. We formulate a proper evidence lower bound, used for inference and hyperparameter learning. 
The aligned functions are treated in the Bayesian way, with variational posterior distributions and the latent warpings are explicitly modelled and marginalised out (rather than taking a MAP estimate as in GP-LVA).
In summary, the model of \cite{kazlauskaite2019gaussian} could also be seen as a partial approximation to our model while we preserve the full model and use approximate inference.
Moreover, due to its use of pseudo-observations, GP-LVA can not handle missing data in a principled way, suffering from underestimated uncertainty particularly in the areas of missing data.

\label{sec:monotonic_background}
\paragraph{Monotonic GPs} To model the aligned functions we must account explicitly for the unknown misalignment subject to the constraint that it must be monotone. %
There have been a number of proposals for approximations to define monotonic GPs: for example, 
truncated or finite-dimensional approximations~\citep{Maatouk:2017, Lopez-Lopera:2019};
incorporating virtual derivative information~\citep{Riihimaki:2010};
projections onto spaces of monotone functions~\citep{Lin:2014}; 
or through non-linear transformations~\citep{Andersen:2018}. 
Instead, we define a guaranteed monotonic stochastic process through a differential flow that provides smooth solutions that are guaranteed monotonic across the entire domain without distorted uncertainty estimation~\citep{ustyuzhaninov2020monotonic}. 
We extend this work with a modified model, better considered as an Ordinary Differential Equation (ODE) with an uncertain drift function, rather than the Stochastic Differential Equation (SDE) of~\cite{ustyuzhaninov2020monotonic}, and provide a new approach for efficient inference.

\section{Model}\label{sec:model}
Consider a data vector $\y_j=\{y_{jn}\}_{n=1}^N$, where $y_{jn}$ is a noisy observation of the function $f_j(\x_{jn})$ and a corresponding input vector $\x_j=\{\x_{jn}\}_{n=1}^N$ for each of $J$ tasks with a length %
of $N$ observations.
For clarity of notation, we will consider tasks to be of the same length; in the case of different lengths, the $\mathsf{vec}$ operator should be replaced with concatenation.
Let $\f_j$ denote the values of the function $f_j(\cdot)$ at inputs $\x_j$. We denote all input data as $\X=[\x_1,..,\x_J]$ and the observed data matrix as  $\Y=[\y_1,..,\y_J]$. Stacked vectors of the observed data and inputs is then denoted as $\y=\mathsf{vec}(\Y)$ and $\x=\mathsf{vec}(\X)$ respectively. Similarly, $\F=[\f_1,..,\f_J]$ and $\f=\mathsf{vec}(\F)$.

\subsection{Multi-Task Gaussian Processes (MTGPs)}

First, we introduce a latent variable version of the standard MTGP formulation. 
Typically, in ICM the correlations between tasks are modelled with a free-form covariance (in the absence of task-specific features).
Similarly to ~\cite{Stegle:2011} and ~\cite{ dai2017efficient}, we choose a more flexible approach and use latent variables to model the inter-task dependencies. 
Each task is assumed to have a corresponding latent variable $\z_j\in \mathbb{R}^Q$. We put a spherical Gaussian prior on the latent variables $\z_j \sim \mathcal{N}(\z_j|\0, \I)$. 
The functions are assumed to be a joint sample from a GP with a separable kernel over the latent and input spaces taking the form
$\f \given \z, \X \sim \mathcal{GP}(\f \given \0, \K)$,
where $K_{jn,j'n'} = k_{\psi}(\z_j,\z_{j'}) \, k_{\theta}(\x_{n},\x_{n'})$ is the covariance between the $n$-th input of $\f_j(\cdot)$ and the $n'$-th input of $\f_{j'}(\cdot)$. The kernel $k_{\psi}(\cdot,\cdot)$ acting on latent variables determines similarities across the tasks, and kernel $k_{\theta}(\cdot,\cdot)$ affects the properties of each of the tasks separately. This approach allows for the explicit incorporation of priors on the inter-task dependencies.

\subsection{Aligned Multi-Task GPs (AMTGPs)}

In a standard MTGP model, the tasks are assumed to be \emph{aligned} across inputs. However, especially in realistic scenarios where the input is time, these tasks might be misaligned due to various unmodelled factors. %
To overcome this, we account for temporal misalignments between tasks by warping the inputs with latent monotonic functions; this reflects the assumption that misalignment manifests as delays and phase shifts but not as non-causal permutations of time. 
The values of $\f_j$ are modelled using inferred aligned input values $\g_j$.

\label{sec:monotonic_warps_model}
\paragraph{Monotonic Warps} Here we consider warps to be independent between tasks (\emph{e.g.}~sampling errors or phase noise) that we model as smooth monotonic functions. For each task $j$, the alignment is modelled with a monotonic function $g_j(x_{jn})$ and the corresponding values of the function for all inputs $\x_j$ are denoted by $\g_j$. 
As discussed in Sec.~\ref{sec:monotonic_background}, there are a number of different approaches to constrain a GP to be monotonic.
We build upon the monotonic GP flow solution proposed by~\cite{ustyuzhaninov2020monotonic}.
There, a stochastic process is defined as a fixed time, initial value solution to a Stochastic Differential Equation (SDE). 
Subject to constraints on how the SDE is defined (using a GP field) and the inference procedure, every sampled solution is guaranteed not to permute the inputs and, therefore, remain monotone.
In contrast to the SDE formulation, we pose a monotonic process as the solution to an Ordinary Differential Equation (ODE) $\dd{u} = w(u) \dd{\tau}$ but where the drift function $w(\cdot)$ is uncertain; we place a GP prior over the drift function $w(u) \sim \mathcal{GP}\big(\mathbf{0}, K_{\omega}(u, u)\big)$. 
We thus define the monotonic warping process $g_j(x)$ as the solution, at $\tau = T$, to the ODE:
\begin{equation}
    g_j(x) := u_{j}(\tau = T ; x) = \int_{0}^{T} w_{j}\big( u(\tau) \big) \, \dd{\tau}
\end{equation} subject to initial condition $u(\tau = 0) := x \,$.
To draw a sample $\g_j^{(s)}$ from the process we first draw a posterior \emph{function} sample from the GP $w^{(s)}(\cdot)$ and solve the resulting ODE jointly for all elements in $\x_j$. The use of an ODE rather than an SDE has the advantage of guaranteed smoothness, from the GP prior on $w(\cdot)$, and allows the use of higher-order adaptive solvers, \emph{e.g.}~Runge-Kutta~\citep{schober2014rungekutta}.

The difficulty presented is the requirement to draw a single \emph{function} sample from the GP for  integration by the ODE solver. 
Typically, we draw joint samples from a GP posterior only for a given finite set of input locations.
For the ODE, we do not know a priori all the input locations; they are only revealed sequentially as the solver progresses and depend on function evaluations for previous values of $\tau$. 
We solve this problem using a recent result in efficient path-wise sampling from GP posteriors from~\cite{wilson2020pathsampling}.
This allows us, not only, to evaluate the sampled function $w^{(s)}(\cdot)$ sequentially, but also to perform the evaluation efficiently without performing expensive Cholesky operations (which scale cubically with the number of posterior samples).
Further details of this inference procedure are provided in Sec.~\ref{sec:monotonic_inference}.
We note a concurrent work by \cite{hegde2021bayesian} that similarly uses path-wise sampling from a GP to infer posterior of an ODE system.

\paragraph{Joint Distribution} The joint probability distribution factorises as
\begin{equation}
\begin{split}
        p(\y, \f, \z,\g \given \X, \beta, \theta, \psi) = p(\f \given \z, \g, \theta,\psi) \qquad \qquad \;\\
    \prod_{j=1}^J p(\g_j \given \x_j, w_j) \, p(w_j) \, p(\z_j) \prod_{n=1}^N p(y_{jn} \given f_{jn}, \beta)  \,.
\end{split}
\end{equation}
The terms in the joint distribution are:
\begin{align}
    \y \given \f, \beta &\sim \mathcal{N}(\y \given \f, \beta^{-1}\mathbf{I}_{JN}) , \nonumber\\
    \f \given \z,\g &\sim \mathcal{GP}\big(\f \given \0, K_\psi(\z_j,\z_{j'}) \odot K_{\theta}(\g_{j,n},\g_{j',n'})\big) ,\nonumber\\
    \g_j \given \x_j, w_j &\sim \mathsf{Monotonic\ Process} \big(\g_j \given \x_j, w_j \big) , \\
    w_j &\sim \mathcal{GP} \big(w_j \given \mathbf{0}, K_{\omega}(u_j,u_j)\big) , \nonumber\\[2pt]
    \z_j &\sim \mathcal{N}(\z_j \given \0, \mathbf{I}_Q) , \nonumber
\end{align}
where $\odot$ denotes a tensor product such that $[K_{\f,\f}]_{jn,j'n'} = [K_\psi(\z_j,\z_{j'})]_{j,j'} [K_{\theta}(\g_{j,n},\g_{j',n'})]_{jn,j'n'}$.
The functional values $\f$ are fully correlated across all inputs and tasks that leads to the problematic complexity of $\mathcal{O}(J^3N^3)$. Since all $\{\g_j\}$ are now different, we can no longer utilise the Kronecker structure, as was suggested in previous work, \emph{e.g.}~\cite{zhe2019scalable}. 
To address this %
we formulate a stochastic variational inference scheme, following the framework of~\cite{hensman2013gaussian}.

\section{Inference} \label{sec:inference}
Several parts of the model pose distinct challenges for inference. Firstly, the covariance of $\f$ depends on both latent variables $\z$ and the warps $\g$, hence we can not marginalize them out in closed form. Secondly, even if we use point estimates for $\z$ and $\g$ (\emph{e.g.}~MAP), the resulting covariance matrix would be of size $JN \times JN$, which is prohibitively expensive to invert. Notice, that we cannot use Kronecker decomposition of the covariance, a typical efficiency trick in multi-task GP models ~\citep{Stegle:2011, dai2017efficient}, in the case of misaligned or missing data. To deal with these issues, and avoid point estimates, we adopt a sparse Stochastic Variational Inference (SVI) scheme.

We wish to compute the marginal likelihood of the data
$p(\y \given \X) = \int p(\y \given \g, \z) \, p(\g \given \X) \, p(\z) \, \dd\z \, \dd\g$.
This integral is intractable as both latent variables $\z$ and warps $\g$ appear nonlinearly inside the inverse of the covariance matrix. 
To address this, we use a variational approach and introduce separable distributions over the latent variables $q(\z) := \prod_{j=1}^J q(\z_j)$ and warps $q(\g) := \prod_{j=1}^J q(\g_j)$ to approximate the true posterior $p(\z, \g \given \y,\X)$. 
The log marginal likelihood can then be bounded using Jensen's inequality:
\begin{multline}
     \log{p(\y \given \X)} %
     \geq \, \EX_{q(\z)q(\g)}\big[\log{p(\y \given \z,\g)}\big]\\
      - \KL\diverge{q(\z)}{p(\z)} -\KL\diverge{q(\g)}{p(\g)} \,.
\end{multline}
The expectation is still intractable, but we can further bound $\mathcal{L}_1 := \log p(\y \given \z,\g)$ using the sparse VI approach of~\cite{titsias2009variational}.

\subsection{Sparse Stochastic Variational Inference}

To make progress, we augment our model by introducing a set of inducing variables.
Consider a set of $M$ auxiliary variables $\h \in \mathbb{R}^M$ evaluated at some artificial pseudo-inputs $[ \Xaux,\Zaux ]$, where $\Xaux \in \mathbb{R}^M$ and $\Zaux \in \mathbb{R}^{M\times Q}$. We may then define an augmented joint distribution as 
\begin{multline}
p(\y,\f, \h, \z,\g  \given \X, \Xaux, \Zaux) = p(\y \given \f)  \\
p(\f \given \h, \z, \g, \Xaux, \Zaux) \, p(\h \given \Xaux, \Zaux) \, p(\g \given \X) \, p(\z) \,.
\end{multline}

Following the approach of~\citep{titsias2009variational}, we define a sparse approximation to the posterior distribution over $\f$ using the inducing variables.
Omitting the dependence on $\X$ for clarity, the exact posterior over $\f$ in the augmented model can be described by the predictive Gaussian distribution%
\begin{equation}
\begin{split}
     p(\f \given \y, \g, \z) 
     & = \int p(\f \given \h, \y, \g, \z) \, p(\h \given \y, \g, \z) \, \dd \h \,.
\end{split}
\label{eq:exact_post}
\end{equation}
Suppose that $\h$ is a sufficient statistic for $\f$, meaning that for any new inputs $[\Xstar,\Zstar]$ and the corresponding function values $\Fstar$, we have $\Fstar \perp \f \given \h  $ or $p(\Fstar \given \h,\f) = p(\Fstar \given \h)$. 
Similarly to~\cite{titsias2009variational} (see supplement for more details),
it follows %
that we can drop the dependence on $\y$ in the posterior such that  $p(\f \given \h, \y, \g, \z) = p(\f \given \h, \g, \z)$.
Under this assumption, we can write an approximation to the exact posterior in \eqref{eq:exact_post} as ${q(\f) = \int p(\f  \given \h, \g, \z) \, q(\h) \dd \h}$, where we specify that ${q(\h) := \mathcal{N}(\h \given \m_{\mathrm{h}}, \S__{\mathrm{h}})}$. The variational distribution over $\f$ and $\h$ is then $q(\f, \h) = p(\f \given \h, \g, \z) \, q(\h)$.

Now, using the augmented model and the variational distribution $q(\f, \h)$, we can write the lower bound on $\mathcal{L}_1$ as
\begin{equation}
\begin{split}
    \mathcal{L}_1 &= \log p(\y \given \z,\g) \\
    &= \log{\int p(\f | \h, \z, \g) q(\h) \frac{p(\y \given \f) p(\h)}{q(\h)} \dd \h \dd \f}\\
    &\geq \mathcal{L}_2 - \KL\diverge{q(\h)}{p(\h)} \;, \\
    \mathcal{L}_2 &:= \int q(\h) \Big[\int p(\f  \given  \h, \g, \z) \log{ p(\y \given \f) }\dd\f \Big]\dd\h \;.
\end{split}
\label{eq:L1}
\end{equation}
While it is possible to ``collapse'' the distribution $q(\h)$ by finding its optimal parameters~\citep{titsias2010bayesian}, we choose to follow the stochastic VI approach of~\cite{hensman2013gaussian} and keep the explicit representation of the inducing variables.

After marginalizing out $\f$ and $\h$ in the $\mathcal{L}_2$ term of (\ref{eq:L1}), 
please see the supplement for detailed derivation, 
the overall lower bound $\mathcal{L} \leq \log{p(\y)}$ takes the form
\begin{align}
     \mathcal{L} &= \sum^{J}_{j=1}\Big\{  \EX_{q(\z_j)q(\g_j)}\big[\log \mathcal{N}(\y_j \given \K_{f_{j}h}\K_{hh}^{-1}\m, \beta^{-1}\I) \nonumber\\
     &\;- \frac{1}{2} \mathsf{Tr}[\Lambda_j\S_] - \frac{\beta}{2} \mathsf{Tr}[\Sigma_j] \big] - \KL\diverge{q(\z_j)}{p(\z_j)} \label{eqn:final_lower_bound} \\
     &\;- \KL\diverge{q(\g_j)}{p(\g_j)} \Big\} - \KL\diverge{q(\h)}{p(\h)} ,   \nonumber
\end{align}
where we have matrices $\Lambda_j := \beta \,\K_{hh}^{-1}\K_{hf_{j}}\K_{f_{j}h}\K_{hh}^{-1}$ and ${\Sigma_j := \K_{f_{j},f_{j}} - \K_{f_{j}h}\K_{hh}^{-1}\K_{hf_{j}}}$.
The bound is factorised over sequences; combined with the separable kernels, we only need to compute the following expectations $\EX_{q(\z_i)}\big[ K(\z_j,\Zaux)\big]$, $\EX_{q(\z_i)}\big[K(\Zaux,\z_j)K(\z_j, \Zaux)\big]$,  $\EX_{q(\g_j)}\big[K(\g_{j}, \Xaux)\big]$ and $\EX_{q(\g_j)}\big[K(\Xaux, \g_{j}) K(\g_{j}, \Xaux)\big]$. In general, these can be approximated with sampling; the expectations under $q(\z)$ can be computed analytically for some kernels, \emph{e.g.}~the squared exponential.

Using sparse VI, the GP methodology allows for the Bayesian treatment of latent variables and warps, as well as reducing the time complexity to $\mathcal{O}(JNM^2)$. Stochastic VI adds the possibility of further complexity reduction through training using mini-batches of tasks.
\subsection{Efficient SVI for Monotonic Warps}
\label{sec:monotonic_inference}

Calculation of~\eqref{eqn:final_lower_bound} requires taking expectations over the warps under the approximate posteriors $\{q(\g_j)\}$.
We estimate this bound by drawing samples from the respective monotonic processes. 
In Sec.~\ref{sec:monotonic_warps_model}, we outlined the sampling procedure as drawing a function sample from each GP posterior $w^{(s)}_j(\cdot)$ and then solving the initial value ODE to obtain samples from $q_j(\g_j)$.
As the inputs are unknown a priori, we follow~\cite{ustyuzhaninov2020monotonic} and~\cite{Hedge:2019},
and specify the field using a sparse variational GP~\citep{titsias2009variational}.
For each sequence, we define inducing locations $\tilde{\mathbf{U}}_j$ and pseudo-outputs $\w_j$, and learn an approximate variational posterior $q(\w_j) = \mathcal{N}(\w_j \given \m_{\mathrm{w},j}, \S__{\mathrm{w},j})$.

As the warps are smooth, we found it most efficient to solve the ODE using a simple Euler stepping approach with 10 steps over $\tau \in [0, 1]$ taking gradients with respect to the variational parameters and kernel hyperparameters.
The solver requires the sequential evaluation of a single functional sample $w^{(s)}_j(\cdot)$ at arbitrary locations. 
Standard approaches would require all inputs to be known and a large covariance factorised.
Instead, we make use of an efficient approximation scheme using path-wise samples from~\citep{wilson2020pathsampling}.
We combined Matherson's Rule with a weight-space approximation to sample from the prior using random Fourier features~\citep{rahimi2008randomfeatures}. These samples may then be conditioned on the inducing-locations and samples from their corresponding pseudo-output distributions.
\newcommand{\om}{\boldsymbol{\Omega}}%
Let $\om_j$ be a set of $F$ random Fourier features for the kernel with hyperparameters $\omega_j$ and $\mathbf{b}_j$ be a set of draws from a uniform distribution over $[0, 2\pi)$ such that $\om_j, \mathbf{b}_j \in \mathbb{R}^{F}$.
Then $\phi_j(u) := \sqrt{2 \sigma_{\omega} / F} \cos(\om_j u + \mathbf{b}_j)$ defines a feature space such that $K_{\omega_j}(\U,\U') \approx \boldsymbol{\phi}_j^{\top}\!(\U) \, \boldsymbol{\phi}_j(\U')$.
If we draw samples $\boldsymbol{\alpha}^{(s)} \sim \mathcal{N}(\0, I_{F})$ and $\w^{(s)}_j \sim q(\w_j)$ then
\begin{align}
    w^{(s)}_j(u) &:= \phi_j^{\top}(u) \, \boldsymbol{\alpha}^{(s)} + \boldsymbol{\beta}(u) \;, \\
    \boldsymbol{\beta}(u) &:= K_{\omega_j}(u, \tilde{\U}_j) \, K_{\omega_j}^{-1}(\tilde{\U}_j,\tilde{\U}_j) \big(\w^{(s)}_j - \phi_j^{\top}(\tilde{\U}_j) \, \boldsymbol{\alpha}^{(s)} \big)  \nonumber 
\end{align}
gives a single functional draw of $w_j^{(s)}(\cdot)$ for arbitrary $u$.
Thus, fixing $\boldsymbol{\alpha}^{(s)}$ and $\w^{({s})}_j$ during the ODE solver loop, we efficiently solve for samples $\g_j^{(s)}$ with complexity $\mathcal{O}(N)$. %
\begin{figure*}[t]
\centering
\begin{subfigure}[b]{0.16\textwidth}
     \centering
     \includegraphics[width=\textwidth]{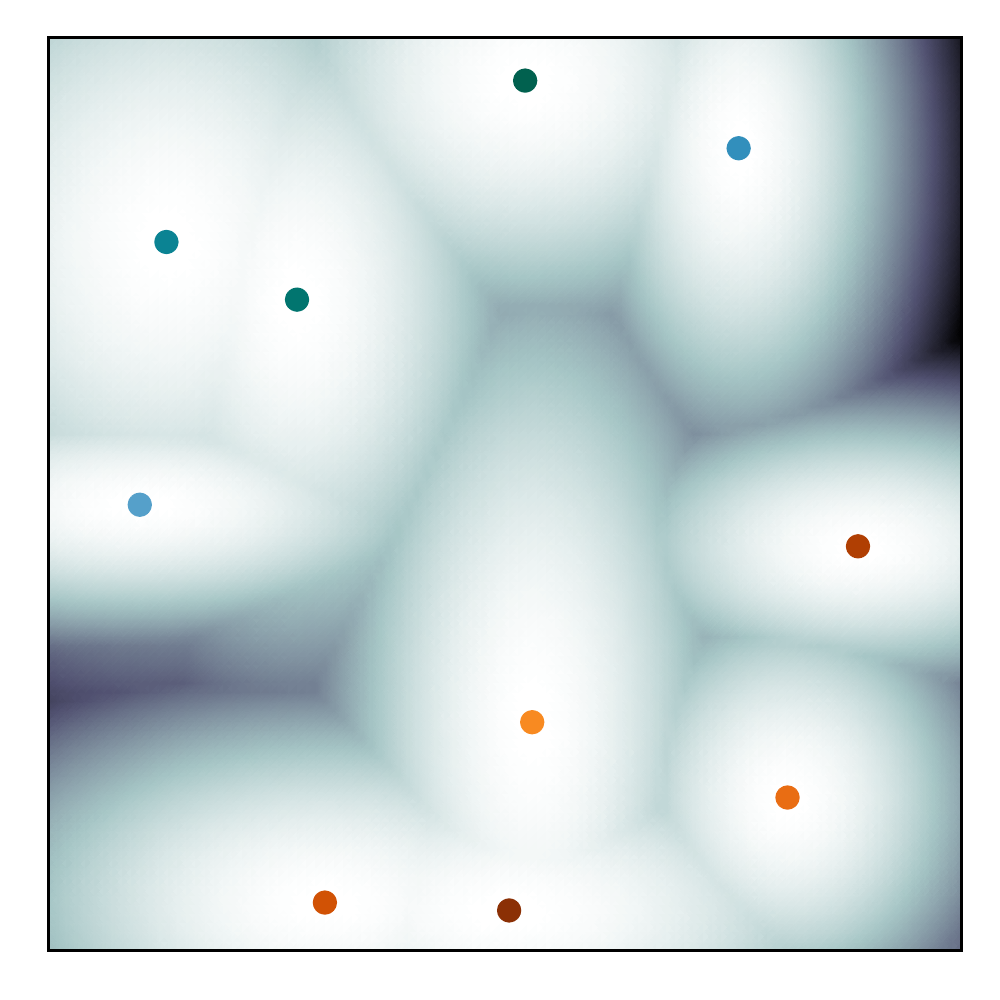}
     \caption{MTGP $\textbf{Z}$}
     \label{fig:synthetic_latent_mtgp}
 \end{subfigure}
 \begin{subfigure}[b]{0.4\textwidth}
     \centering
     \includegraphics[width=\textwidth]{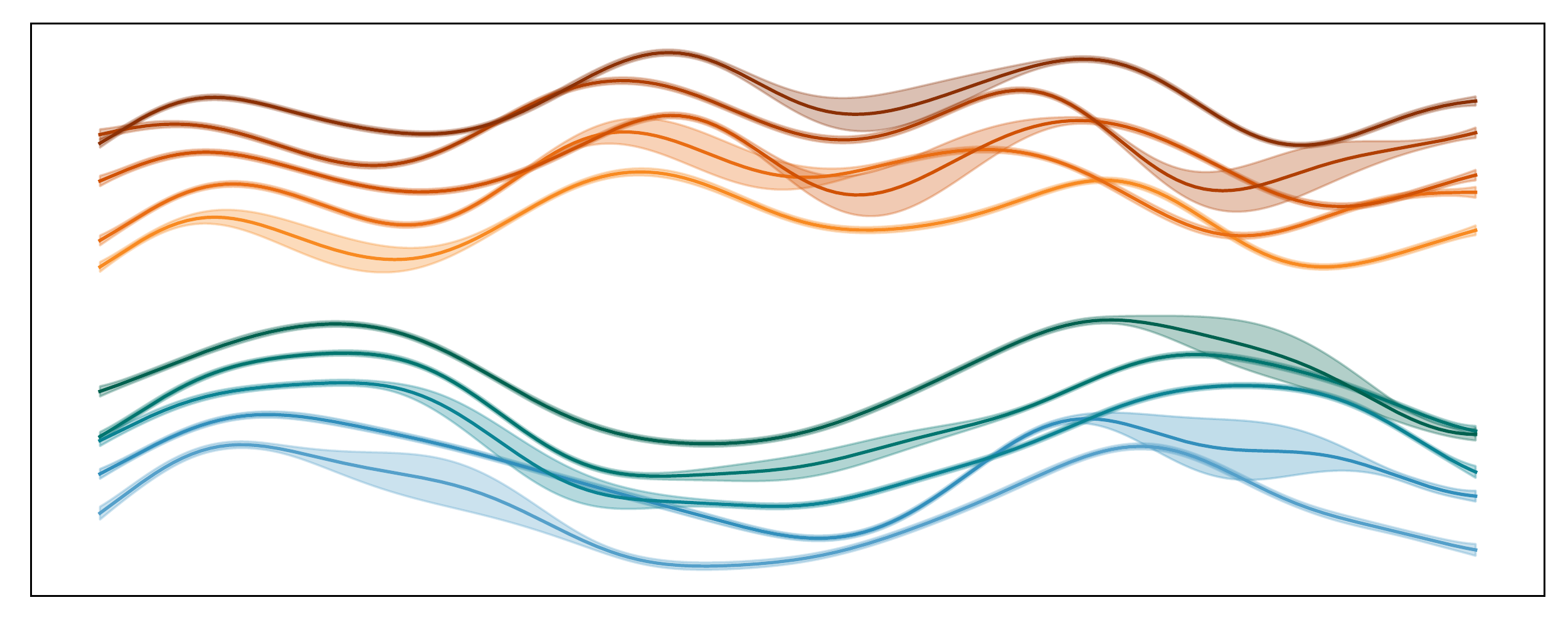}
     \caption{MTGP function posteriors (unaligned)}
     \label{fig:synthetic_f_post_mtgp}
 \end{subfigure}
 \begin{subfigure}[b]{0.4\textwidth}
     \centering
     \includegraphics[width=\textwidth]{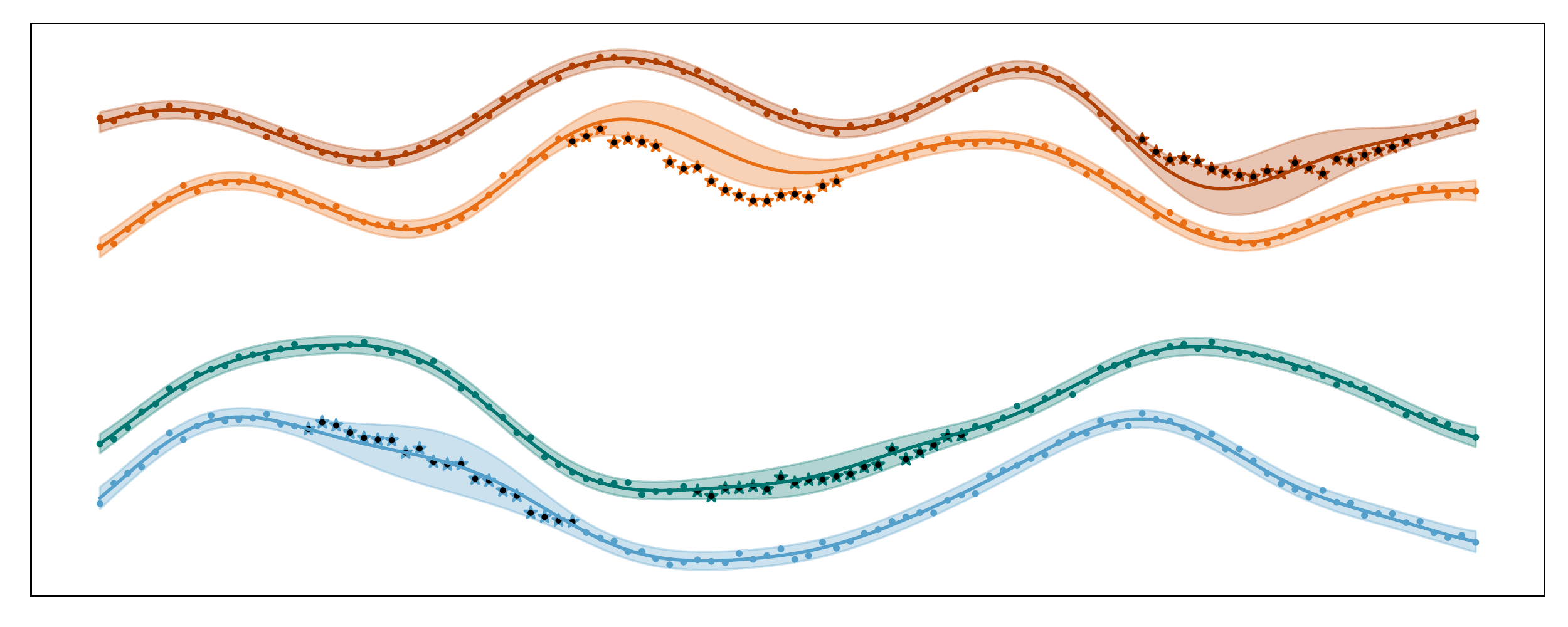}
     \caption{MTGP missing data examples}
     \label{fig:synthetic_data_fit_mtgp}
 \end{subfigure}\\
  \begin{subfigure}[b]{0.16\textwidth}
     \centering
     \includegraphics[width=\textwidth]{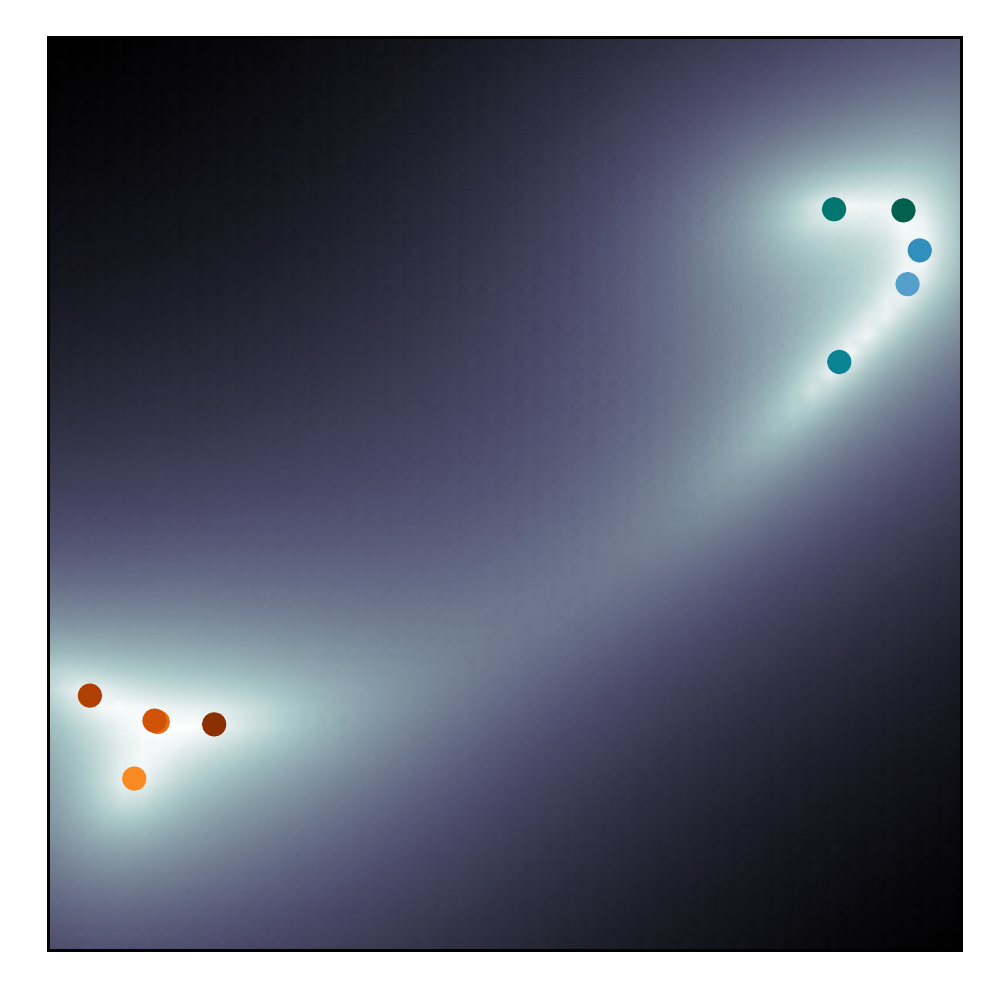}
     \caption{GP-LVA~$\textbf{Z}$}
     \label{fig:synthetic_latent_gplva}
 \end{subfigure}
 \begin{subfigure}[b]{0.4\textwidth}
     \centering
     \includegraphics[width=\textwidth]{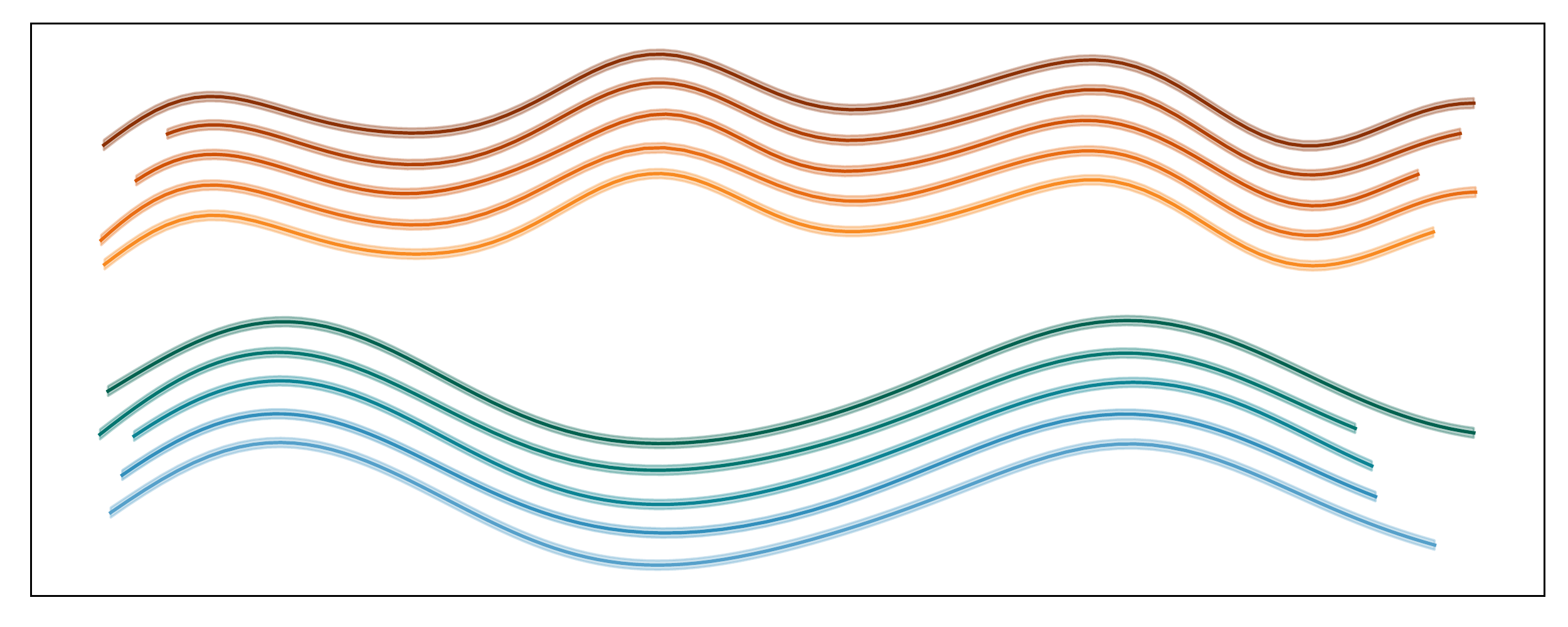}
     \caption{GP-LVA~function posteriors (aligned)}
     \label{fig:synthetic_f_post_gplva}
 \end{subfigure}
 \begin{subfigure}[b]{0.4\textwidth}
     \centering
     \includegraphics[width=\textwidth]{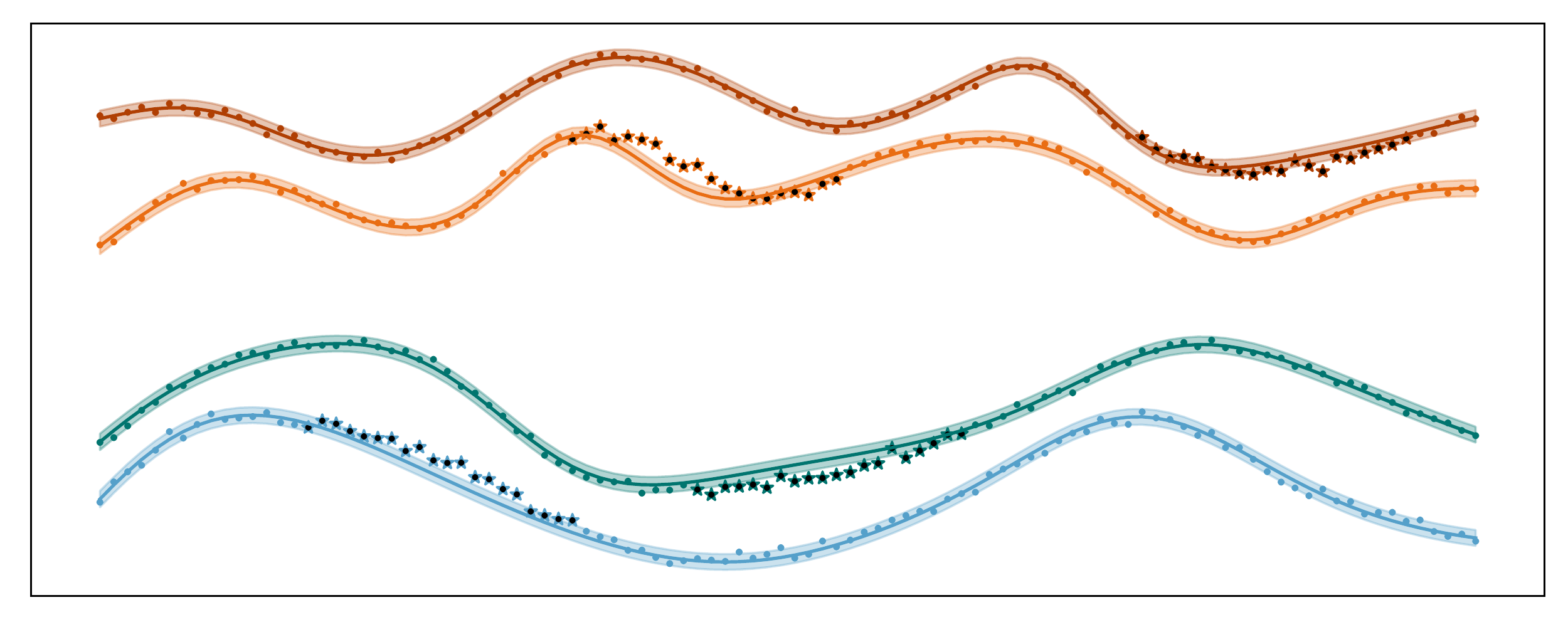}
     \caption{GP-LVA~missing data examples}
     \label{fig:synthetic_data_fit_gplva}
 \end{subfigure}\\
  \begin{subfigure}[b]{0.16\textwidth}
     \centering
     \includegraphics[width=\textwidth]{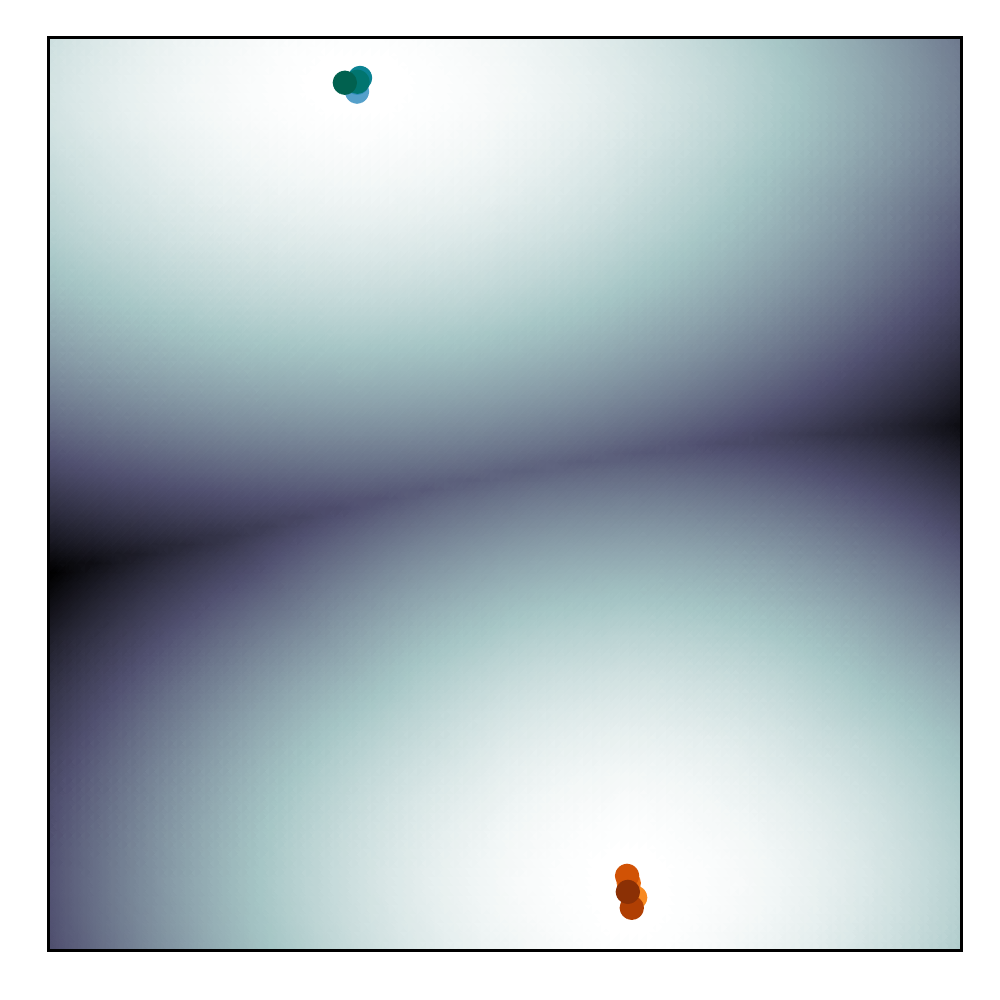}
     \caption{AMTGP $\textbf{Z}$}
     \label{fig:synthetic_latent_amtgp}
 \end{subfigure}
 \begin{subfigure}[b]{0.4\textwidth}
     \centering
     \includegraphics[width=\textwidth]{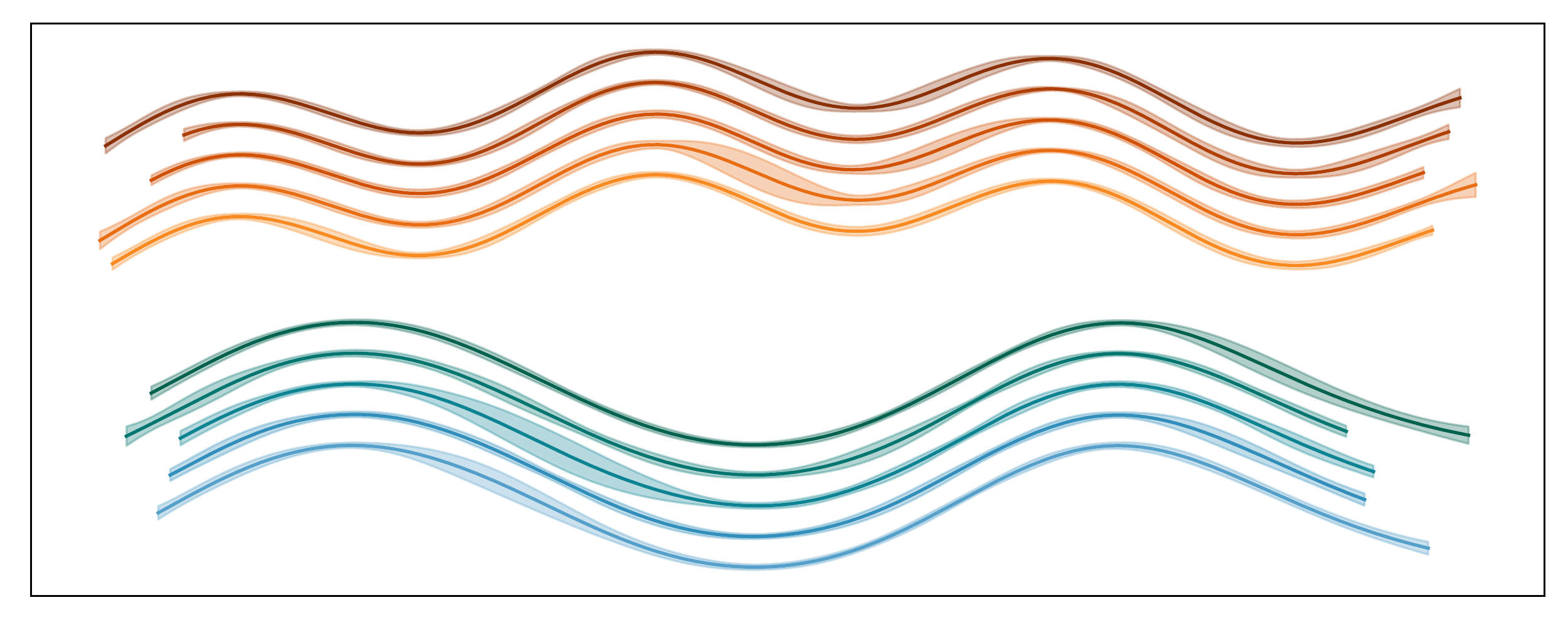}
     \caption{AMTGP function posteriors (aligned)}
     \label{fig:synthetic_f_post_amtgp}
 \end{subfigure}
 \begin{subfigure}[b]{0.4\textwidth}
     \centering
     \includegraphics[width=\textwidth]{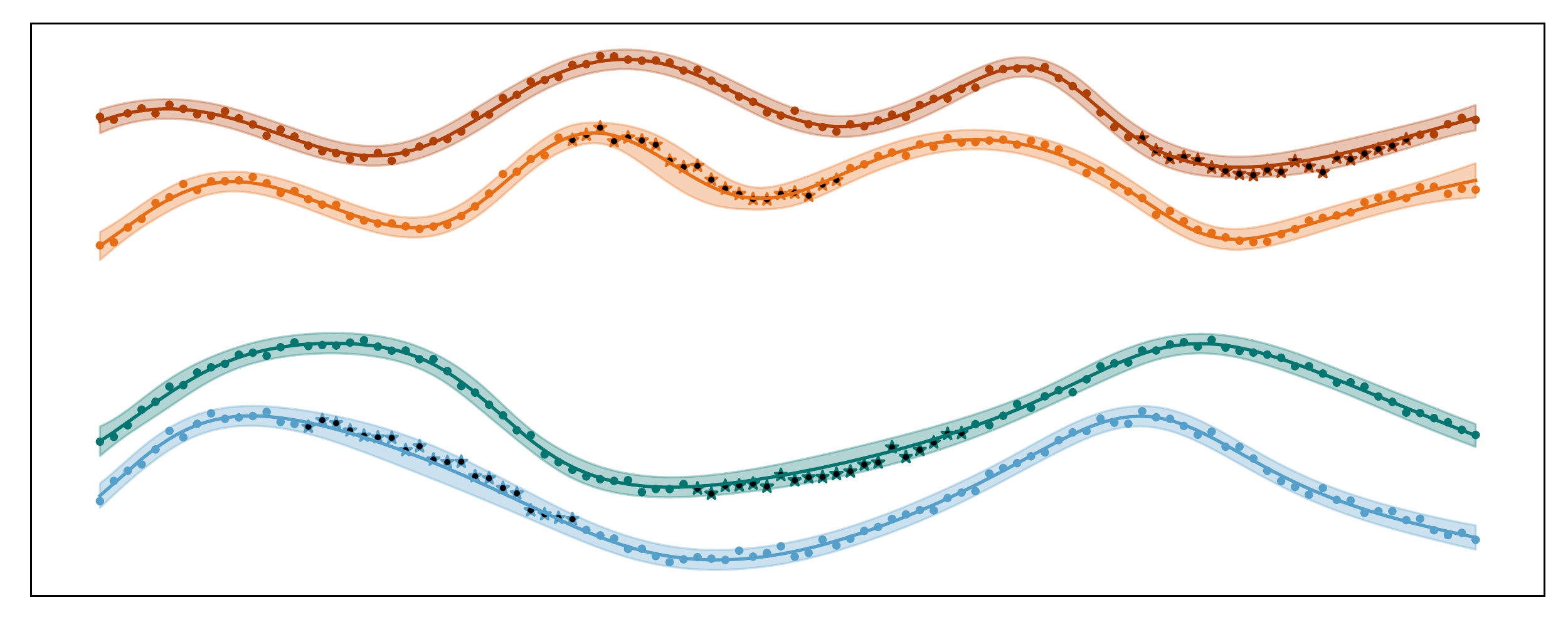}
     \caption{AMTGP missing data examples}
     \label{fig:synthetic_data_fit_amtgp}
 \end{subfigure}
\caption{MTGP (top row), GP-LVA~(middle row) and full AMTGP (bottom row) results on synthetic data for missing data scenario S3.  \subref{fig:synthetic_latent_mtgp}, \subref{fig:synthetic_latent_gplva}~and~\subref{fig:synthetic_latent_amtgp} show log-scaled posterior over the latent space. \subref{fig:synthetic_f_post_mtgp}, \subref{fig:synthetic_f_post_gplva}~and~\subref{fig:synthetic_f_post_amtgp} show posterior over $f$ with $2\sigma$ uncertainty bars.
\subref{fig:synthetic_latent_amtgp}, \subref{fig:synthetic_data_fit_gplva}~and~\subref{fig:synthetic_data_fit_amtgp} show the data and corresponding predictive distributions for two examples from the two different groups of tasks; this clustering \subref{fig:synthetic_latent_amtgp} is correctly identified by AMTGP. 
Data points shown in black are the missing values and plots %
are vertically offset for clarity.
MTGP has to introduce large error bars to account for the missing data~\subref{fig:synthetic_latent_mtgp} and, while it aligns correctly, 
GP-LVA~becomes overconfident~\subref{fig:synthetic_data_fit_gplva}; 
in contrast, full AMTGP correctly accounts for the uncertainty in the warps and accurately models the missing data distribution~\subref{fig:synthetic_data_fit_amtgp}.}
\label{fig:synth_comparison}
\end{figure*}

\subsection{Learning} 
Training alternates two steps: %
(1)~using natural gradients for the variational distributions of the inducing variables $q(\h)=\mathcal{N}(\h \given \m_{\mathrm{h}}, \S__{\mathrm{h}})$ and $q(\w_j) = \mathcal{N}(\w_j \given \m_{\mathrm{w},j}, \S__{\mathrm{w},j})$ (see~\cite{hensman2013gaussian} for details); 
and (2)~estimating $q(\z_j)=\mathcal{N}(\z_j \given \m_{\mathrm{z},j}, \mathsf{diag}( \s_{\mathrm{z},j}))$ alongside the noise precision $\beta$ and kernel hyperparameters $\theta,\{\omega_j\}$ using the Adam optimizer~\citep{kingma2014adam}. 
We fix the latent space lengthscale and variance hyperparameters $\psi$ to 1 to avoid excessive parameterization and  initialise the latent variables $\z$ using linear PCA.
We use the \texttt{GPflow} framework~\citep{GPflow2017} and the \texttt{GPflowSampling} path sampling toolkit~\citep{wilson2020pathsampling}.

\begin{table*}[t]
\centering
\caption{Results on the Synthetic Data.}\vspace{-0.25cm}
\begin{adjustbox}{max width=1.\textwidth}
\begin{tabular}{rrrrrrr}%
\toprule
 & \multicolumn{2}{c}{S1} & \multicolumn{2}{c}{S2} & \multicolumn{2}{c}{S3}  \\
\midrule
 & \multicolumn{1}{c}{Train} & \multicolumn{1}{c}{Test} & \multicolumn{1}{c}{Train} & \multicolumn{1}{c}{Test} & \multicolumn{1}{c}{Train} & \multicolumn{1}{c}{Test} \\

\midrule

MTGP (SMSE) & 0.0069 $\pm$ 0.0003 & 0.0102 $\pm$ 0.0009 & 0.0072 $\pm$ 0.0006 & 0.1573 $\pm$ 0.0624 & 0.0068 $\pm$ 0.0002 & 0.157\phantom{0} $\pm$ 0.0663\\

GP-LVA (SMSE) & 0.0082 $\pm$ 0.0004 & 0.0112 $\pm$ 0.0008 & 0.0085 $\pm$ 0.0009 & 1.3053 $\pm$ 1.0803 & 0.0086 $\pm$ 0.0024 & 0.1123 $\pm$ 0.0727 \\

\textbf{M-AMTGP} (SMSE)  & 0.0061 $\pm$ 0.0002  & 0.0097 $\pm$ 0.0009 & 0.0066 $\pm$ 0.0006 & 0.0534 $\pm$ 0.0181 & 0.0062 $\pm$ 0.0001 & 0.0528 $\pm$ 0.0235\\

\textbf{AMTGP}  (SMSE) & 0.0076 $\pm$ 0.0002  & 0.0099 $\pm$ 0.0007 & 0.0079 $\pm$ 0.0007 & 0.052\phantom{0} $\pm$ 0.0193 & 0.0076 $\pm$ 0.0002 & 0.058\phantom{0} $\pm$ 0.0235\\

\midrule

MTGP (SNLP)  & -1930.7 $\pm$ 13.5  & -456.4 $\pm$ \phantom{0}7.6 & -1919.8 $\pm$ 32.4 & -258.5 $\pm$ \phantom{000}52.4 & -1935.7 $\pm$ 11.2 & -167.1 $\pm$ \phantom{00}72.4\\

GP-LVA (SNLP) & -1840.6 $\pm$ 19.7 & -399.7 $\pm$ 16.0 & -1806.4 $\pm$ 41.4 & 14232.6 $\pm$ 12776.9 & -1810.5 $\pm$ 74.5 & 1115.6 $\pm$ 1126.2 \\

\textbf{M-AMTGP}  (SNLP)  &  -2024.9 $\pm$ 10.3 & -460.7 $\pm$ 11.8 &-1997.0 $\pm$ 34.5 & -181.6 $\pm$ \phantom{00}128.3 & -2015.7 $\pm$ \phantom{0}8.1 & -61.9 $\pm$  \phantom{0}242.2\\

\textbf{AMTGP}  (SNLP)  &  -1836.0 $\pm$ 11.8 & -442.0 $\pm$ \phantom{0}7.4 & -1826.6 $\pm$ 36.6 & -245.4 $\pm$ \phantom{000}92.7 & -1812.3 $\pm$ 18.5 & -156.3 $\pm$ \phantom{0}141.3\\

\bottomrule

\end{tabular}
\label{table:synthetic_results}
\end{adjustbox}
\end{table*}

\section{Experiments} \label{sec:experiments}
To show that multi-task GP learning and inference %
benefits from alignment, we compare our AMTGP model against a version without the alignment functionality, denoted MTGP. 
MTGP can be seen as a fully Bayesian version of Latent Variable Multiple Output Gaussian Processes \citep{dai2017efficient}.
To illustrate the benefits of marginalising out the warps, we also add results for the aligned model using point MAP estimates for the warps as in GP-LVA (denoted \MAPAMTGP). The MAP estimates are obtained by optimising a set of auxiliary variables (constrained to be monotone) under a GP prior as proposed by GP-LVA~\citep{kazlauskaite2019gaussian}.
We also make comparison to GP-LVA.

We evaluate AMTGP on synthetic data as well as three real datasets: dynamic emotional facial expressions~\citep{livingstone2018ryerson}, heartbeat sounds~\citep{Bentley:2011}, and respiratory motion traces~\citep{ernst2011compensating} (please see the supplement).
We perform a quantitative evaluation on the task of predicting missing data in three scenarios: 
(S1) data missing at random, 
(S2) a continuous segment of data missing at the same location for all tasks, 
and (S3) continuous segments of data missing at different locations for each task. 
The performance of the two approaches is compared using both the standardised mean squared error (SMSE) and the standardised negative log probability density (SNLP)~\citep{williams2006gaussian}.
The results are presented with statistics over 10 random data amputations.
For all experiments, a Mat\'{e}rn 5/2 kernel is used for the warp differential field GP prior. %
For fair comparison, the number of inducing points for each model is chosen from the ELBO for the full dataset.

\begin{figure*}[t]
\centering
\begin{subfigure}[b]{0.16\textwidth}
     \centering
     \includegraphics[width=\textwidth]{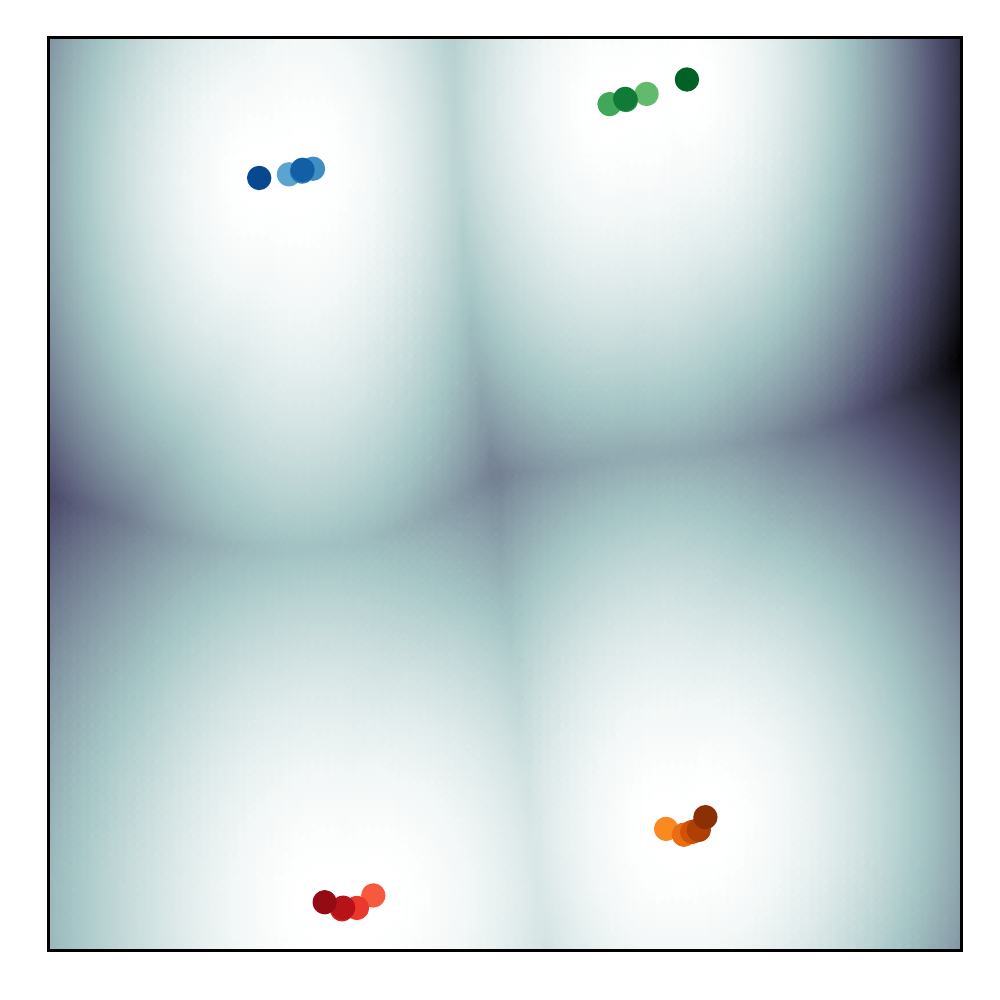}
     \caption{MTGP $\textbf{Z}$}
     \label{fig:face_exp_1_latent_mtgp}
 \end{subfigure}
 \begin{subfigure}[b]{0.4\textwidth}
     \centering
     \includegraphics[width=\textwidth]{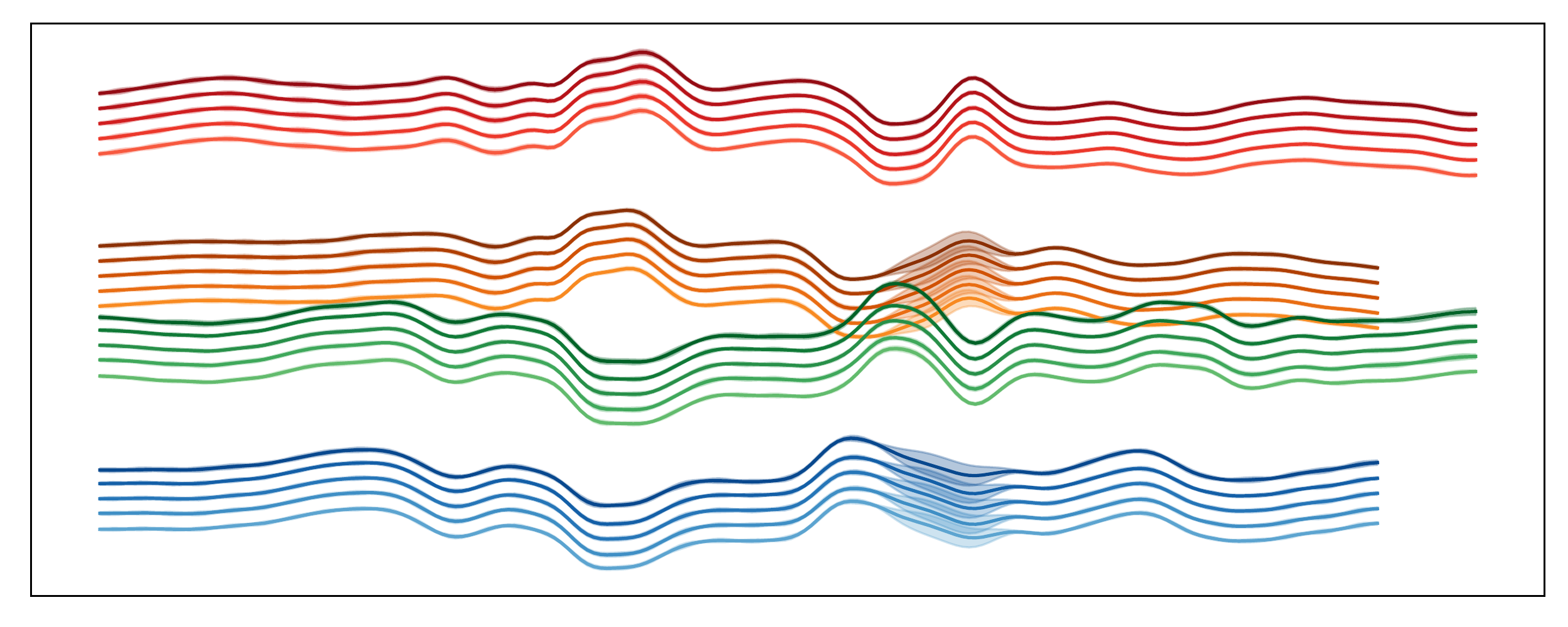}
     \caption{MTGP function posteriors (unaligned)}
     \label{fig:face_exp_1_f_post_mtgp}
 \end{subfigure}
 \begin{subfigure}[b]{0.4\textwidth}
     \centering
     \includegraphics[width=\textwidth]{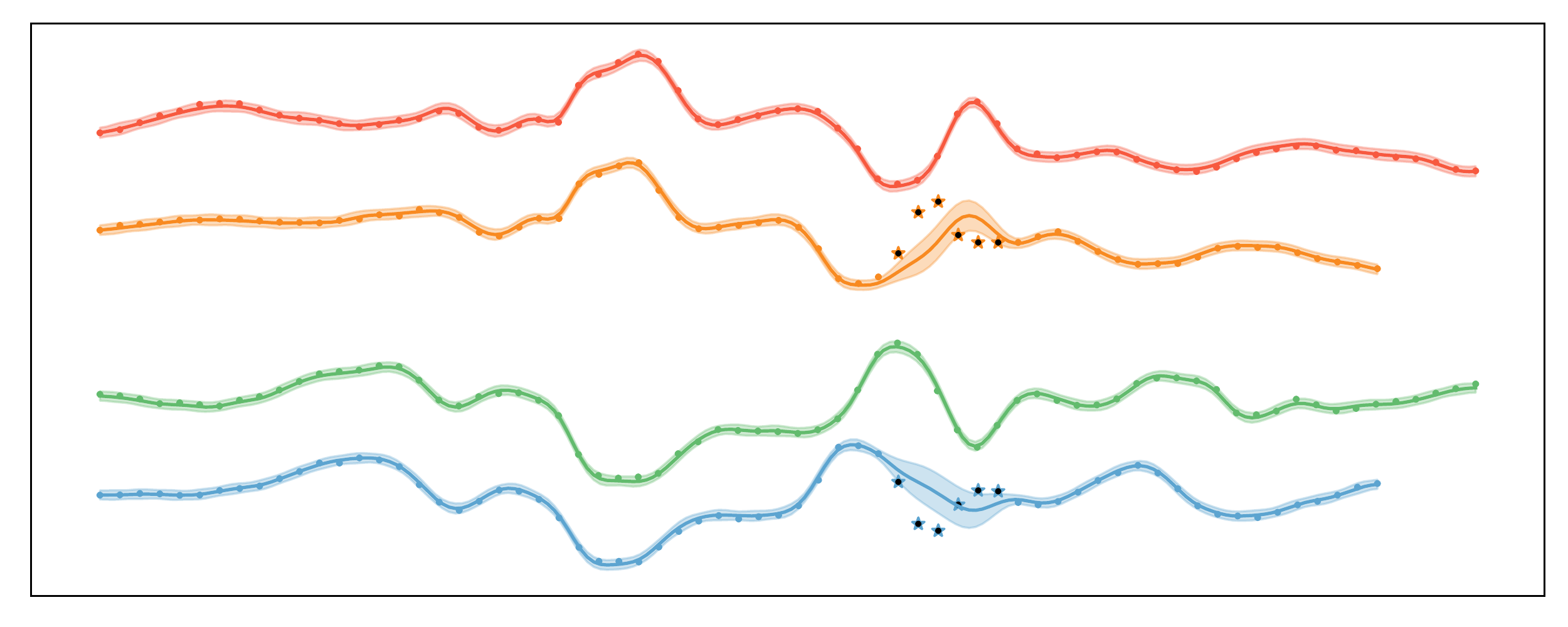}
     \caption{MTGP missing data examples}
     \label{fig:face_exp_1_data_fit_mtgp}
 \end{subfigure}\\
  \begin{subfigure}[b]{0.16\textwidth}
     \centering
     \includegraphics[width=\textwidth]{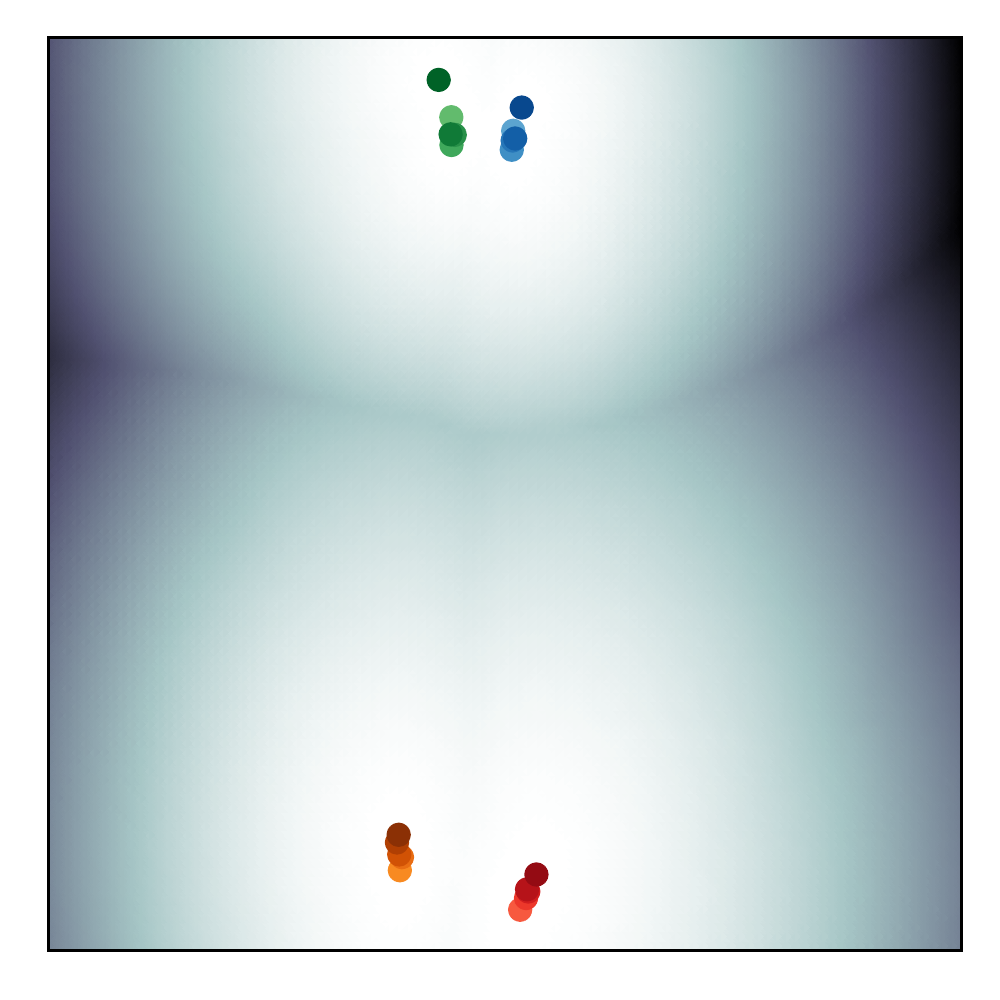}
     \caption{AMTGP $\textbf{Z}$}
     \label{fig:face_exp_1_latent_amtgp}
 \end{subfigure}
 \begin{subfigure}[b]{0.4\textwidth}
     \centering
     \includegraphics[width=\textwidth]{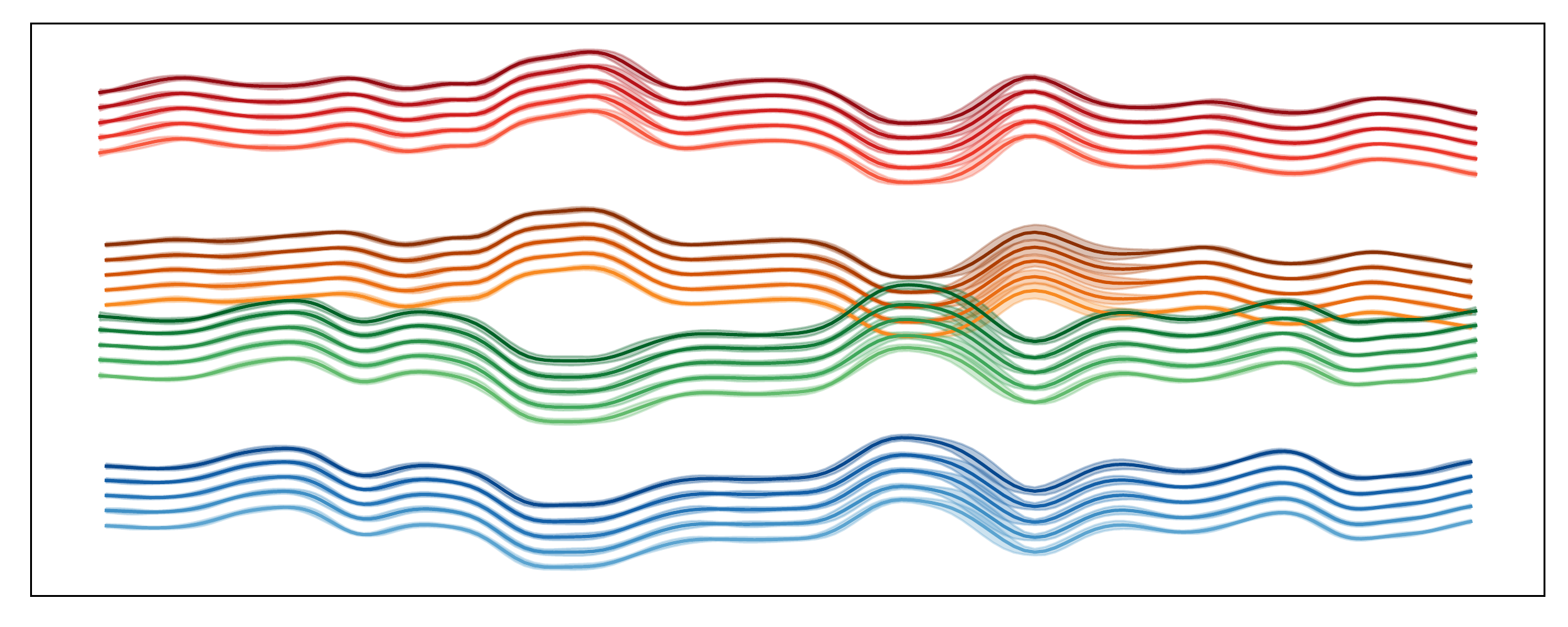}
     \caption{AMTGP function posteriors (aligned)}
     \label{fig:face_exp_1_f_post_amtgp}
 \end{subfigure}
 \begin{subfigure}[b]{0.4\textwidth}
     \centering
     \includegraphics[width=\textwidth]{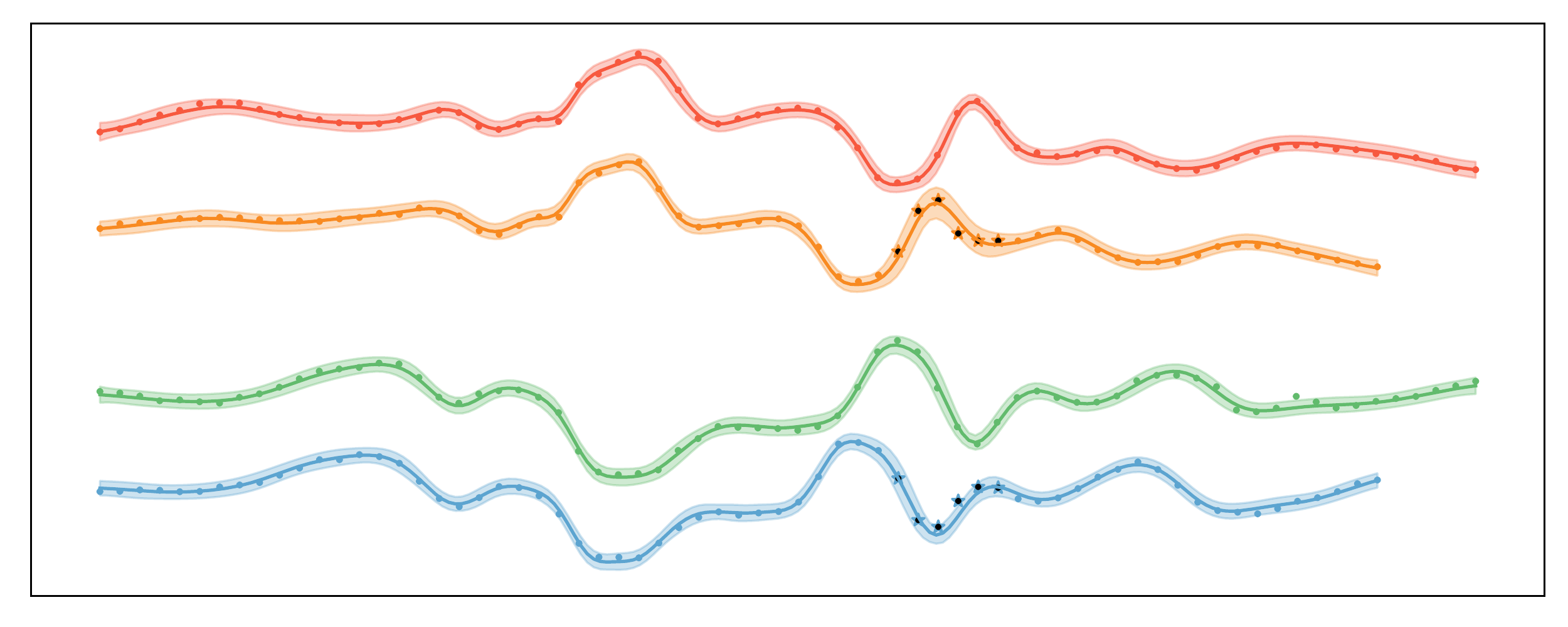}
     \caption{AMTGP missing data examples}
     \label{fig:face_exp_1_data_fit_amtgp}
 \end{subfigure}
\caption{Experiment on the facial landmark data. Red and orange hues are the upper lip coordinates in 2 recordings, green and blue are the corresponding lower lip coordinates. Top row shows MTGP~and the bottom row is AMTGP. \subref{fig:face_exp_1_data_fit_mtgp}~and~\subref{fig:face_exp_1_data_fit_amtgp} show one lower and one upper lip point for each recording. In contrast to AMTGP, the predictions under MTGP~\subref{fig:face_exp_1_data_fit_mtgp} are out of phase with the missing data.}
\label{fig:face_exp1}
\end{figure*}

\begin{table*}[t]
\centering
\caption{Results on Real Datasets} \vspace{-0.25cm}
\begin{adjustbox}{max width=0.65\textwidth}
\begin{tabular}{rrrrrrr}%
\toprule
 & \multicolumn{2}{c}{Facial Expressions}  & \multicolumn{2}{c}{Heartbeat Sounds} \\
\midrule
 & \multicolumn{1}{c}{Train} & \multicolumn{1}{c}{Test} & \multicolumn{1}{c}{Train} & \multicolumn{1}{c}{Test}  \\

\midrule

MTGP (SMSE) & 0.0027 $\pm$ 0.0001 & 1.8327 $\pm$ 2.9569 & 0.0329 $\pm$ 0.0038 & 0.0544 $\pm$ 0.0103 & \\

GP-LVA (SMSE) & 0.0216 $\pm$ 0.0021 & 0.5166 $\pm$ 0.4626 & -\phantom{000000} & -\phantom{000000}\\

\textbf{M-AMTGP}  (SMSE)  & 0.0018 $\pm$ 0.0001 & 0.8172 $\pm$ 1.6459  & 0.0138 $\pm$ 0.0005 & 0.0284 $\pm$ 0.0138 \\

\textbf{AMTGP}  (SMSE)  & 0.0059 $\pm$ 0.0002 & 0.6686 $\pm$ 1.1964 & 0.0181 $\pm$ 0.0014 & 0.0276 $\pm$ 0.0046 \\

\midrule

MTGP (SNLP)  &-3469.7 $\pm$ 30.0 & 89.5 $\pm$ 228.5 & -1674.0 $\pm$ 41.3 & -378.4 $\pm$ 25.1 \\

GP-LVA (SNLP) & -2472.2 $\pm$ 57.8 & 86.3 $\pm$ 119.9 & -\phantom{0000} & -\phantom{0000}\\

\textbf{M-AMTGP} (SNLP)  & -3803.6 $\pm$ 46.0 & -44.5 $\pm$ \phantom{0}35.4 & -2231.1 $\pm$ 21.2 & -489.1 $\pm$ 65.9 \\

\textbf{AMTGP}  (SNLP)  & -2910.4 $\pm$ 48.4 & -56.4 $\pm$ \phantom{0}42.3 & -2006.6 $\pm$ 36.5 & -466.5 $\pm$ 21.6 \\

\bottomrule
\end{tabular}
\label{table:real_data_results}
\end{adjustbox}
\end{table*}

\paragraph{Synthetic Data}
We generate synthetic data by taking two 1-D functions and applying five random monotonic warps to each, adding i.i.d. Gaussian noise, to produce ten misaligned tasks.
Missing data prediction performance is compared across the tasks in the three scenarios S1 - S3; %
$20\%$ of the full data were removed and an SE kernel used. 
The results are summarised in Table~\ref{table:synthetic_results} and Fig.~\ref{fig:synth_comparison}.
The latent variable posterior distribution (Figs.~\ref{fig:synthetic_latent_mtgp},\subref{fig:synthetic_latent_amtgp}) shows that AMTGP and GP-LVA correctly identify the two underlying groups of tasks and that MTGP is unable to detect correlations between misaligned versions of the same task. 
The unaligned MTGP result~(Fig.~\ref{fig:synthetic_data_fit_mtgp}) does not share information correctly and over-fits, resulting in large error bars and poor test performance. Whilst the \MAPAMTGP~aligns the data and improves the mean, the point estimate of the warp is overconfident. With the full marginalisation of the warps, AMTGP is able to both align correctly and model the uncertainty accurately (Fig.~\ref{fig:synthetic_data_fit_amtgp}) resulting in improved performance for both SMSE and SNLP. Notably, in the scenarios with missing segments, S2 and S3, GP-LVA has very poor uncertainty estimation, confirming the detrimental effect of pseudo-observations on missing data reconstruction.

\paragraph{Facial Expressions}
We also test our method on a dataset of dynamic emotional facial expressions RAVDESS \citep{CC,livingstone2018ryerson}. This dataset contains recordings of people saying a short phrase with different emotions.
We use mouth landmark coordinate sequences extracted from the data, share warping functions across all coordinates from each recording, and use a Mat\'{e}rn5/2 kernel.
We use two instances of the same phrase by the same person and ten mouth coordinates.
Scenario S2 with $10\%$ signal removal is employed on one instance and the other left intact.
Both models are able to group lower and upper lip coordinates in each recording, but AMTGP also detects the similarity across recording instances, resulting in only two final clusters (Fig.~\ref{fig:face_exp_1_latent_amtgp}). The missing data prediction of AMTGP is influenced by the behaviour of the other observed instance (Fig.~\ref{fig:face_exp_1_data_fit_amtgp}), while MTGP is unable to use this information resulting in phase errors (Fig.~\ref{fig:face_exp_1_data_fit_mtgp}).

\paragraph{Heartbeat Sounds}
We also consider sequences of heartbeat sounds recorded by a digital stethoscope~\citep{pascal-chsc-2011}. A normal heart sound has a clear  ``lub dub, lub dub'' pattern that varies temporally depending on the age, health, and state of the subject.
The models are tested in scenario S1 with $20\%$ missing data, a Mat\'{e}rn 5/2 kernel is used, and the results are presented in Table~\ref{table:real_data_results}. 
As indicated by the latent space posterior (figures provided in the supplement), AMTGP uncovers correlations within groups of ``lubs'' and ``dubs'', resulting in better predictive performance.

\section{Conclusion and Limitations} \label{sec:colclusions}
AMTGP performs multi-task learning under GP priors that address the problem of temporal noise or warping in time-series data; it extends existing work on multi-task GPs to the case of warped inputs. 
We derive the variational bound, leveraging SVI and path-wise sampling for efficient fully Bayesian inference. 
We provide multiple examples to confirm the intuition that temporal alignment can and should be treated as an integral part of a multi-task model. 
We show, that while modelling uncertainty in the warps is not critical for alignment, it is beneficial when making predictions far from existing data, when MAP can be over-confident.
Monotonic warps are natural for time-series, however, other domains, \emph{e.g.}~spatial or image, other transformations may be more appropriate, \emph{e.g.}~rotations or translation. Our approach could be extended to infer the parameters of some other transformation function. %
Whilst modelling the warps adds a computational overhead, our efficient path-wise approach ensures a linear scaling in $N$. In practice, differentiating through the ODE solver is currently the main limiting factor. That said, the lack of Kronecker form also introduces extra complexity linearly in $J$. Those limitations will be considered in future work.

\subsubsection*{Acknowledgements}
This work was supported by the Swedish Foundation for Strategic Research, Grant No. RIT15-0107 (EACare). Authors would like to thank Akshaya Thippur for his feedback on the manuscript.

\section*{References}

\bibliography{references}

\newpage

\onecolumn
\aistatstitle{Supplementary Materials}
\appendix
\section{PGM}
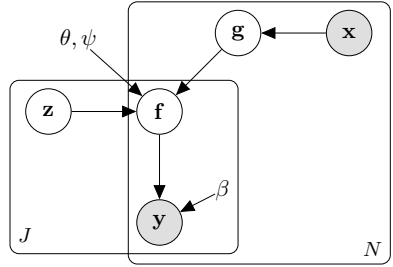
\begin{figure}[ht]
\centering
  \resizebox{0.3\textwidth}{!}{%

\begin{tikzpicture} [scale=1]

\node[obs] (y) {$\mathbf{y}$};

\node[latent, above=of y] (f) {$\mathbf{f}$};
\node[latent, above right=of f] (g) {$\mathbf{g}$};
\node[const, right=of y, xshift=-0.5cm, yshift=0.5cm] (beta) {$\beta$};
\node[const, above left=of f] (theta) {$\theta, \psi$};
\node[latent, left=of f] (z) {$\mathbf{z}$};
\node[obs, right=of g] (x) {$\mathbf{x}$};

\edge {g} {f};
\edge {f} {y};
\edge {z} {f};
\edge {x} {g};
\edge {beta} {y};
\edge {theta} {f};

{\tikzset{plate caption/.append style={below=-8pt of #1.south west}}
\plate {} {(z)(f)(y)(beta)} {$J$} ;}
{\tikzset{plate caption/.append style={below=-2pt of #1.south east}}
\plate {} {(x)(g)(f)(y)(beta)} {$N$} ;}

\end{tikzpicture}   }
  \caption{Generative model of \textbf{AMTGP}. The observed variables are shown in grey. The latent variables $\g$ are the warps that map the inputs $\x$ to the aligned input values. The latent variables $\z$ encode the inter-task correlations. }
  \label{fig:model_pgm}
\end{figure}

\section{Derivations}
\subsection{Sufficient Statistic Assumption}

Following~\cite{titsias2009variational} we can compute the predictive posterior at the new locations $[\Xstar, \Zstar]$ using the augmented joint:

\begin{align}
     p(\Fstar \given \y, \g, \z, \Xstar, \Zstar) &= \int p(\Fstar, \f, \h \given \y, \g, \z, \Xstar, \Zstar) \dd \f \dd \h \\
     &= \int p(\Fstar\given  \f, \h, \g, \z, \Xstar, \Zstar) p(\f \given \h, \y, \g, \z) p(\h\given \y, \g, \z) \dd \f \dd \h \nonumber
\end{align}

Let us assume that $\h$ is a sufficient statistic for $\f$, meaning:
$\Fstar\perp\f|\h  $ or $p(\Fstar \given \h, \f) = p(\Fstar \given \h)$. Thus we have,
$p(\Fstar\given  \f, \h, \g, \z, \Xstar, \Zstar) = p(\Fstar \given \h, \g, \z, \Xstar, \Zstar)$.

Due to of the assumption and the fact that $\y$ is a noisy version of $\f$, it follows that

\begin{align}
   p(\Fstar \given \h, \y, \g, \z) 
    &= \frac{\int p(\Fstar, \f, \h, \y \given \g, \z) \dd \f}{\int p(\Fstar, \f, \h, \y \given \g, \z) \dd \f \dd \Fstar} \nonumber\\
   &= \frac{\int p(\y \given \f) p(\Fstar, \f, \h \given \g, \z) \dd \f}{\int p(\y \given \f) p(\Fstar, \f, \h \given \g, \z) \dd \f \dd \Fstar} \\
   & = \frac{\int p(\y \given \f) p(\Fstar \given \h, \g, \z) p(\f \given \h, \g, \z) p(\h \given \g, \z) \dd \f}{\int p(\y \given \f) p(\Fstar \given \h, \g, \z) p(\f \given \h, \g, \z) p(\h \given \g, \z) \dd \f \dd \Fstar} \nonumber\\
   &= \frac{p(\y \given \h, \g, \z) p(\Fstar, \h \given \g, \z) }{ p(\y \given \h, \g, \z) p(\h \given \g, \z)} = p(\Fstar \given \h, \g, \z) \nonumber
\end{align}

Hence, $p(\f \given \h, \y, \g, \z) = p(\f \given \h, \g, \z)$.

\subsection{$\mathcal{L}_2$ Bound}

 $\mathcal{L}_2$ is defined as $\mathcal{L}_2 = \int q(\h) p(\f  \given  \h, \g, \z) \log{ p(\y \given \f) }\dd \f \dd \h$, where ${q(\h) := \mathcal{N}(\h \given \m, \S_)}$.  Marginalising out $\f$  we get

\begin{align}
    \mathcal{L}_2 &= \int q(\h) p(\f  \given  \h, \g, \z) \log{ p(\y \given \f) }\dd \f \dd \h \\
    &= \EX_{q(\h)}\Big[ \int p(\f  \given  \h, \g, \z) \log{ p(\y \given \f) }  \dd \f \Big]\\
    &= \EX_{q(\h)}\Big[ -\frac{\beta}{2}(\y - \boldsymbol{\mu})^{T}(\y - \boldsymbol{\mu}) - \frac{JN}{2}\log{2\pi} 
     - \frac{1}{2}\log{|\beta^{-1}\I|} - \frac{\beta}{2}\mathsf{Tr}[\Sigma]\Big] 
\end{align}
where $\boldsymbol{\mu} := \K_{fh}\K_{hh}^{-1} \h$ and $\Sigma := \K_{ff} - \K_{fh}\K_{hh}^{-1}\K_{hf}$. 

Taking expectations under the distribution $q(\h)$ we arrive at the final result:
\begin{align}
    \mathcal{L}_2 &=\sum^{J}_{j=1}\Big\{\log \mathcal{N}\big(\y_j \given \K_{f_{j}h}\K_{hh}^{-1}\m, \beta^{-1}\I\big) 
     - \frac{1}{2} \mathsf{Tr}[\Lambda_j\S_] - \frac{\beta}{2} \mathsf{Tr}[\Sigma_j] \Big\} 
\end{align}
where the matrices $\Lambda_j := \beta \,\K_{hh}^{-1}\K_{hf_{j}}\K_{f_{j}h}\K_{hh}^{-1}$ and ${\Sigma_j := \K_{f_{j},f_{j}} - \K_{f_{j}h}\K_{hh}^{-1}\K_{hf_{j}}}$.

\subsection{M-AMTGP: MAP estimate of the warps}
For comparison, we include a version of AMTGP where we consider only a point MAP estimate for the monotonic warps following the GP-LVA approach~\citep{kazlauskaite2019gaussian}.

We parameterise each $\g_j$ through a set of auxiliary variables $\U_j\in \mathbb{R}^N$ and enforce monotonicity as follows:
\begin{equation}
     g_{jn} = \sum_{i=1}^n \big[ \mathsf{softmax}(\U_{j}) \big]_i \,,
\end{equation}
under a Gaussian process prior over the values of $\g_j$.
This re-parameterization ensures that the warps are monotonic over the specified range $[0, 1]$. We also add a scale and a shift parameter to map this to the range of the input space.

\subsection{GP-LVA: extension to missing data}
The original GP-LVA \citep{kazlauskaite2019gaussian} was not designed for missing data reconstruction. However, it is possible to use predictive posterior of individual sequences conditioned on both observed data $Y$ and pseudo-observations $S$. While conditioning on $S$ allows for implicit correlation between aligned sequences, it can cause the model severely misbehave. In some scenarios, we found the pseudo-observations to overpower the data, resulting in poor fit to $Y$.

\section{Additional Experiments and Plots}

\begin{figure*}[ht]
\centering
\begin{subfigure}[b]{0.16\textwidth}
     \centering
     \includegraphics[width=\textwidth]{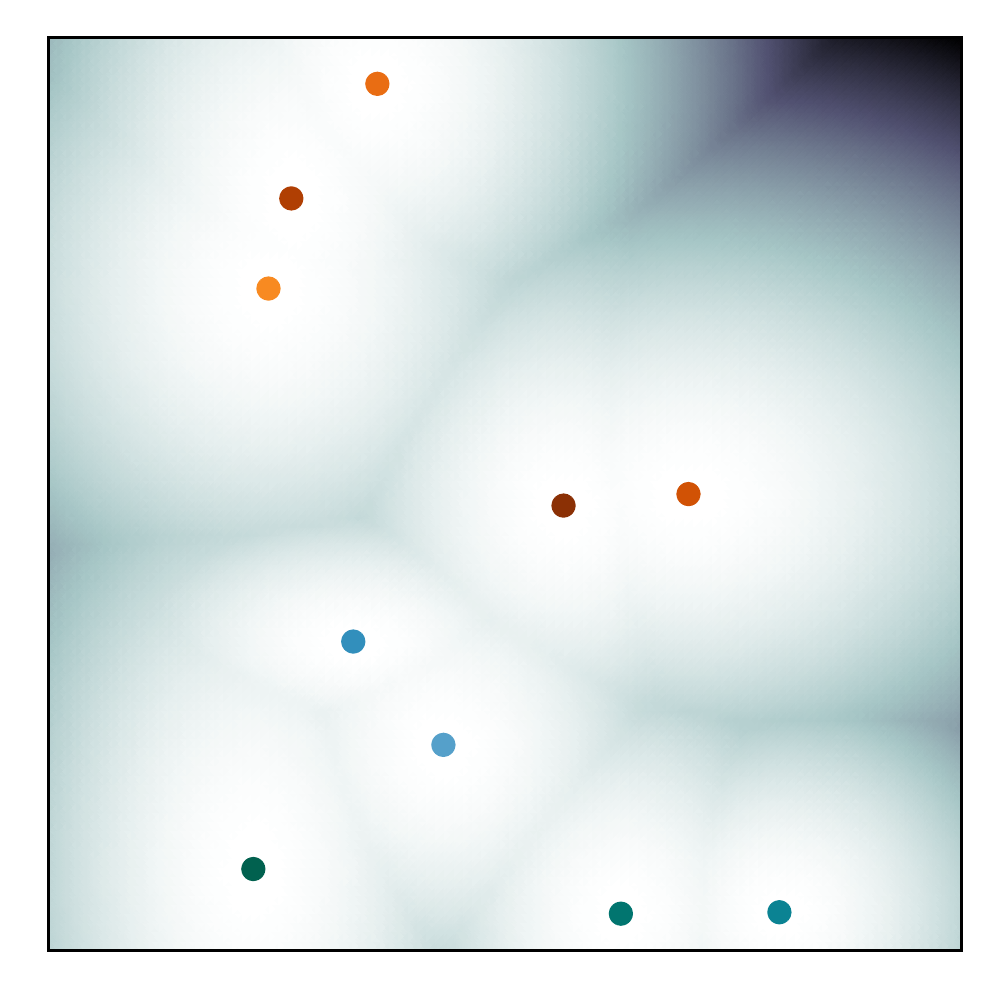}
     \caption{MTGP $\textbf{Z}$}
     \label{fig:hearts_latent_mtgp}
 \end{subfigure}
 \begin{subfigure}[b]{0.4\textwidth}
     \centering
     \includegraphics[width=\textwidth]{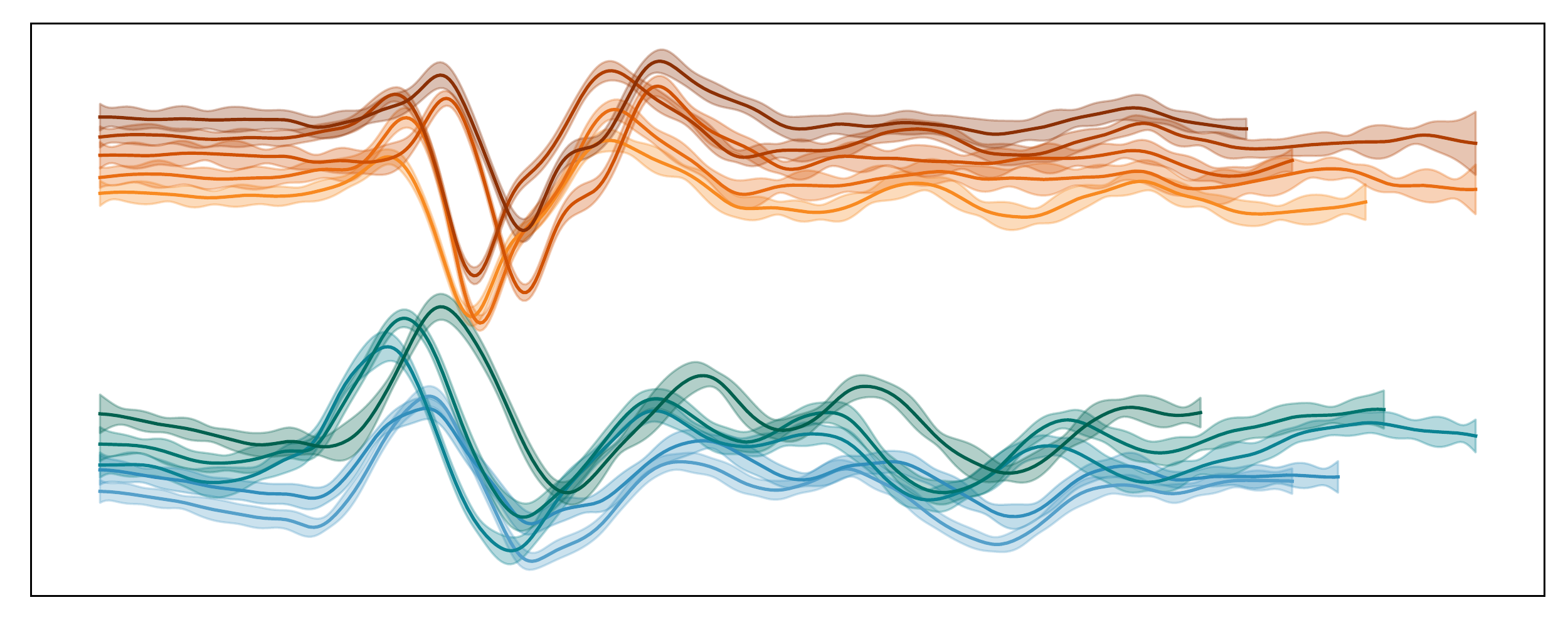}
     \caption{MTGP function posteriors (unaligned)}
     \label{fig:hearts_f_post_mtgp}
 \end{subfigure}
 \begin{subfigure}[b]{0.4\textwidth}
     \centering
     \includegraphics[width=\textwidth]{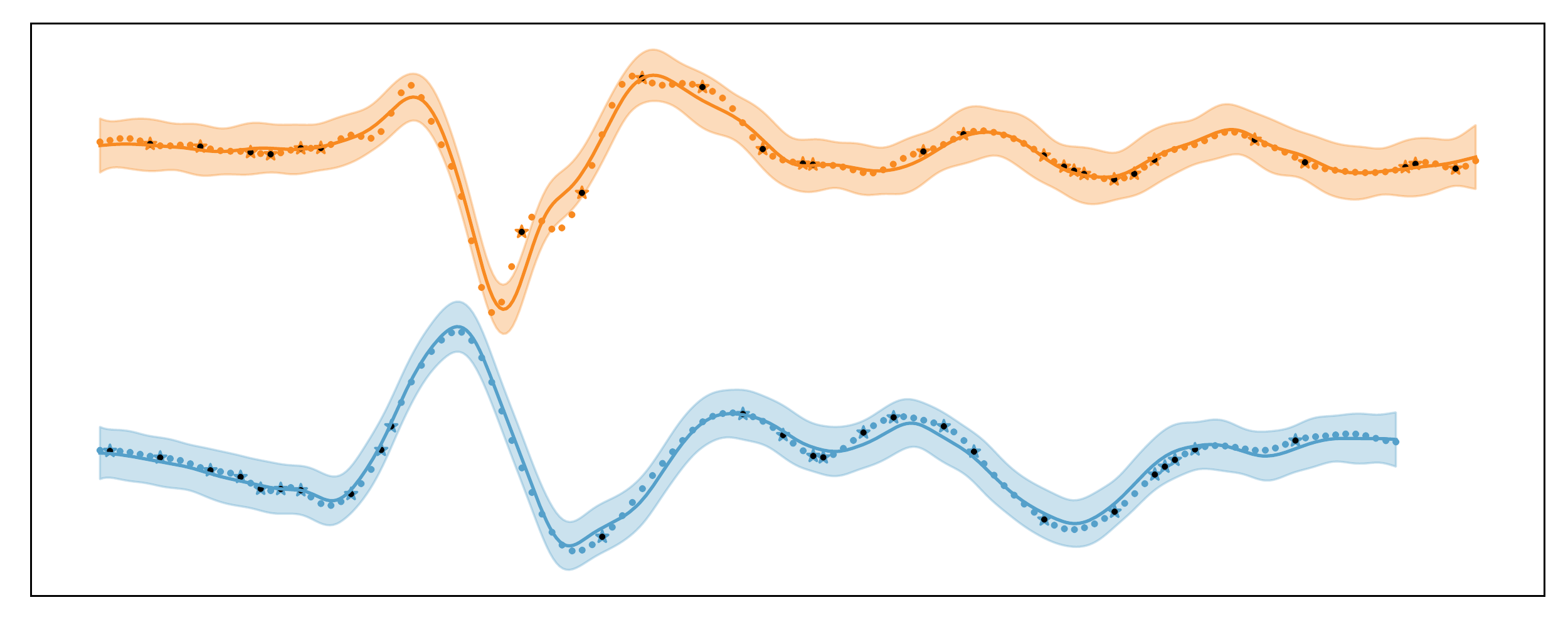}
     \caption{MTGP missing data examples}
     \label{fig:hearts_data_fit_mtgp}
 \end{subfigure}\\
 \begin{subfigure}[b]{0.16\textwidth}
     \centering
     \includegraphics[width=\textwidth]{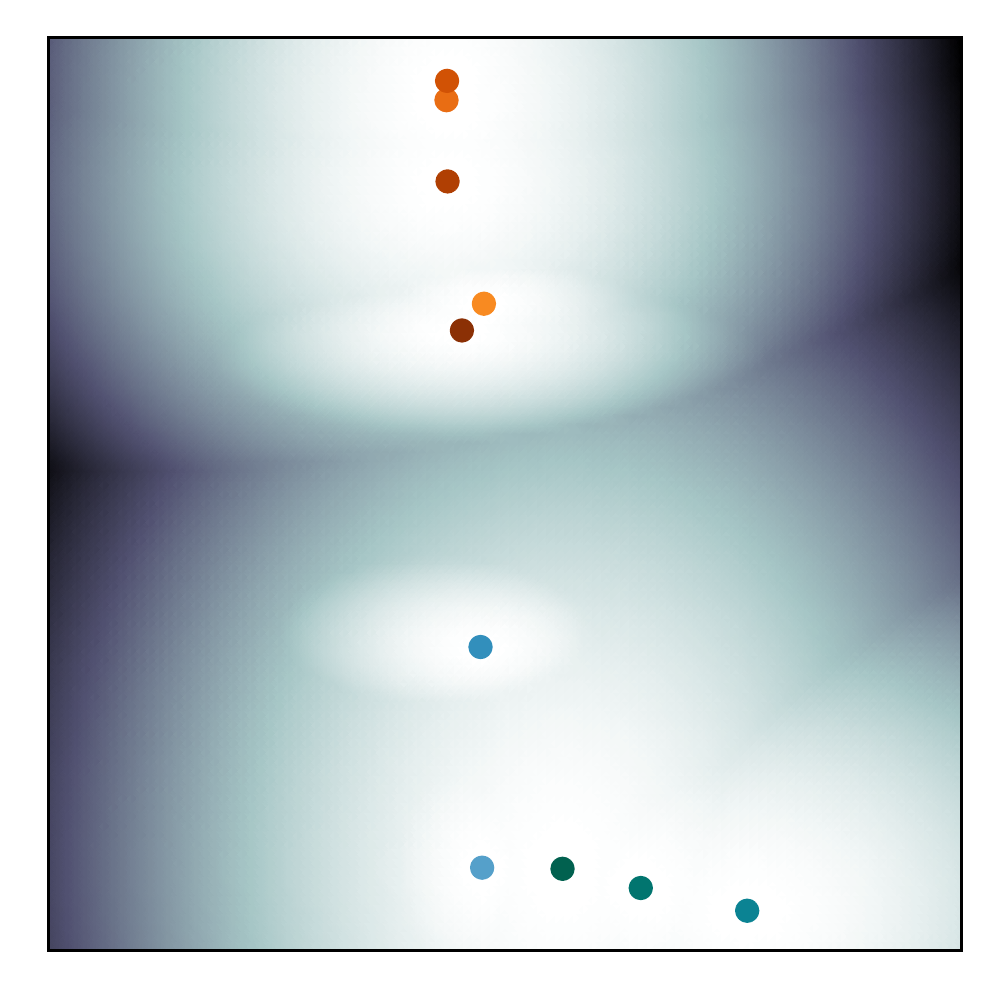}
     \caption{M-AMTGP $\textbf{Z}$}
     \label{fig:hearts_latent_m-amtgp}
 \end{subfigure}
 \begin{subfigure}[b]{0.4\textwidth}
     \centering
     \includegraphics[width=\textwidth]{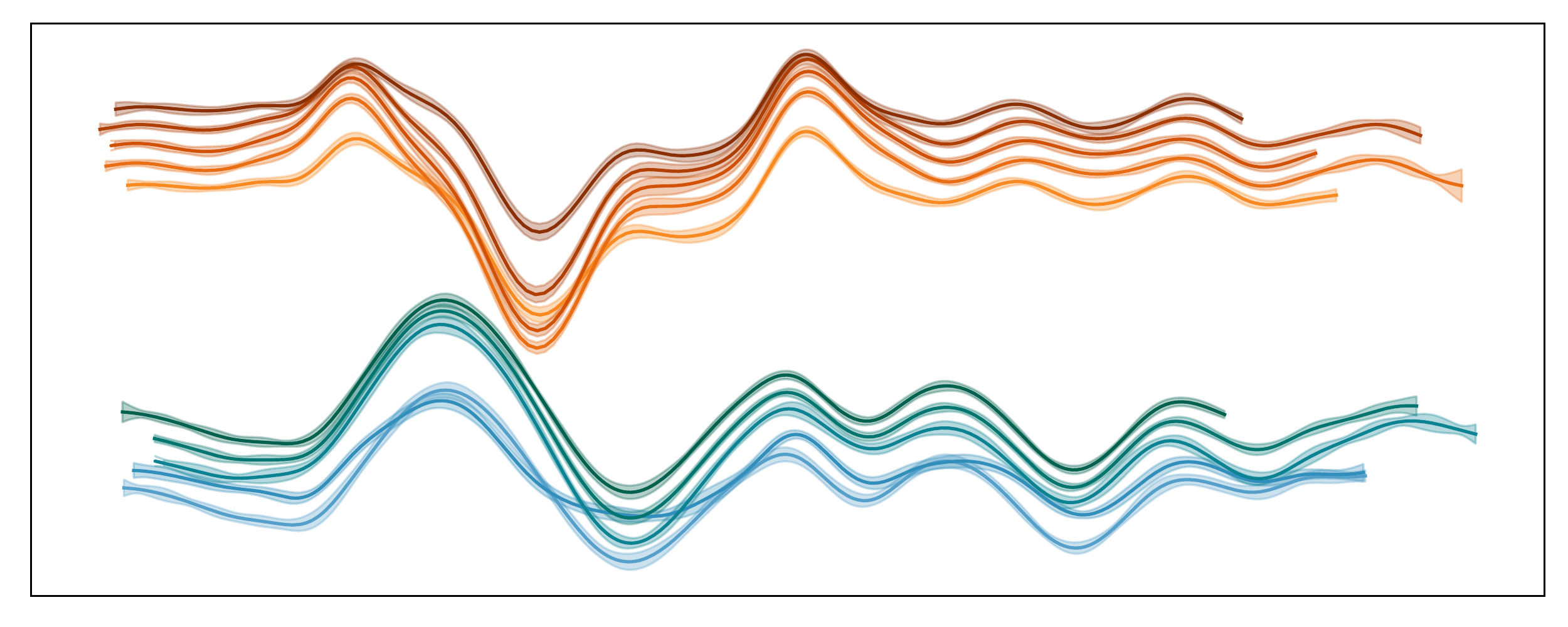}
     \caption{M-AMTGP function posteriors (aligned)}
     \label{fig:hearts_f_post_m-amtgp}
 \end{subfigure}
 \begin{subfigure}[b]{0.4\textwidth}
     \centering
     \includegraphics[width=\textwidth]{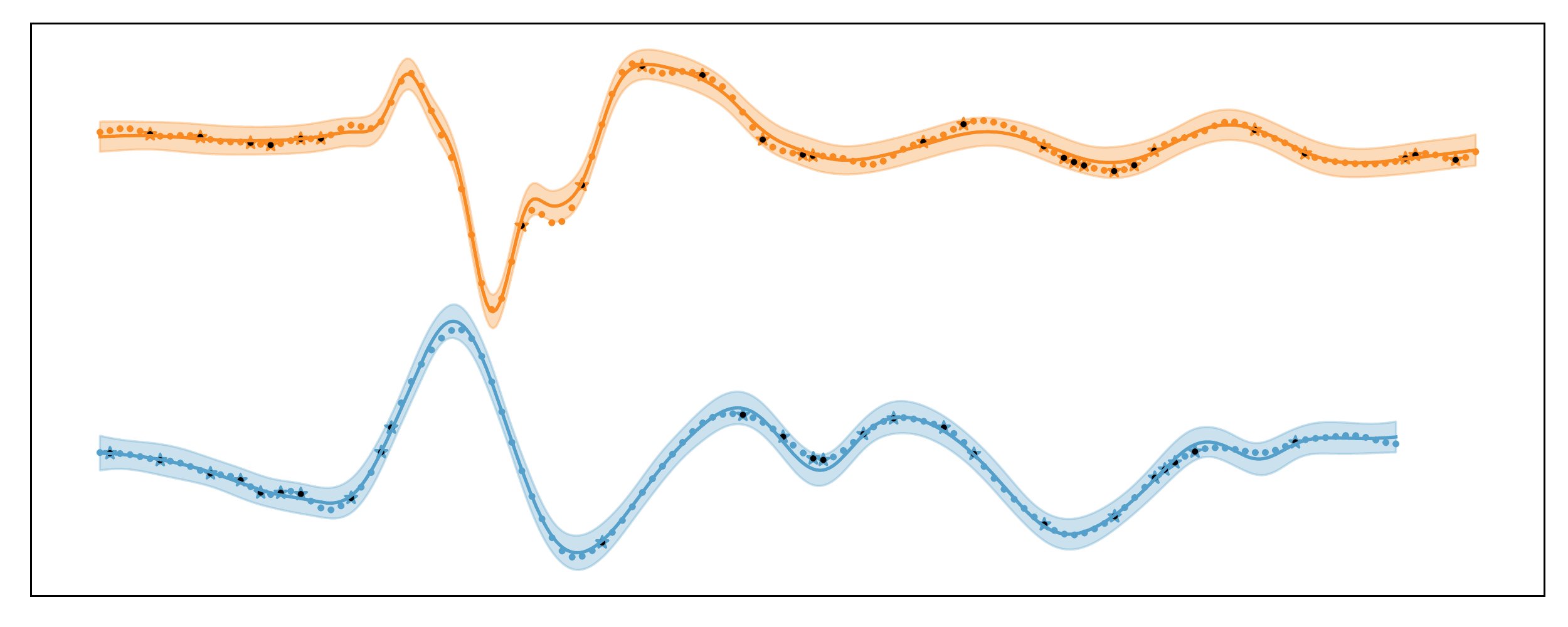}
     \caption{M-AMTGP missing data examples}
     \label{fig:hearts_data_fit_m-amtgp}
 \end{subfigure}\\
 \begin{subfigure}[b]{0.16\textwidth}
     \centering
     \includegraphics[width=\textwidth]{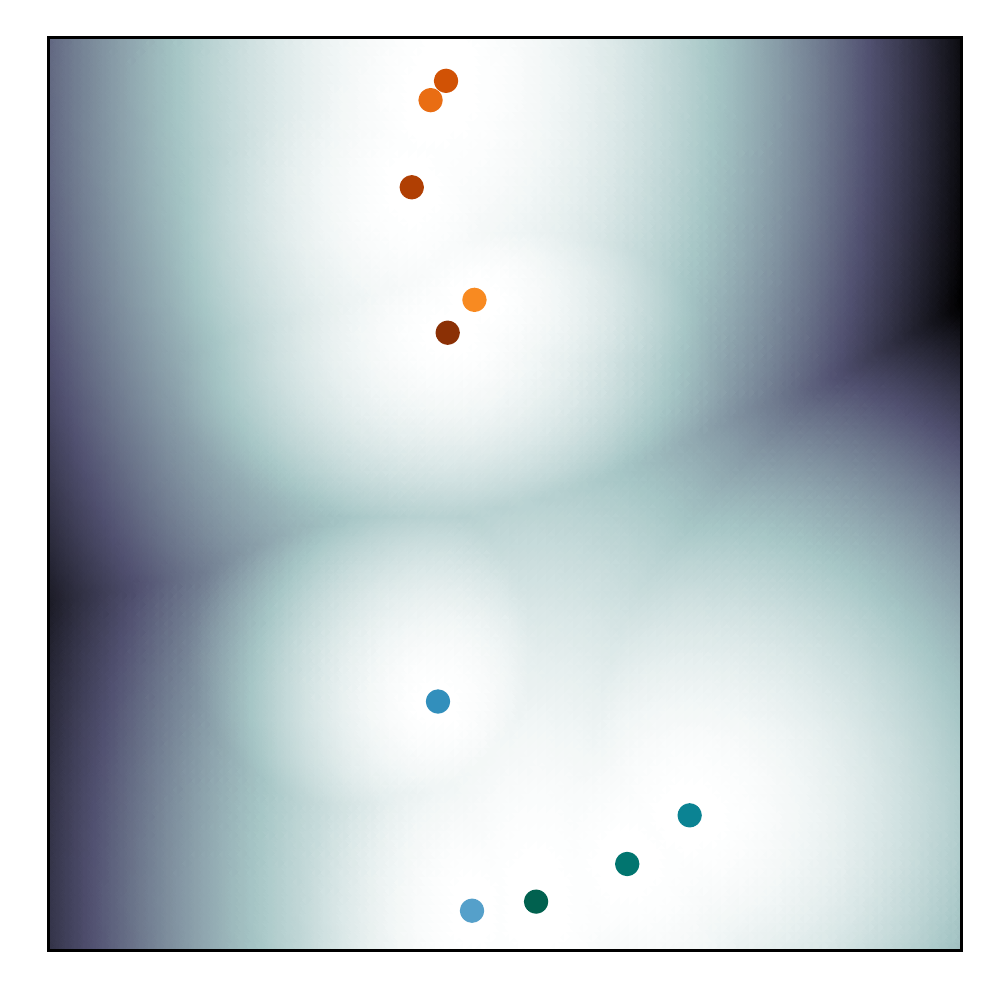}
     \caption{AMTGP $\textbf{Z}$}
     \label{fig:hearts_latent_amtgp}
 \end{subfigure}
 \begin{subfigure}[b]{0.4\textwidth}
     \centering
     \includegraphics[width=\textwidth]{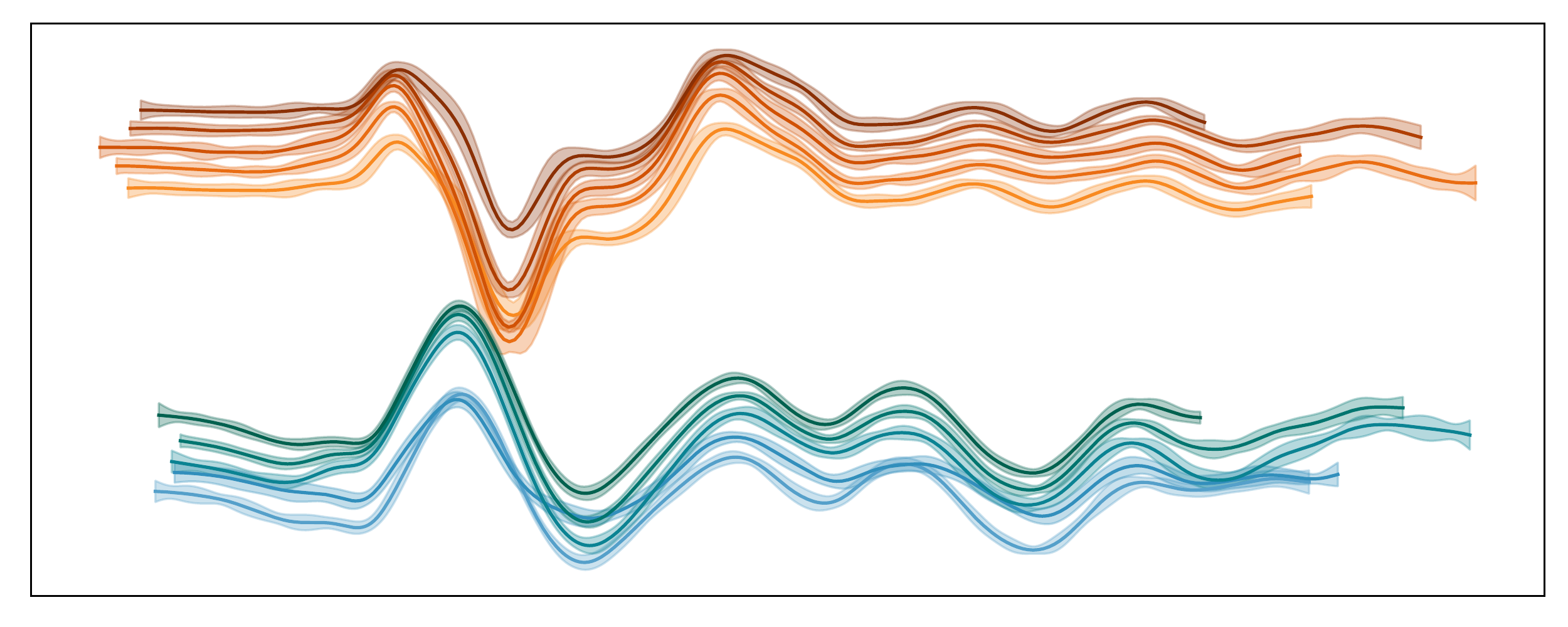}
     \caption{AMTGP function posteriors (aligned)}
     \label{fig:hearts_f_post_amtgp}
 \end{subfigure}
 \begin{subfigure}[b]{0.4\textwidth}
     \centering
     \includegraphics[width=\textwidth]{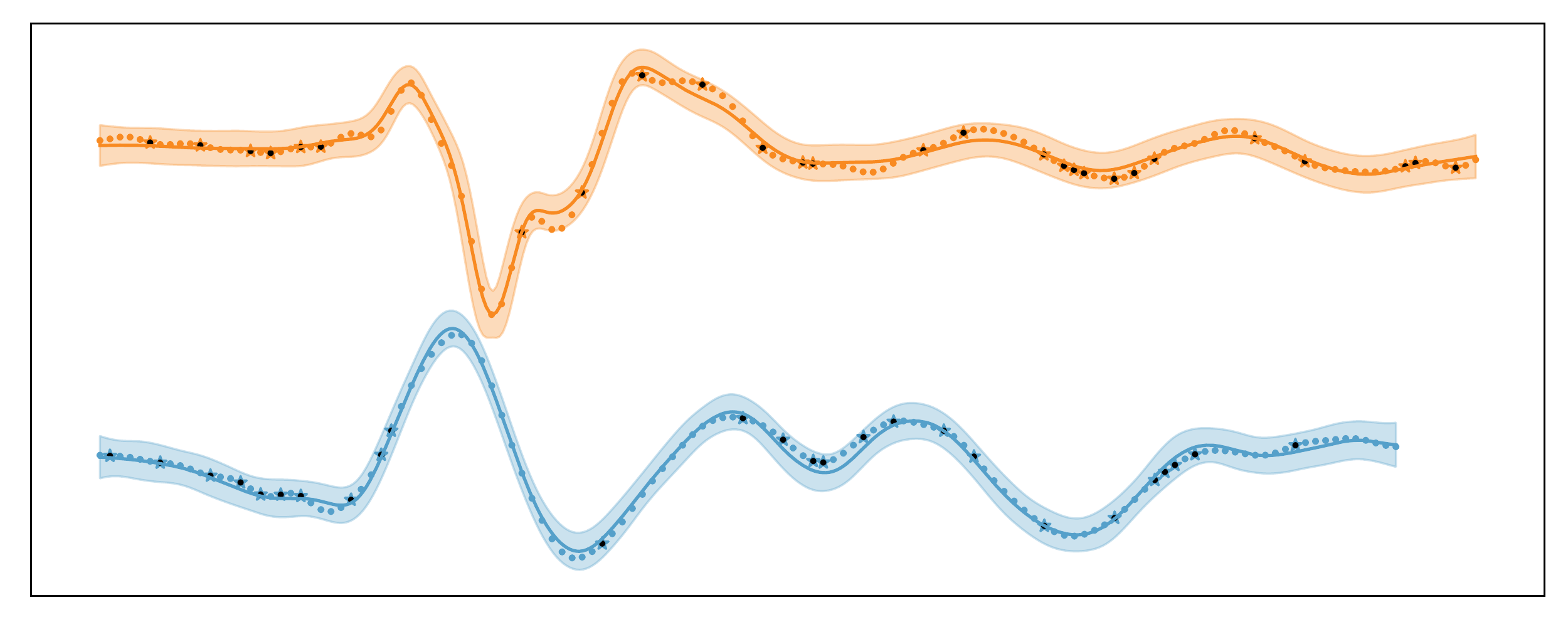}
     \caption{AMTGP missing data examples}
     \label{fig:hearts_data_fit_amtgp}
 \end{subfigure}
\caption{Experiment on the heart beat data. Top row shows MTGP, middle row is M-AMTGP (AMTGP with MAP estimate of warps), bottom row is AMTGP. ``Lubs'' and ``dubs'' are plotted with red and blue respectively. The right column shows one ``lub'' and one ``dub''. $20\%$ of the data is missing at random locations and is shown in black.}
\label{fig:hearts}
\end{figure*}

\subsection{Heartbeat Data}

Figure \ref{fig:hearts} shows results on the Heartbeat data. Details about the data are presented in the main paper. In this experiment, we omit the results for GP-LVA, since we could not achieve a good fit from this model even with no data missing. In particular, GP-LVA aligned pseudo-observations of both "lubs" and "dubs" together in one single category. This resulted in GP-LVA fitting the data $Y$ well only on one category and severely misfitting on another. This clearly shows the downside of pseudo-observations trick.

\subsection{Respiratory Motion Traces}
We perform additional experiments on physiological data \citep{ernst2011compensating}. Specifically, we consider recordings of human liver motion traces. We use 6 markers of liver positions (3 external data and 3 from ultrasound). In this data the misalignment is small and comes from physiological processes.

We perform an experiment where $10\%$ of the data is missing in S3 scenario (continuous segment of data is missing in each task at random locations). Since the data clearly exhibits periodic behavior, we use a sum of a cosine and a squared exponential kernel to model the temporal behavior.

\begin{figure*}[t]
\centering
\begin{subfigure}[b]{0.16\textwidth}
     \centering
     \includegraphics[width=\textwidth]{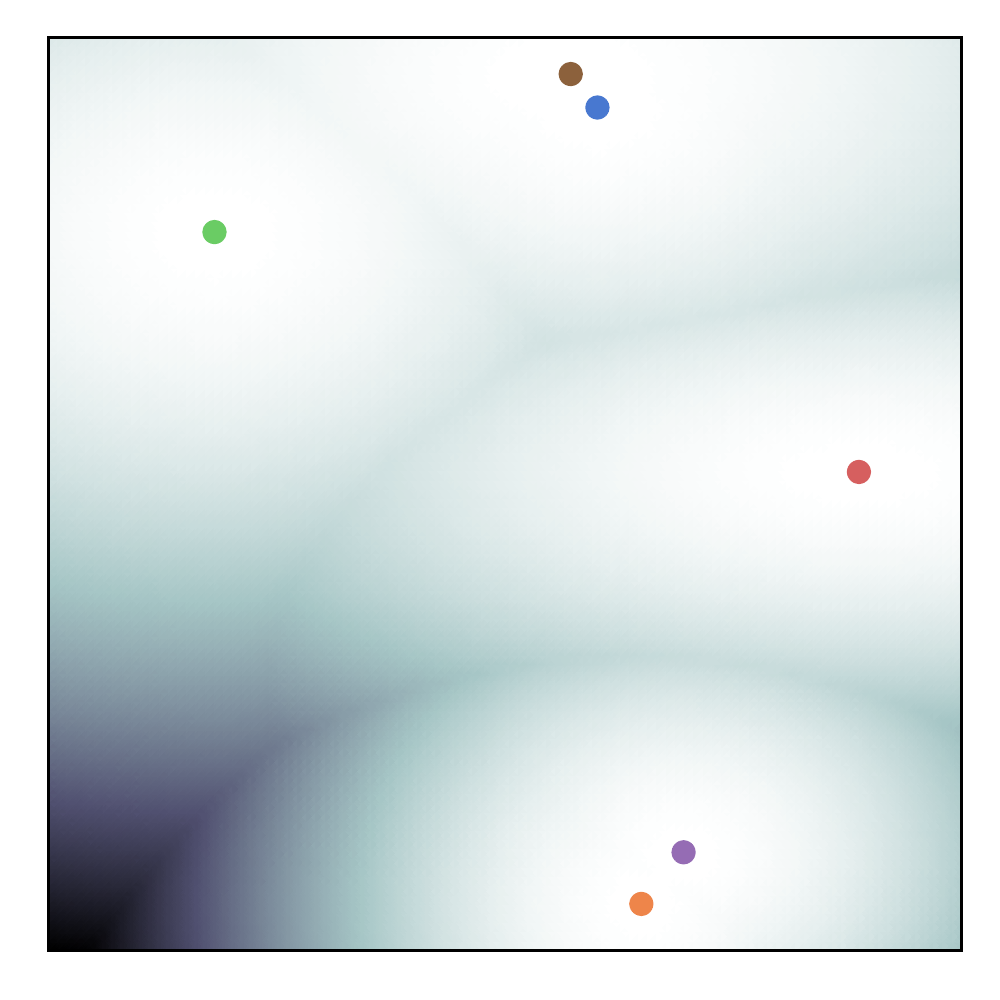}
     \caption{MTGP $\textbf{Z}$}
     \label{fig:liver_latent_mtgp}
 \end{subfigure}
 \begin{subfigure}[b]{0.4\textwidth}
     \centering
     \includegraphics[width=\textwidth]{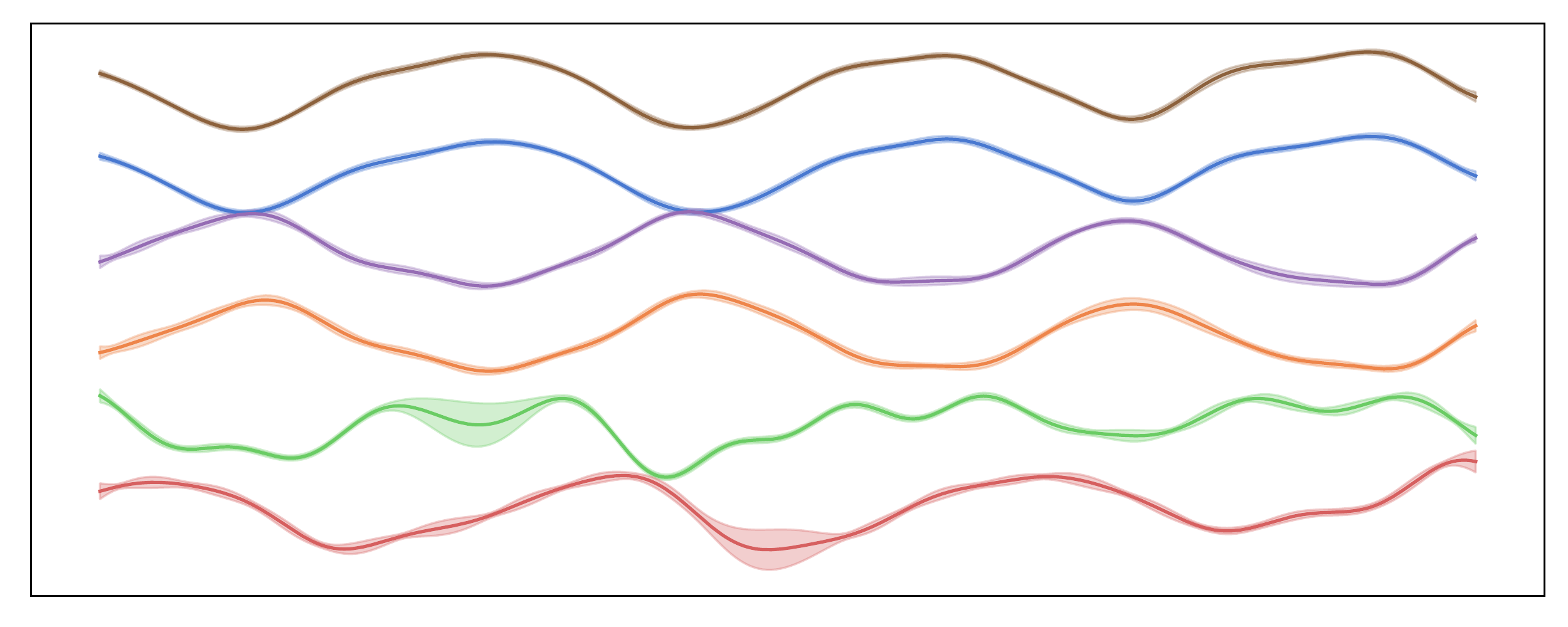}
     \caption{MTGP function posteriors (unaligned)}
     \label{fig:liver_f_post_mtgp}
 \end{subfigure}
 \begin{subfigure}[b]{0.4\textwidth}
     \centering
     \includegraphics[width=\textwidth]{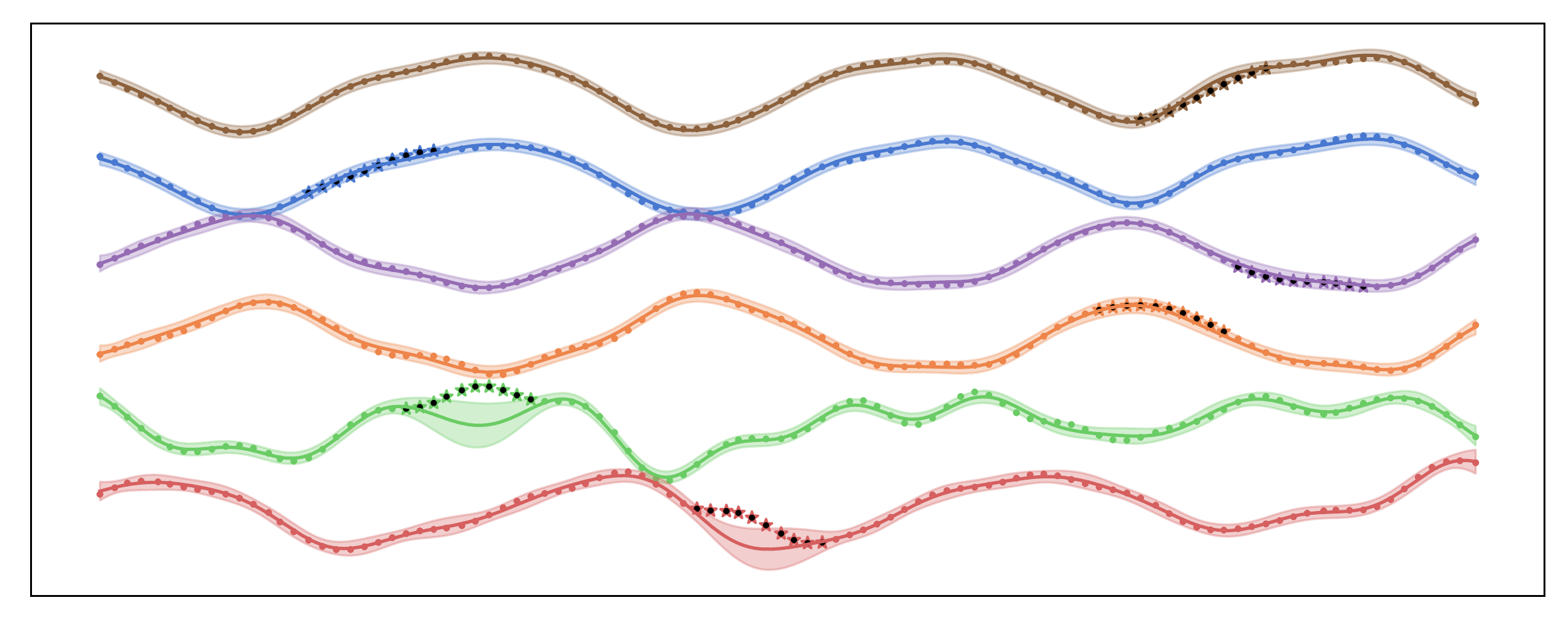}
     \caption{MTGP missing data examples}
     \label{fig:liver_data_fit_mtgp}
 \end{subfigure}\\
 \begin{subfigure}[b]{0.16\textwidth}
     \centering
     \includegraphics[width=\textwidth]{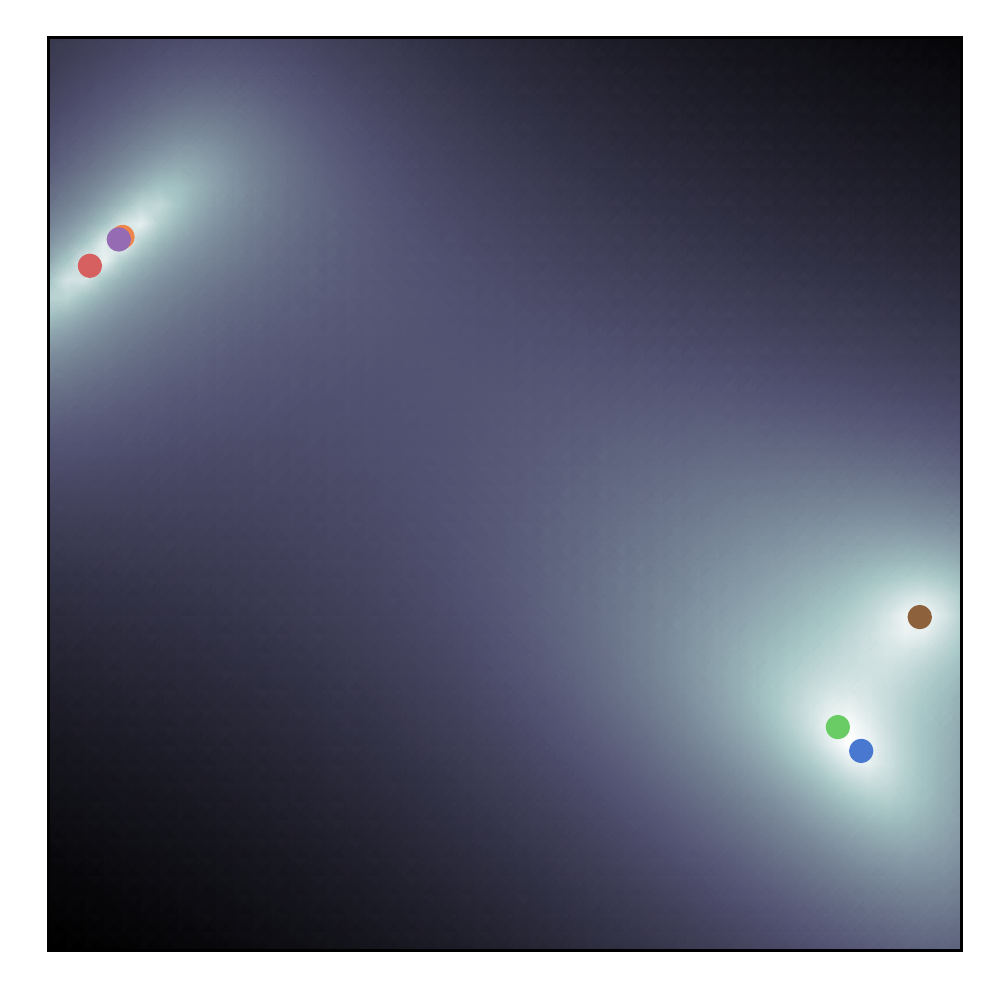}
     \caption{GP-LVA $\textbf{Z}$}
     \label{fig:liver_latent_gplva}
 \end{subfigure}
 \begin{subfigure}[b]{0.4\textwidth}
     \centering
     \includegraphics[width=\textwidth]{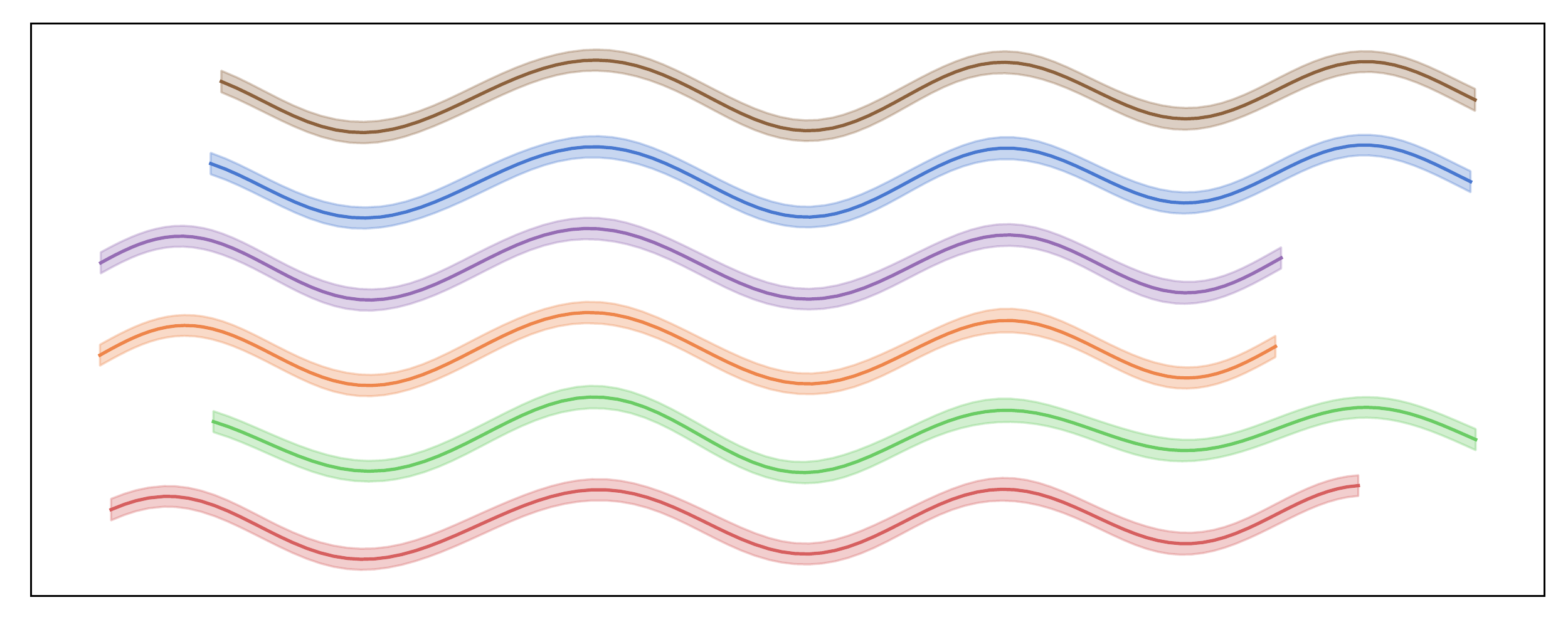}
     \caption{GP-LVA function posteriors (unaligned)}
     \label{fig:liver_f_post_gplva}
 \end{subfigure}
 \begin{subfigure}[b]{0.4\textwidth}
     \centering
     \includegraphics[width=\textwidth]{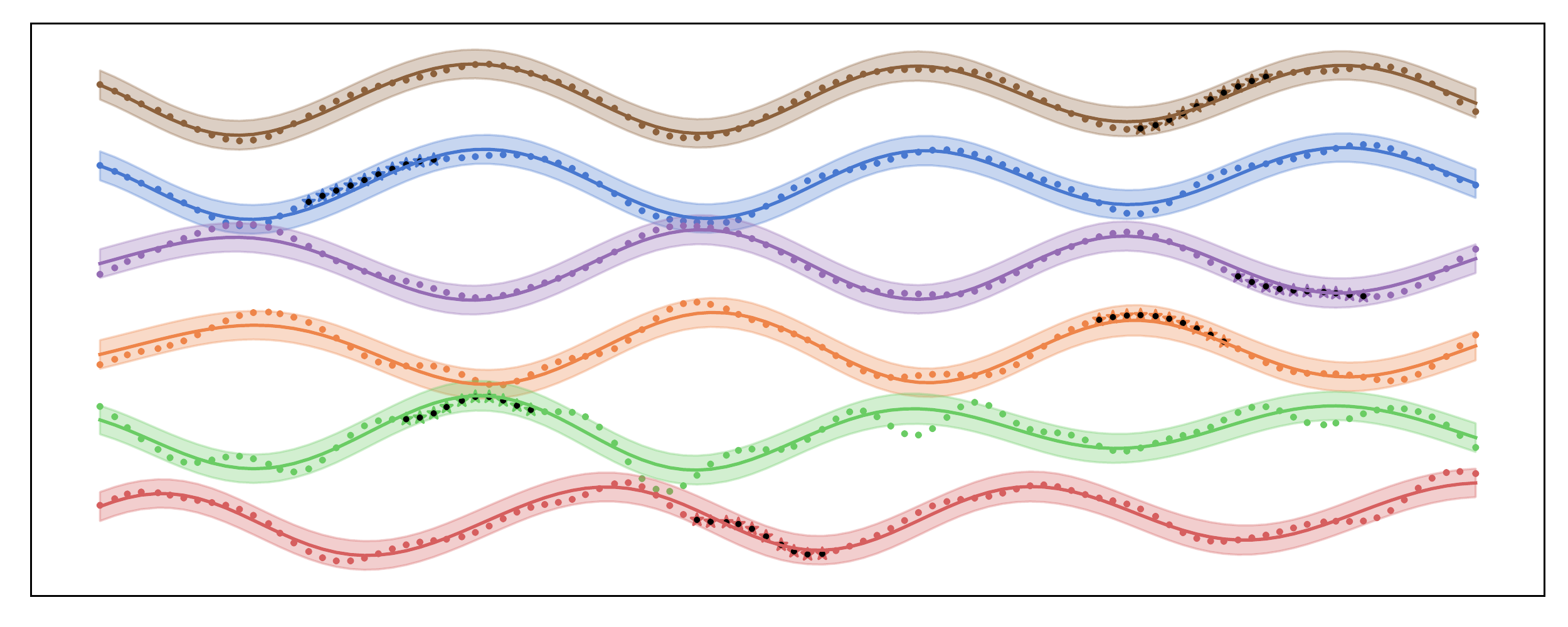}
     \caption{GP-LVA missing data examples}
     \label{fig:liver_data_fit_gplva}
 \end{subfigure}\\
  \begin{subfigure}[b]{0.16\textwidth}
     \centering
     \includegraphics[width=\textwidth]{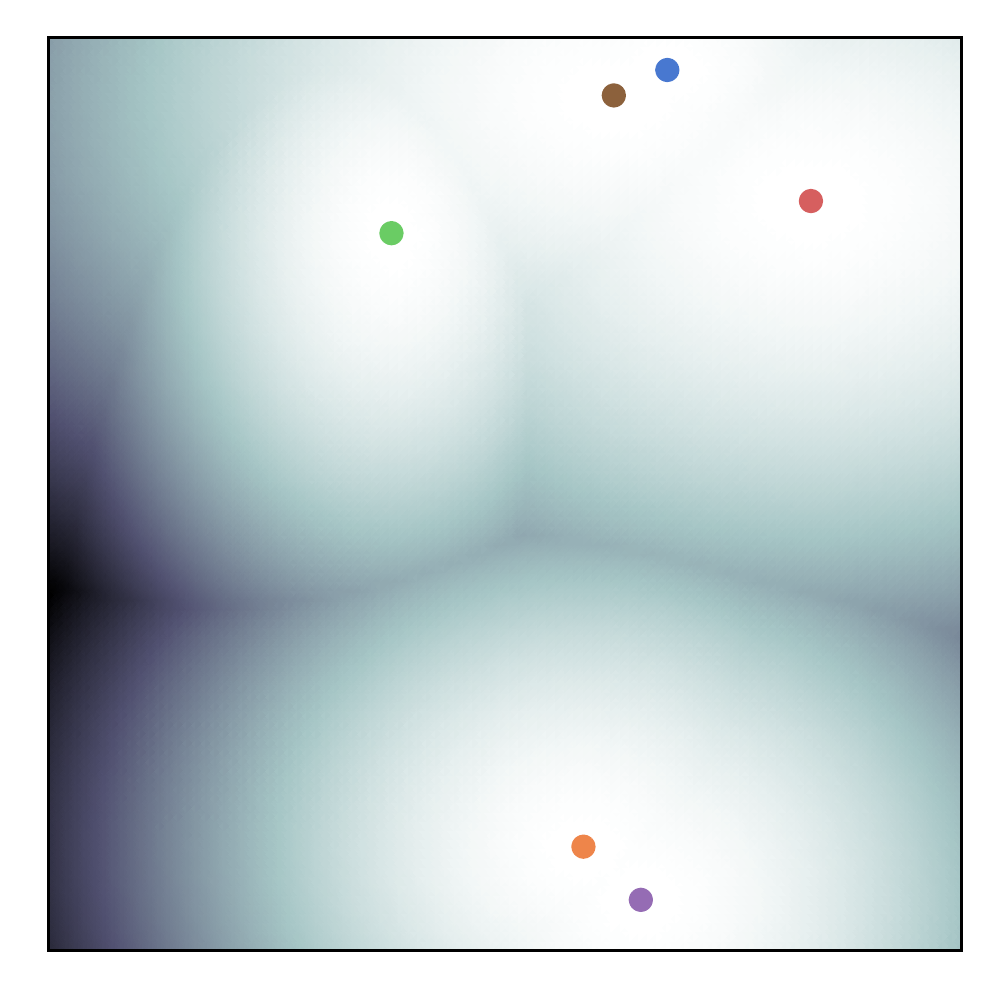}
     \caption{M-AMTGP $\textbf{Z}$}
     \label{fig:liver_latent_m-amtgp}
 \end{subfigure}
 \begin{subfigure}[b]{0.4\textwidth}
     \centering
     \includegraphics[width=\textwidth]{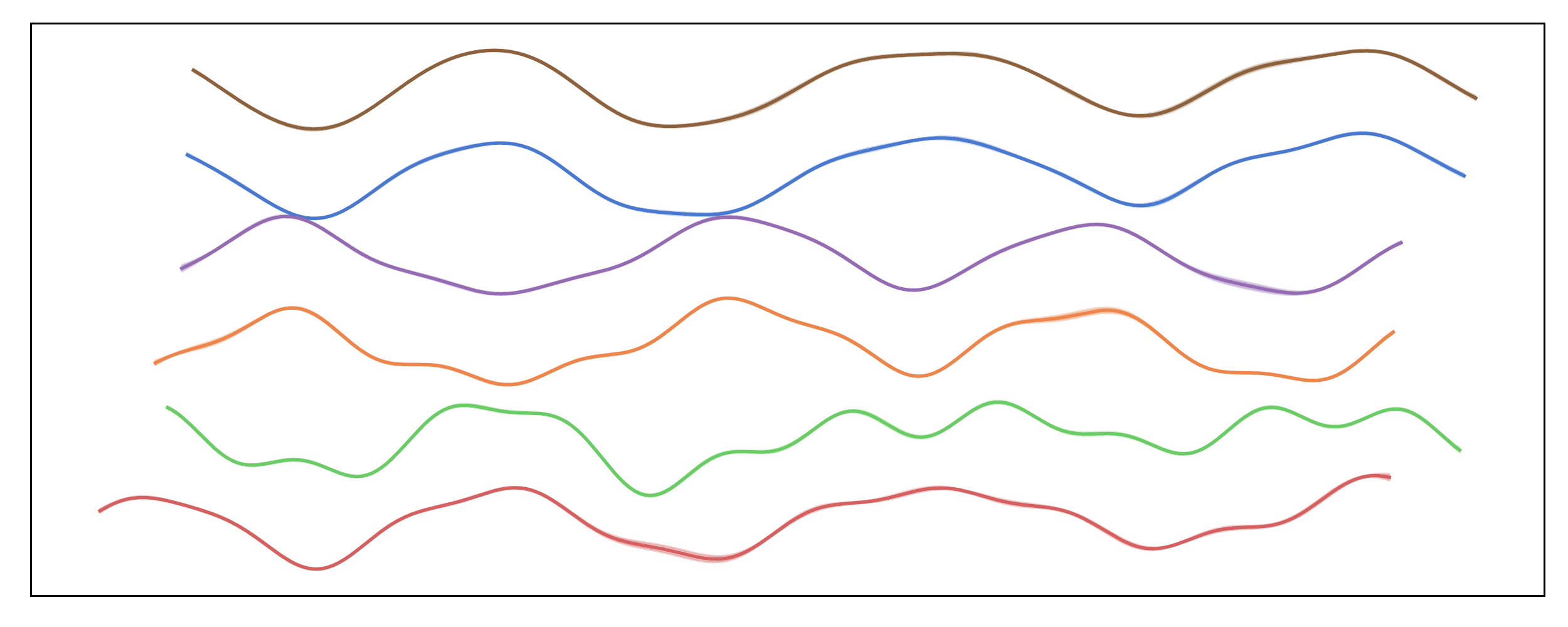}
     \caption{M-AMTGP function posteriors (aligned)}
     \label{fig:liver_f_post_m-amtgp}
 \end{subfigure}
 \begin{subfigure}[b]{0.4\textwidth}
     \centering
     \includegraphics[width=\textwidth]{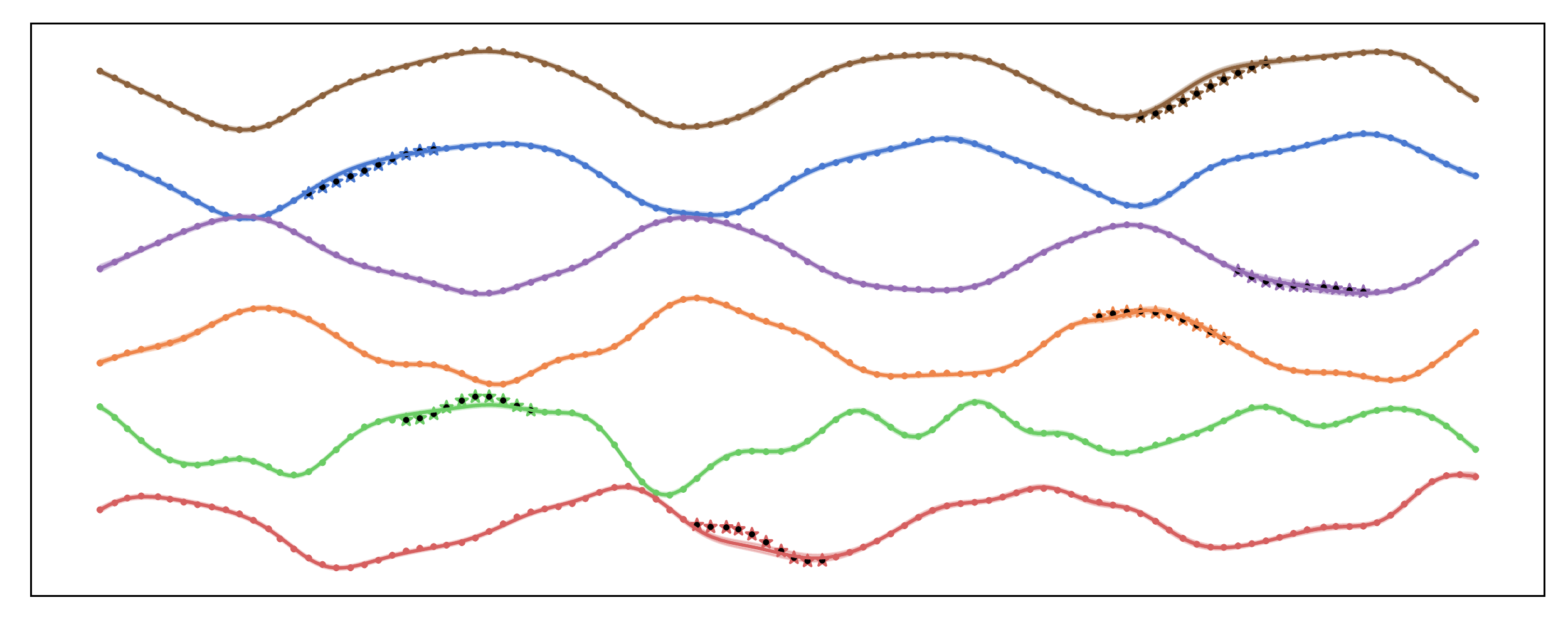}
     \caption{M-AMTGP missing data examples}
     \label{fig:liver_data_fit_m-amtgp}
 \end{subfigure}\\
 \begin{subfigure}[b]{0.16\textwidth}
     \centering
     \includegraphics[width=\textwidth]{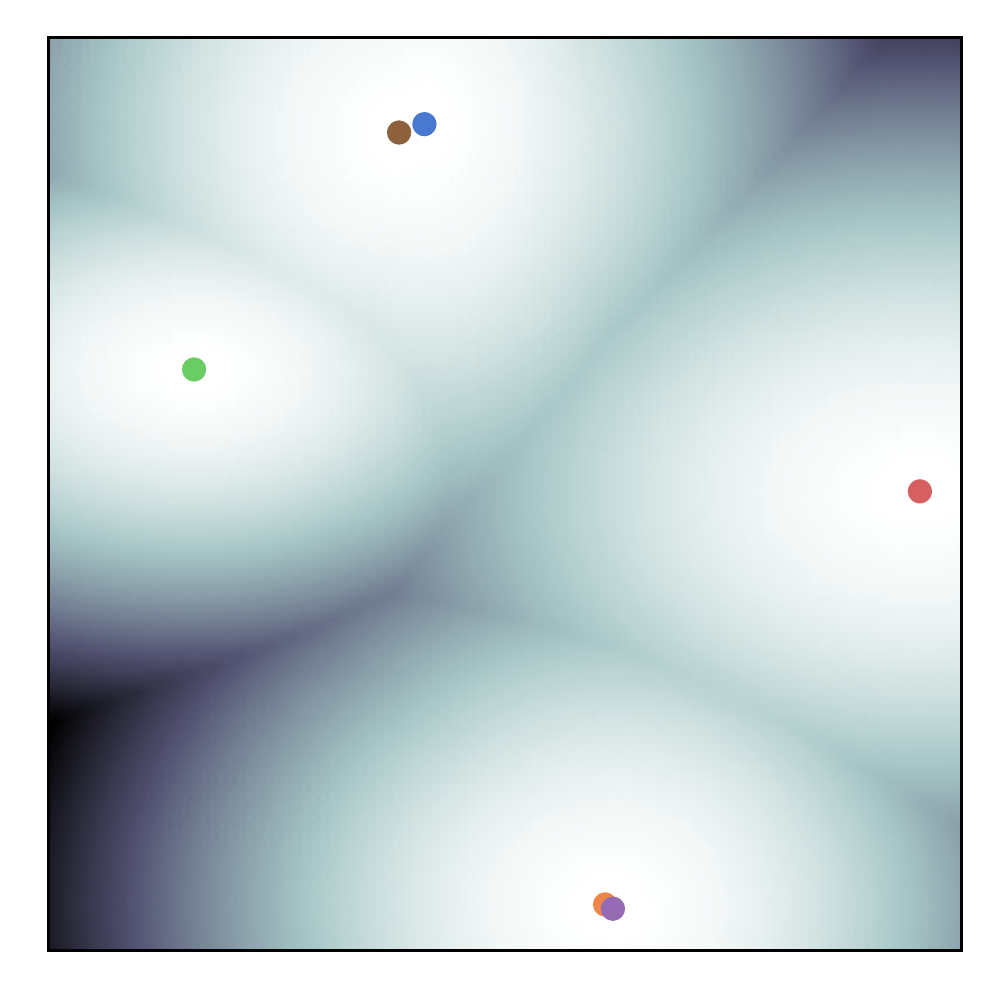}
     \caption{AMTGP $\textbf{Z}$}
     \label{fig:liver_latent_amtgp}
 \end{subfigure}
 \begin{subfigure}[b]{0.4\textwidth}
     \centering
     \includegraphics[width=\textwidth]{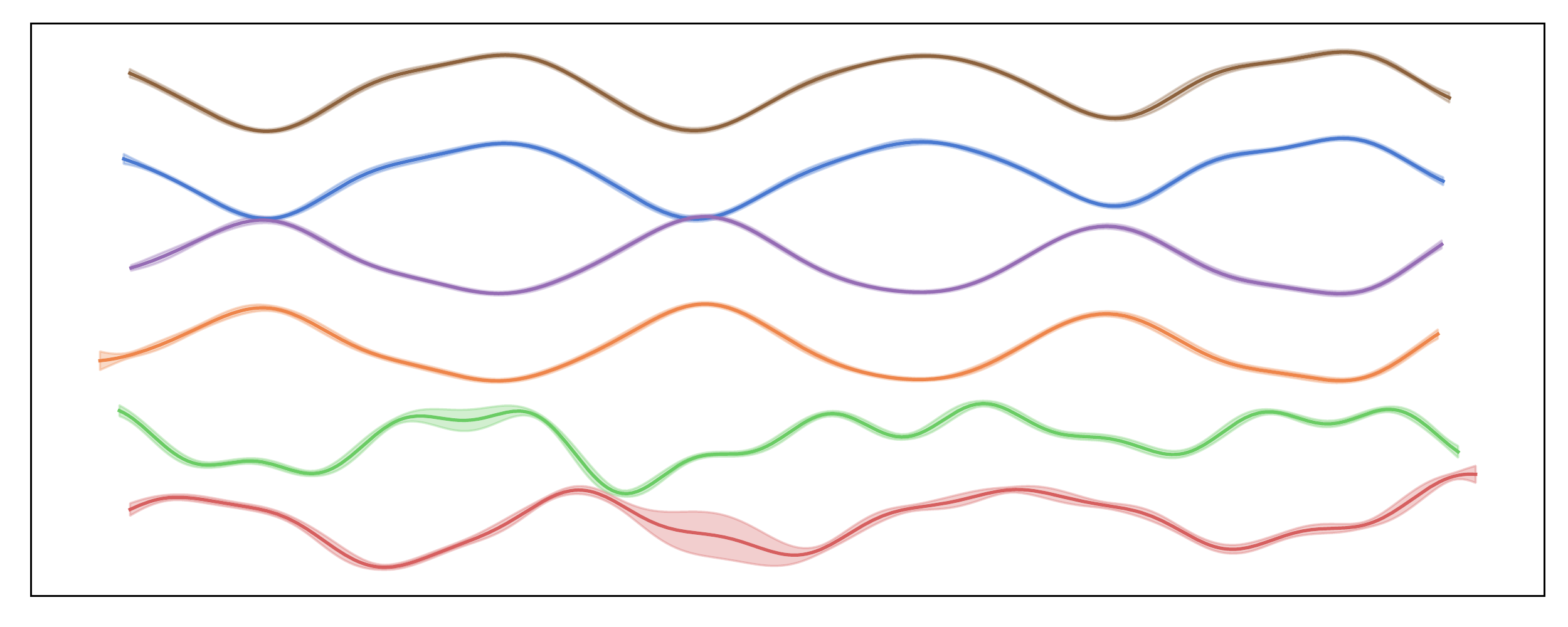}
     \caption{AMTGP function posteriors (aligned)}
     \label{fig:liver_f_post_amtgp}
 \end{subfigure}
 \begin{subfigure}[b]{0.4\textwidth}
     \centering
     \includegraphics[width=\textwidth]{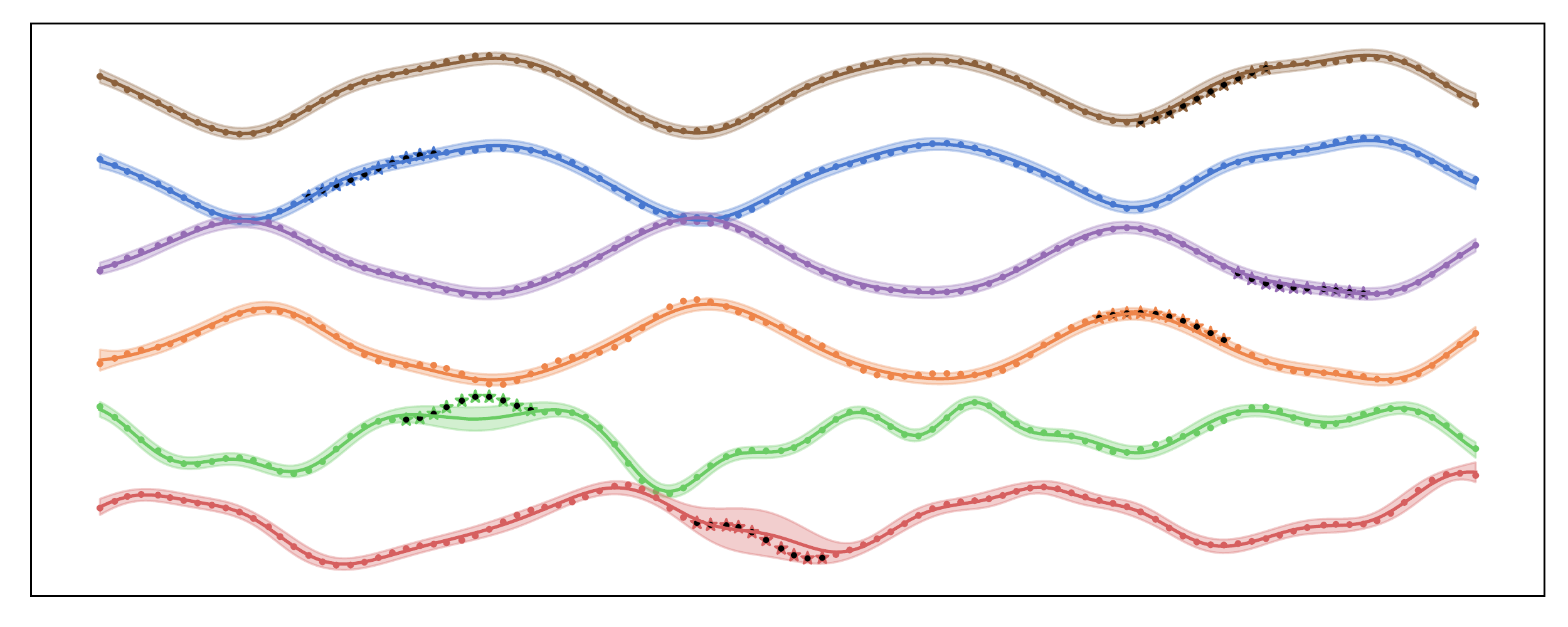}
     \caption{AMTGP missing data examples}
     \label{fig:liver_data_fit_amtgp}
 \end{subfigure}
\caption{Experiment on the liver data. Top row shows MTGP, middle row is M-AMTGP (AMTGP with MAP estimate of warps), bottom row is AMTGP. The right column is plotted with an offset for better visibility. $10\%$ of the data is missing in continuous segments at random locations in each task and is shown in black.}
\label{fig:liver}
\end{figure*}

Results are presented in Table \ref{table:real_data_results_extra} and Figure \ref{fig:liver}. As we can see, all 3 models show similar predictive performance in terms of SMSE. This is due to the fact, that the data has only small misalignment and the correlated tasks exhibit strong periodic behaviour. Importantly, alignments in the proposed AMTGP model and its MAP version, M-AMTGP, do not lead to spurious correlations and negative knowledge transfer between the tasks. The predictive log probability (SNLP) in this experiment once again highlights the effect of Bayesian warp estimation, compared to MAP estimation. M-AMTGP if overconfident in its predictions, while fully Bayesian AMTGP preserves reasonable prediction uncertainty.

 \begin{table*}[h]
\centering
\caption{Results of the additional experiments.}
\begin{adjustbox}{max width=0.75\textwidth}
\begin{tabular}{rrrrrrr}%
\toprule
 & \multicolumn{2}{c}{Respiratory motion traces (liver)} & \multicolumn{2}{c}{Facial Expressions 2} \\
\midrule
 & \multicolumn{1}{c}{Train} & \multicolumn{1}{c}{Test} & \multicolumn{1}{c}{Train} & \multicolumn{1}{c}{Test}  \\

\midrule

MTGP (SMSE) & 0.0084 $\pm$ 0.0047 & 0.1787 $\pm$ 0.1443 & 0.0726 $\pm$ 0.006\phantom{0} & 0.2681 $\pm$ 0.0997  \\

GP-LVA (SMSE) & 0.0362 $\pm$ 0.0024 & 0.3075 $\pm$ 0.3214 & 0.0479 $\pm$ 0.0088 & 0.1804 $\pm$ 0.0614  \\

\textbf{M-AMTGP}  (SMSE)  & 0.001\phantom{0} $\pm$ 0.0005 & 0.1663 $\pm$ 0.1370 & 0.0436 $\pm$ 0.0072  & 0.1785 $\pm$ 0.0753  \\

\textbf{AMTGP}  (SMSE)  &  0.0083 $\pm$ 0.0021  & 0.1919 $\pm$ 0.1533 & 0.054\phantom{0} $\pm$ 0.009\phantom{0} & 0.1666 $\pm$ 0.0941 \\

\midrule

MTGP (SNLP)  & -1215.1 $\pm$ 95.1 & -71.8 $\pm$ \phantom{0}22.9 & -1885.2 $\pm$ \phantom{0}55.0 & -100.7 $\pm$ 41.3 \\

GP-LVA (SNLP)  & -916.7 $\pm$ 20.0 & 83.8 $\pm$ 108.3 & -2445.9 $\pm$ 106.2 & 71.5 $\pm$ 41.6 \\

\textbf{M-AMTGP} (SNLP)  & -1788.2 $\pm$ 98.1 & 435.7 $\pm$ 536.3 & -2310.2 $\pm$ 107.6 & -112.8 $\pm$ 60.3\\

\textbf{AMTGP}  (SNLP)  & -1210.4 $\pm$ 59.5  & -43.6 $\pm$ \phantom{0}46.1 & -2031.6 $\pm$ 101.4 & -131.1 $\pm$ 63.1\\

\bottomrule
\end{tabular}
\label{table:real_data_results_extra}
\end{adjustbox}
\end{table*}

\subsection{Extra Experiment with Facial Expression Data}

Here we present an additional experiment on the facial expressions data. For this experiment, we consider two mouth coordinates (lower and upper lip) and 14 recording instances yielding 28 tasks. We learn one warping function per instance and use scenario S3 (continuous segment of data missing at random location for each task) with $10\%$ of the data removed.

\begin{figure*}[ht]
\centering
\begin{subfigure}[b]{0.16\textwidth}
     \centering
     \includegraphics[width=\textwidth]{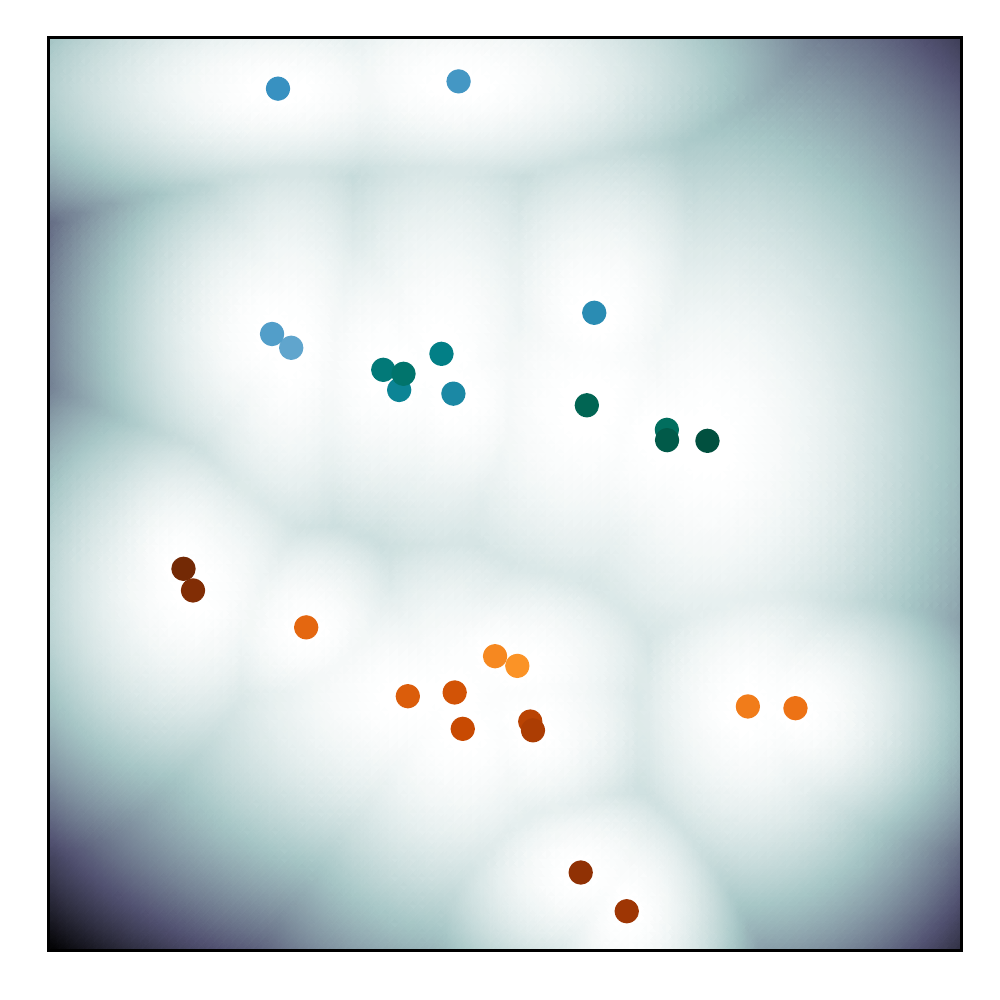}
     \caption{MTGP $\textbf{Z}$}
     \label{fig:face2_latent_mtgp}
 \end{subfigure}
 \begin{subfigure}[b]{0.4\textwidth}
     \centering
     \includegraphics[width=\textwidth]{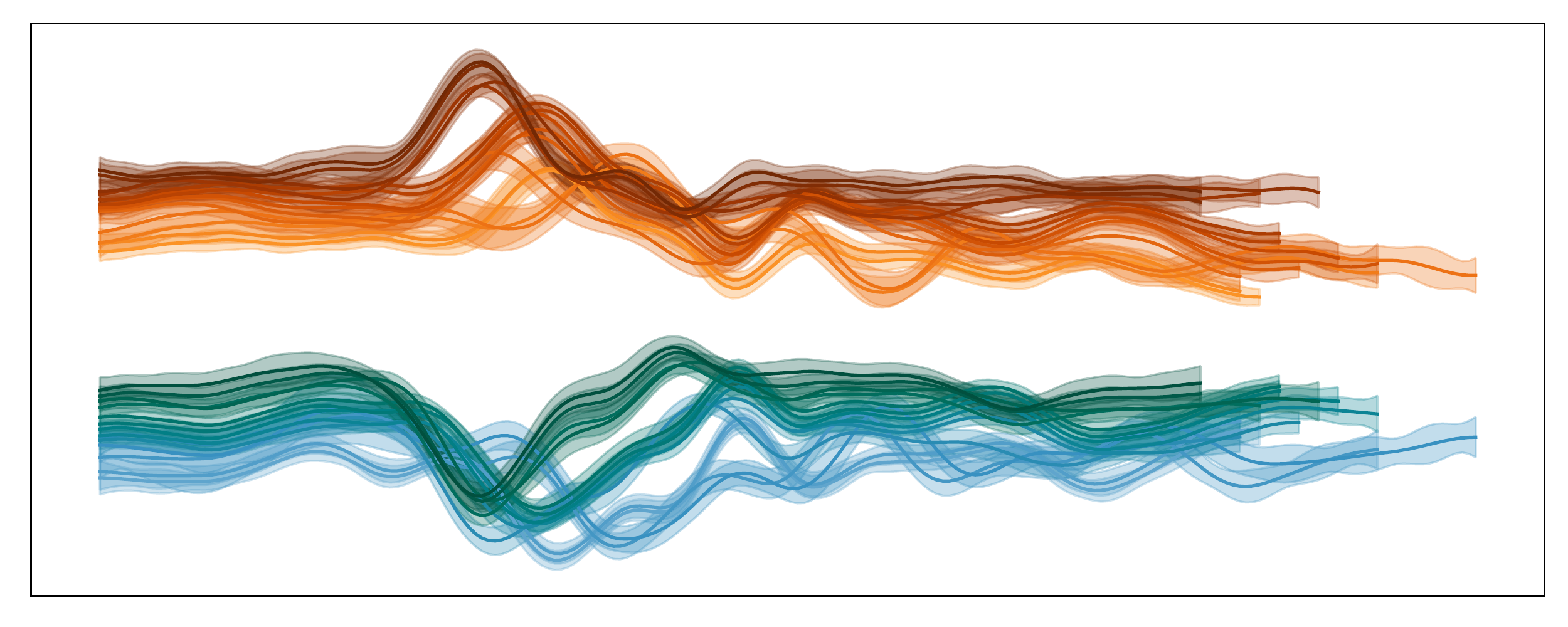}
     \caption{MTGP function posteriors (unaligned)}
     \label{fig:face2_f_post_mtgp}
 \end{subfigure}
 \begin{subfigure}[b]{0.4\textwidth}
     \centering
     \includegraphics[width=\textwidth]{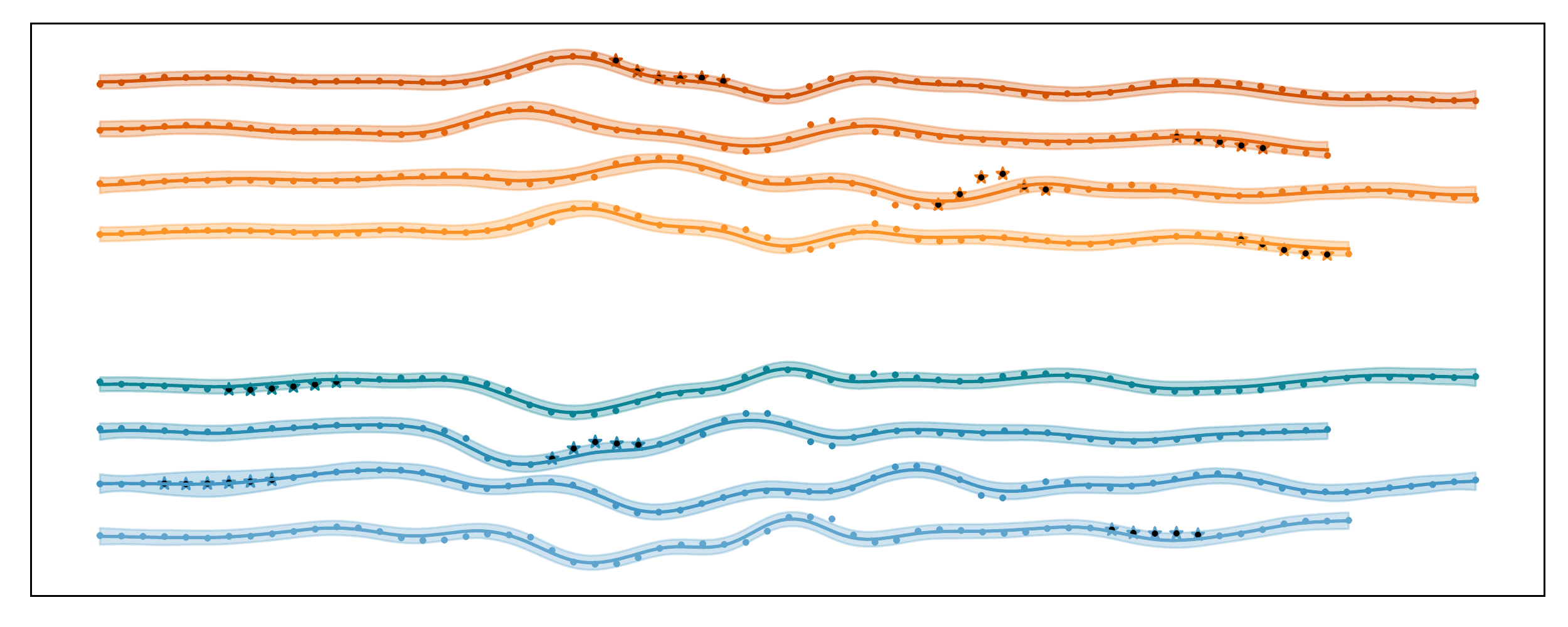}
     \caption{MTGP missing data examples}
     \label{fig:face2_data_fit_mtgp}
 \end{subfigure}\\
 \begin{subfigure}[b]{0.16\textwidth}
     \centering
     \includegraphics[width=\textwidth]{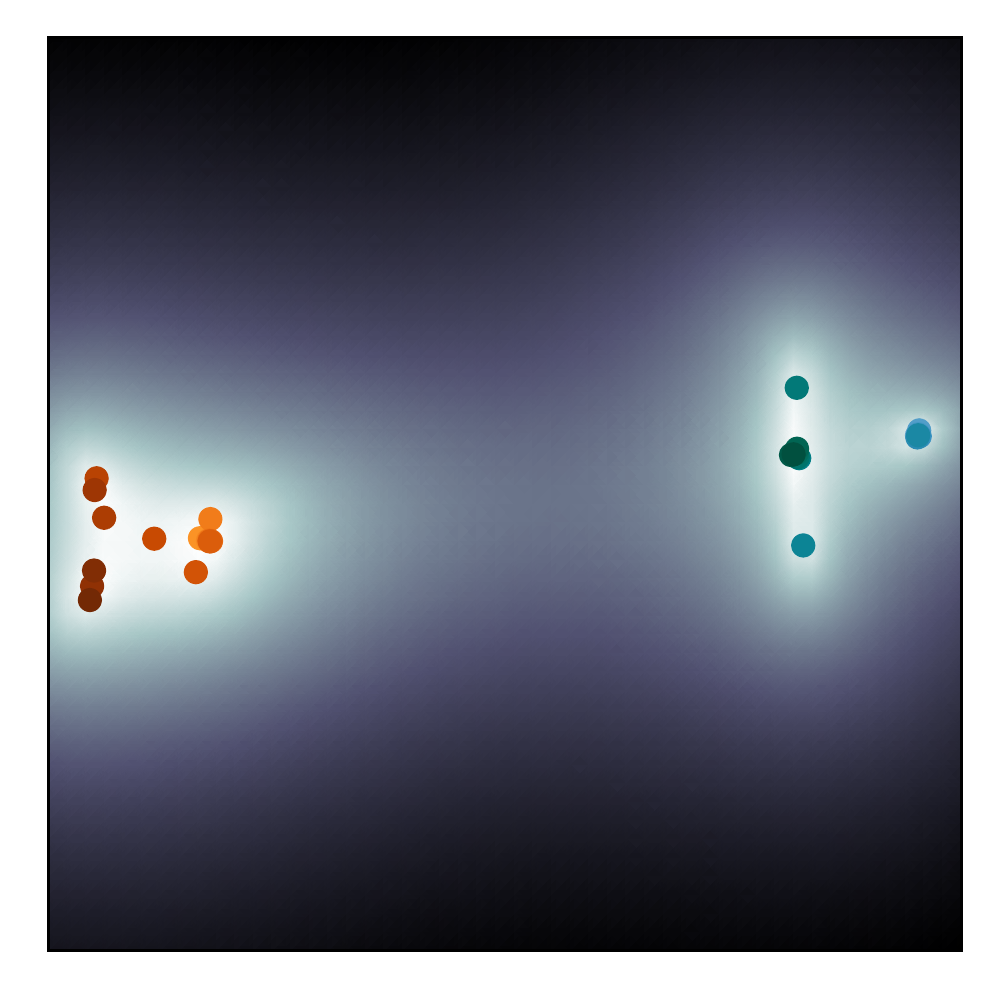}
     \caption{GP-LVA $\textbf{Z}$}
     \label{fig:face2_latent_gplva}
 \end{subfigure}
 \begin{subfigure}[b]{0.4\textwidth}
     \centering
     \includegraphics[width=\textwidth]{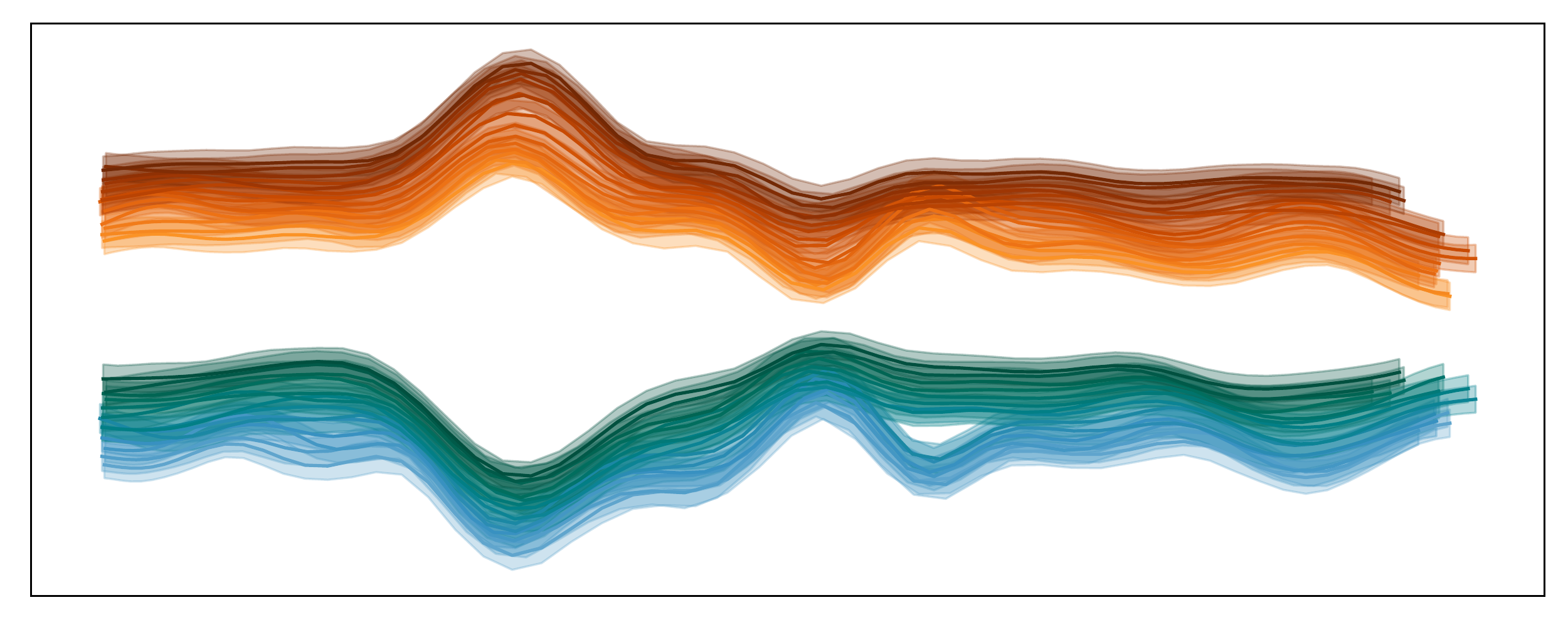}
     \caption{GP-LVA function posteriors (aligned)}
     \label{fig:face2_f_post_gplva}
 \end{subfigure}
 \begin{subfigure}[b]{0.4\textwidth}
     \centering
     \includegraphics[width=\textwidth]{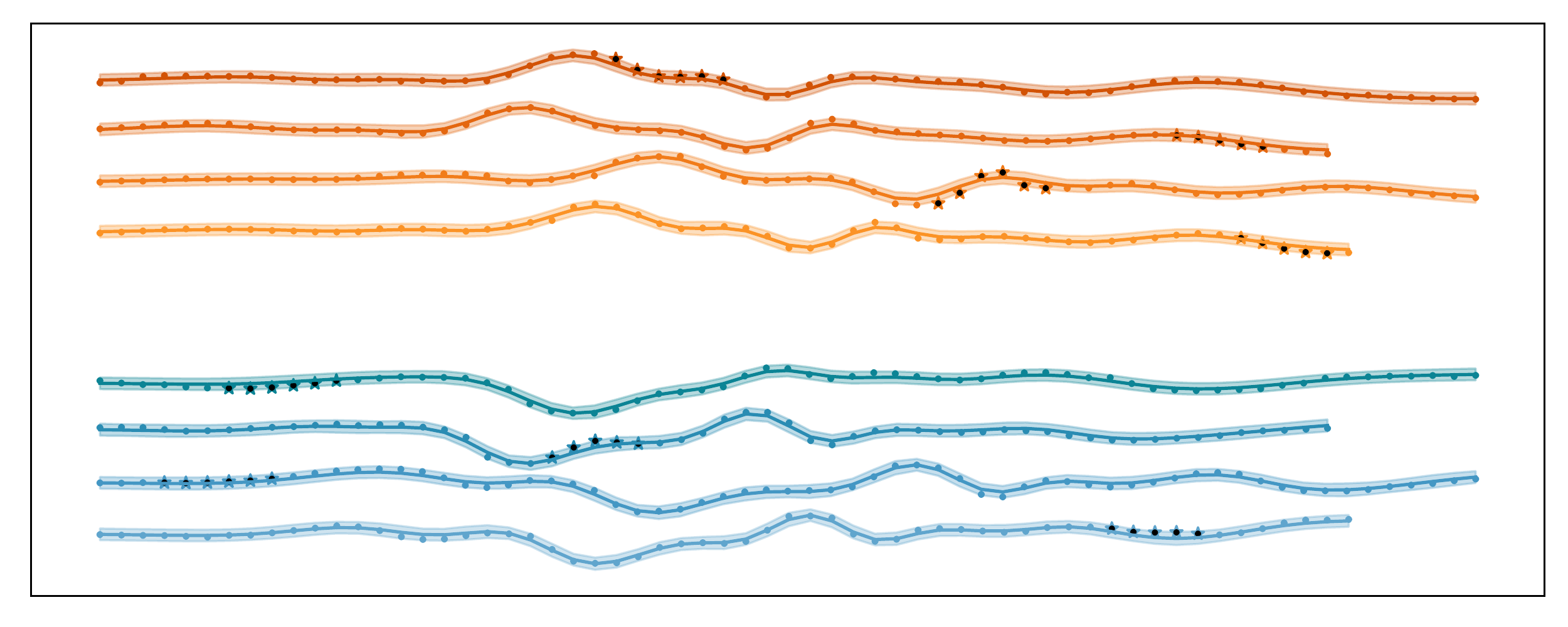}
     \caption{GP-LVA missing data examples}
     \label{fig:face2_data_fit_gplva}
 \end{subfigure}\\
 \begin{subfigure}[b]{0.16\textwidth}
     \centering
     \includegraphics[width=\textwidth]{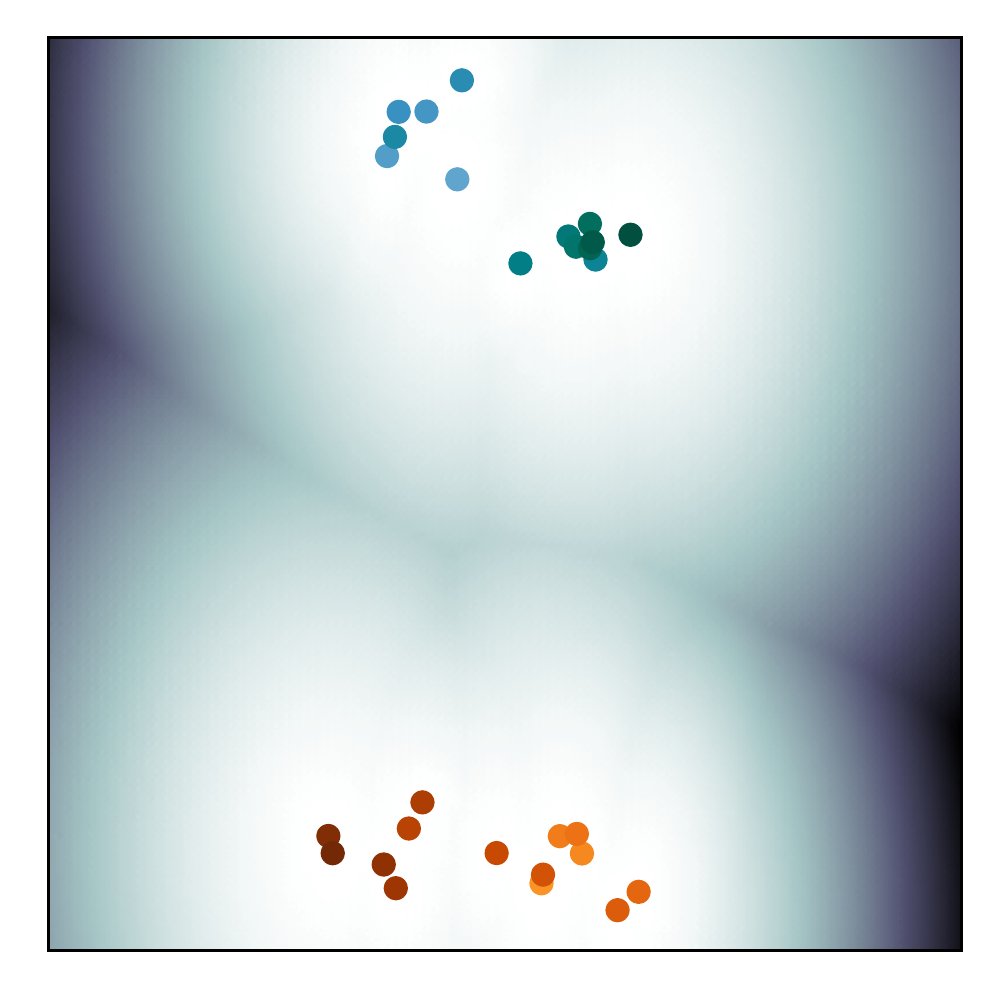}
     \caption{M-AMTGP $\textbf{Z}$}
     \label{fig:face2_latent_mamtgp}
 \end{subfigure}
 \begin{subfigure}[b]{0.4\textwidth}
     \centering
     \includegraphics[width=\textwidth]{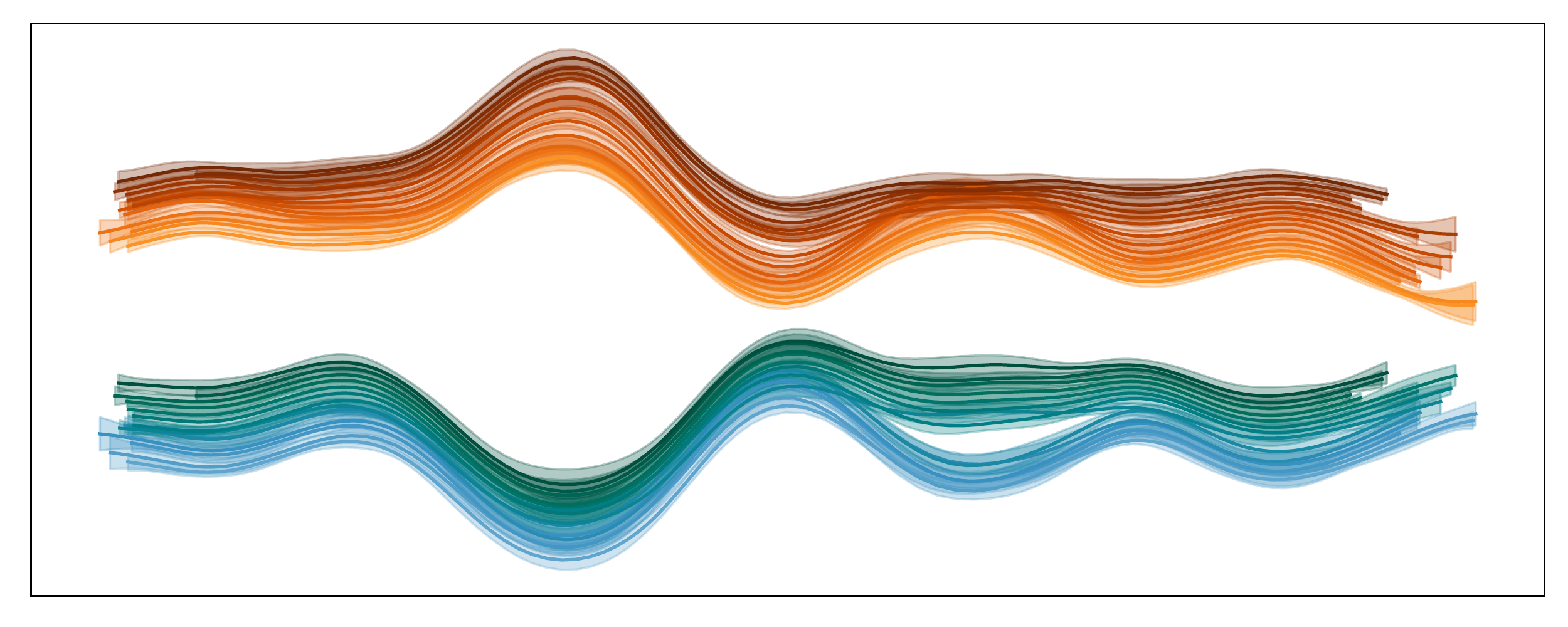}
     \caption{M-AMTGP function posteriors (aligned)}
     \label{fig:face2_f_post_mamtgp}
 \end{subfigure}
 \begin{subfigure}[b]{0.4\textwidth}
     \centering
     \includegraphics[width=\textwidth]{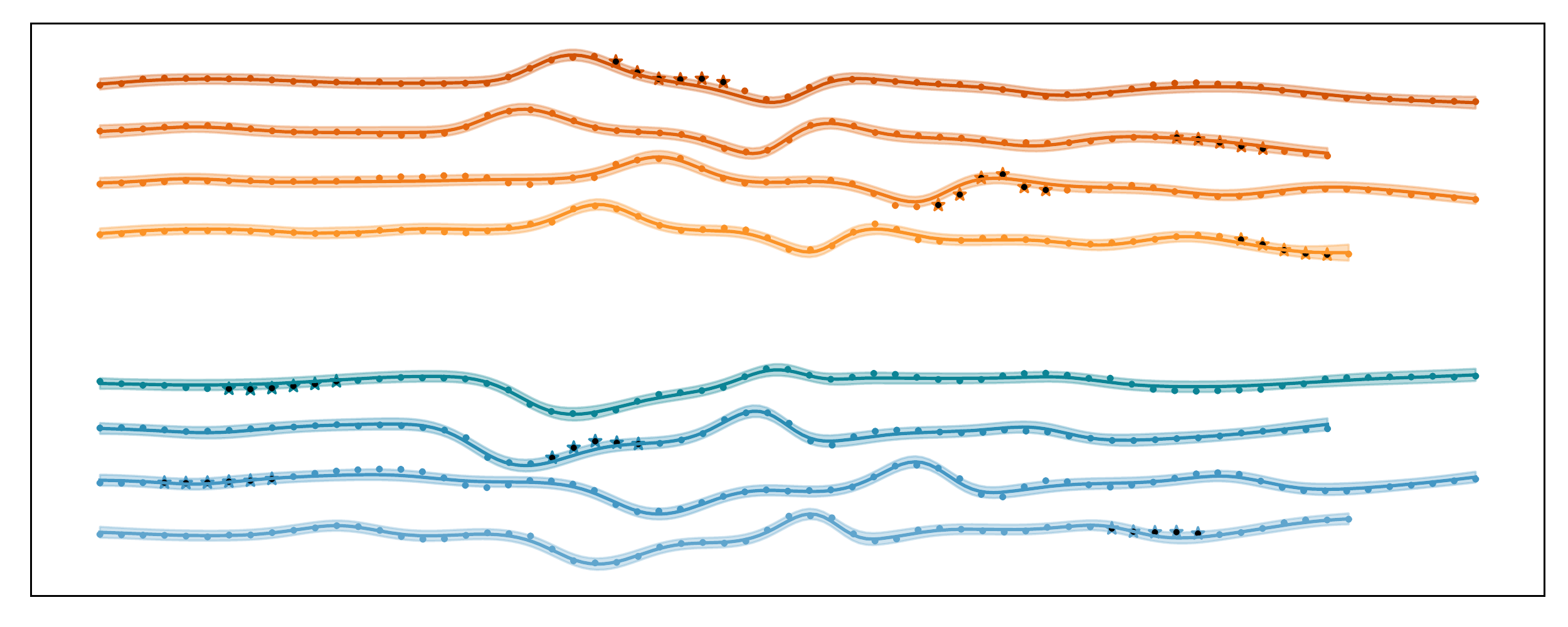}
     \caption{M-AMTGP missing data examples}
     \label{fig:face2_data_fit_mamtgp}
 \end{subfigure}\\
  \begin{subfigure}[b]{0.16\textwidth}
     \centering
     \includegraphics[width=\textwidth]{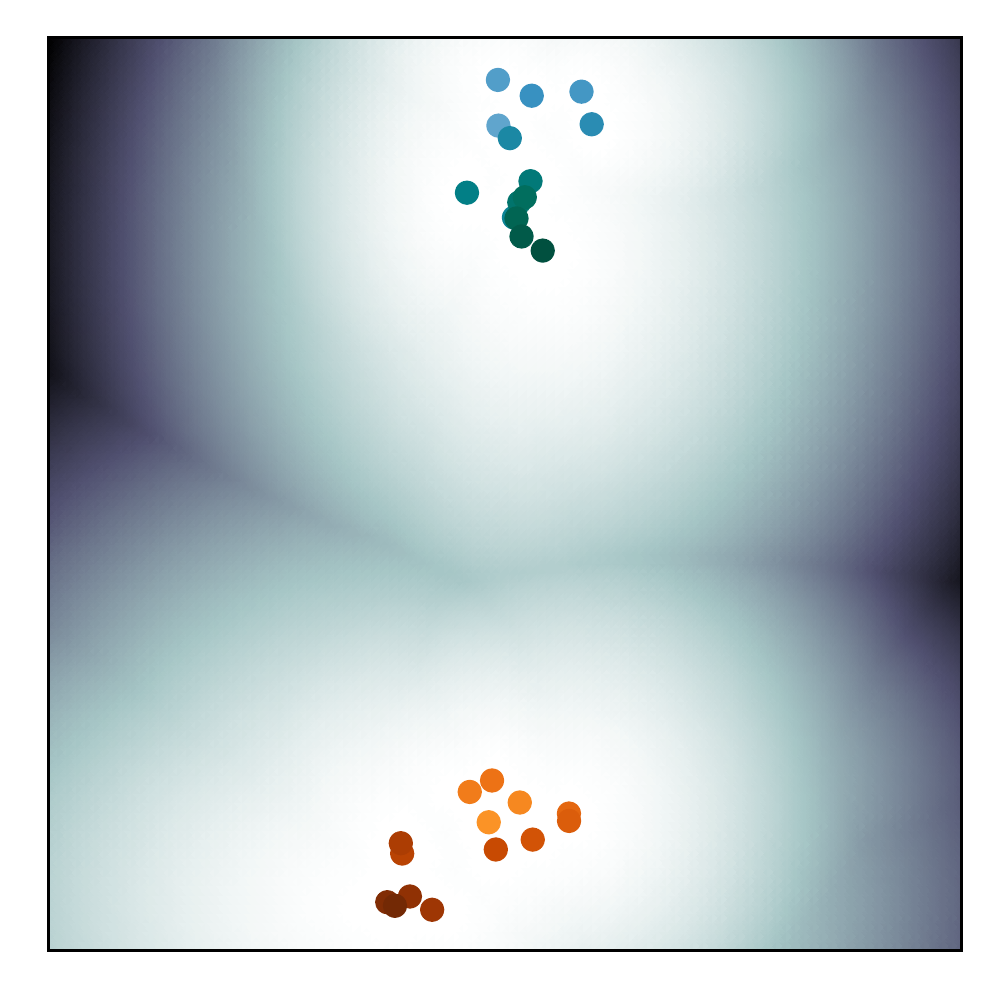}
     \caption{AMTGP $\textbf{Z}$}
     \label{fig:face2_latent_amtgp}
 \end{subfigure}
 \begin{subfigure}[b]{0.4\textwidth}
     \centering
     \includegraphics[width=\textwidth]{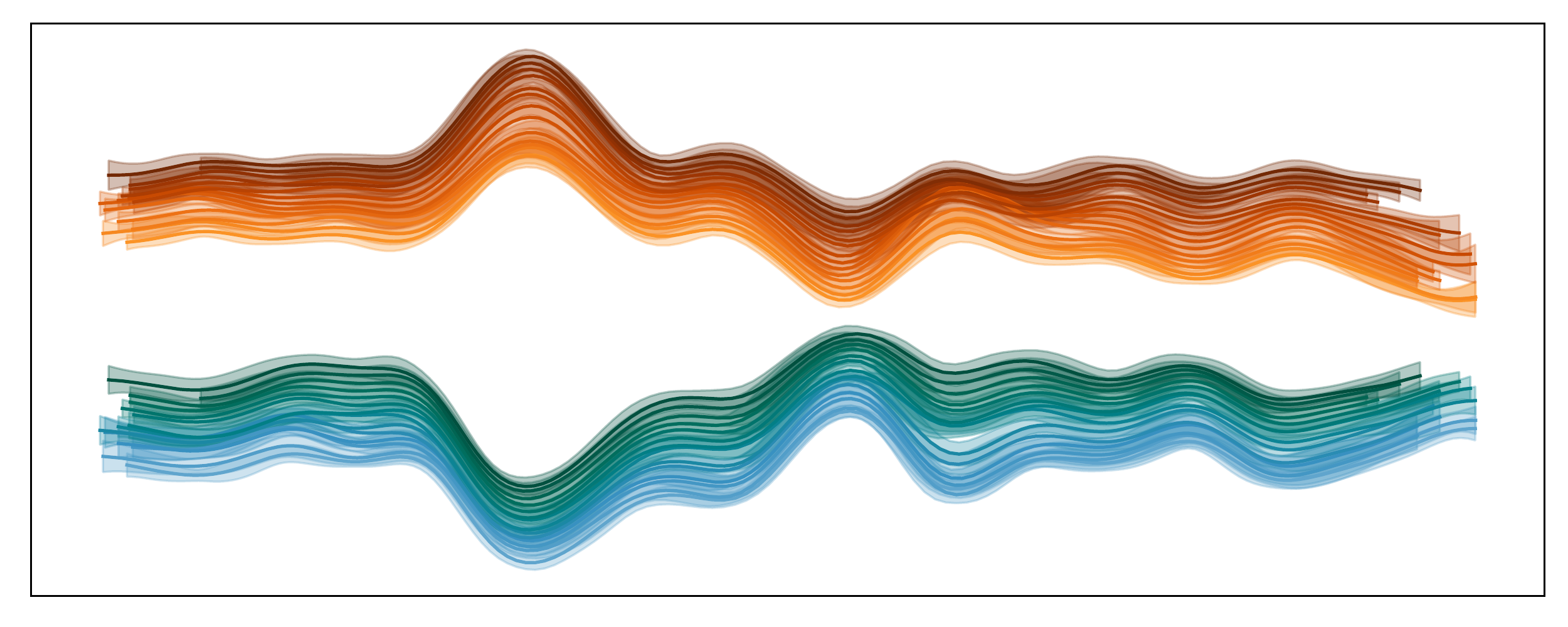}
     \caption{AMTGP function posteriors (aligned)}
     \label{fig:face2_f_post_amtgp}
 \end{subfigure}
 \begin{subfigure}[b]{0.4\textwidth}
     \centering
     \includegraphics[width=\textwidth]{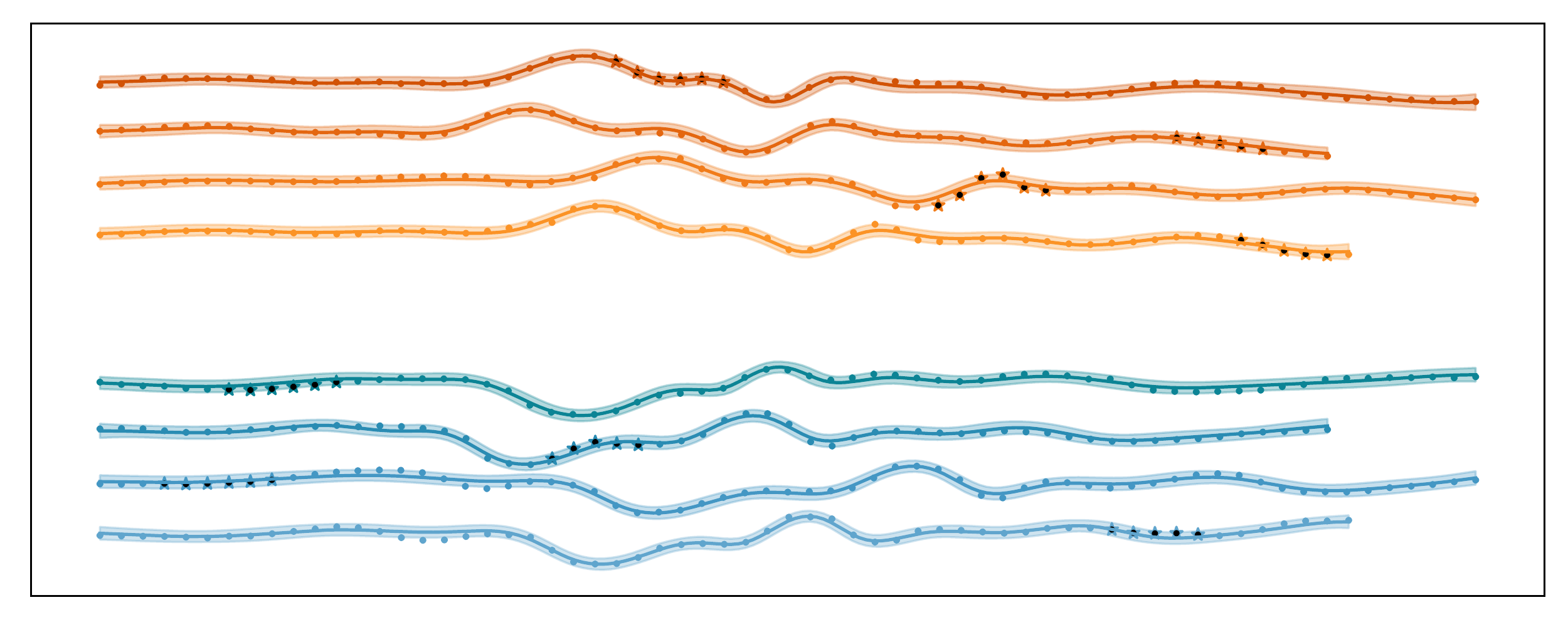}
     \caption{AMTGP missing data examples}
     \label{fig:face2_data_fit_amtgp}
 \end{subfigure}
\caption{Experiment 2 on the facial landmark data. Top row shows MTGP, second row is GP-LVA, third row is M-AMTGP (AMTGP with MAP estimate of warps), bottom row is AMTGP. (a), (d), (g) and (j) show the latent space, (b), (e), (h) and (k) show the posterior over the aligned functions. Data fit is illustrated in (c), (f), (i) and (l) for a few tasks. $10\%$ of the data is missing and is shown in black. Offset and colour coding separates upper and lower lip coordinates.}
\label{fig:face_exp2}
\end{figure*}

The results of the experiment are presented in Table \ref{table:real_data_results_extra} and Figure \ref{fig:face_exp2}. We can see, that unlike MTGP, both AMTGP and M-AMTGP are able to align and group lower and upper lip coordinated. This is reflected in better predictive performance (table \ref{table:real_data_results_extra}) of the aligned models, with fully Bayesian AMTGP being the top model.

\subsection{Sequence Alignment}
Our model can be used not only for misaligned multi-task learning, but also for sequence alignment. In this aspect, our model, Aligned-MTGP can be seen as a more probabilistically solid formulation and generalization of the GP-LVA model~\citep{kazlauskaite2019gaussian}. Here we compare the quality of alignments between GP-LVA, M-AMTGP and AMTGP on four synthetic datasets with known warps from~\cite{kazlauskaite2019gaussian}. Since GP-LVA and M-AMTGP only give point estimate of the warps, we use posterior warp means in AMTGP for comparison. As the task of alignment is underdetermined, to compare the warps we consider relative warps withing each group (one true underlying function).
In Table~\ref{table:warps}, we report the MSE between the true warps and the estimated warps with statistics over different reference warp choices. The the identity warps (\emph{i.e.}~unaligned) are shown as a baseline.  We can see that our method uncovers the true warps on par with GP-LVA while also providing a rigorous formulation of the model and the bound for the inference.

\begin{table}[h!]
\centering
\caption{Warp Recovery on Synthetic Data.}
\begin{adjustbox}{max width=.8\textwidth}
\begin{tabular}{rrrrr}
\toprule
& Identity Warps & GP-LVA \phantom{(O)}& \textbf{M-AMTGP} (Ours)& \textbf{AMTGP} (Ours)\\

\midrule

1 & 0.061 $\pm$ 0.09 & 0.0042 $\pm$ 0.0070 & 0.0017 $\pm$ 0.0026 & 0.0050
$\pm$ 0.0064\\

2 & 0.061 $\pm$ 0.09 & 0.0025 $\pm$ 0.0036 & 0.0024
$\pm$ 0.0036 & 0.0046
$\pm$ 0.0081\\

3 & 0.035 $\pm$ 0.03 & 0.0076 $\pm$ 0.0010 & 0.0055
$\pm$ 0.0088 & 0.0052
$\pm$ 0.0086\\

4 & 0.080 $\pm$ 0.09 & 0.0006 $\pm$ 0.0008 & 0.0004 $\pm$ 0.0004 & 0.0006
$\pm$ 0.0011\\

\bottomrule
\end{tabular}
\label{table:warps}
\end{adjustbox}
\end{table}

The synthetic datasets for testing warps have 10 sequences each, which are generated as follows:
\begin{enumerate}
    \item $\sinc(\pi x)$; $0.6 x^3$ (5/5)
    \item $\sin(\pi x)$; $0.6 x^3$ (5/5)
    \item $\sin(3 x)$; $0.6 x^3$ (5/5)
    \item $\sin(6 x)$ (10)
\end{enumerate}

\begin{figure}[ht]
\centering
\includegraphics[width=0.45\linewidth]{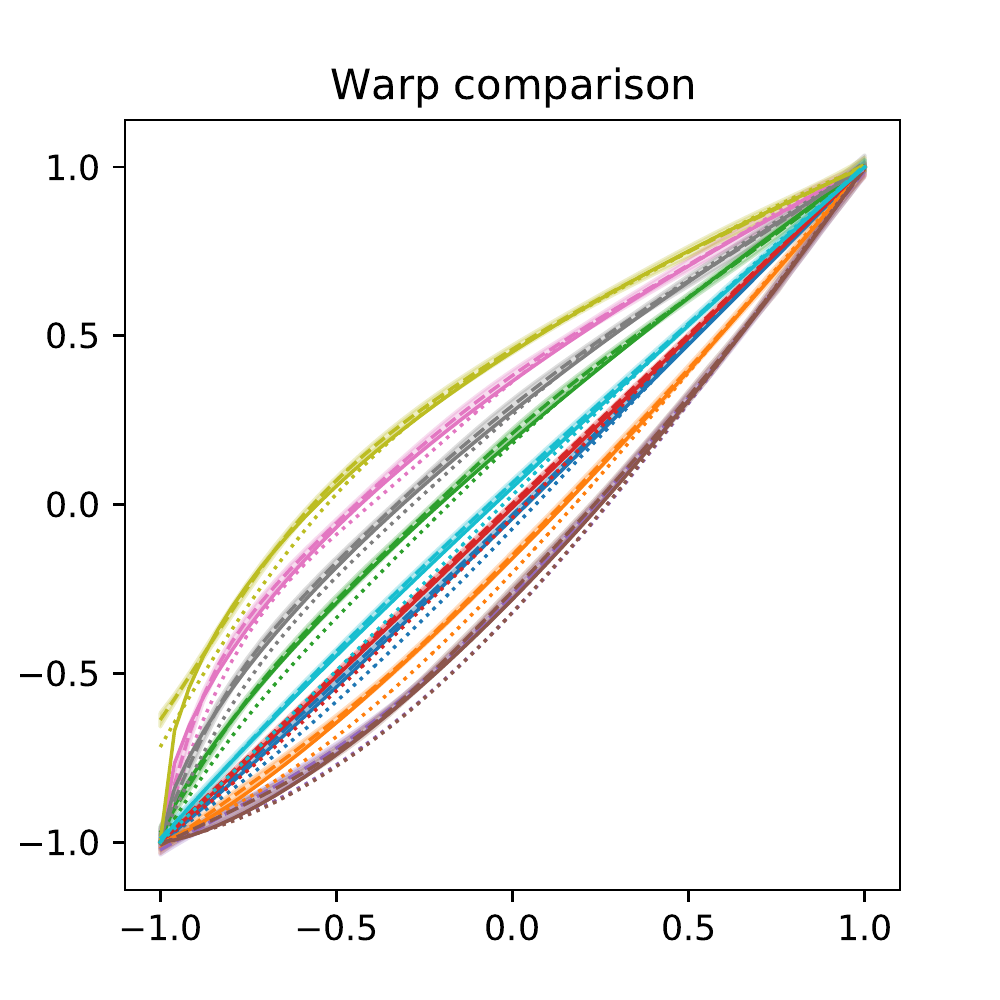}
\caption{Example of warp comparison on synthetic data (dataset 4). Solid lines show the true warps, dotted lines are GP-LVA estimation, dashed lines are AMTGP (our model) shown with 2 standard deviations.}
\label{fig:warps}
\end{figure}

As an example, the comparison of the warp estimation for the case 4 (only one underlying function) is shown in Fig. \ref{fig:warps}.

\section{Model Parameters and Implementation}

\subsection{Model and Training Parameters}

For all experiments, the number of inducing points for each model is chosen based on the saturation of ELBO for the full dataset (see Table \ref{table:incucing_points}). This ensures that the performance is not affected by the quality of sparse approximation. As a result, the aligned models (AMTGP and M-AMTGP) require fewer inducing points.

For the inference over monotonic warps in AMTGP model we use 10 inducing points on a fixed grid for each warp function.

The details about the size of the dataset used in each experiment is provided in Table \ref{table:dataset_size}.

Modeling choices  are summarized in Table \ref{table:model_parameters}.

For optimization we use a combination of Adam and natural gradients. We use natural gradients for $q(h)$ with $\gamma_h=0.5$ for all models and experiments. While it is possible to use natural gradients for parameters of $q(w)$ (underlying GPs in warps), we find that $\gamma_w$ in this case has to be very low and does not improve convergence compared to Adam for those parameters. In synthetic data experiments, we use natural gradient for the warps with $\gamma_w=0.05$, in real data experiments we use Adam.

The number of the training iterations for each experiment was determined based on the loss convergence and is reported in Table \ref{table:model_parameters}.

For all experiments, we train all models on 10 random data amputations and report mean and standard deviation of SMSE and SNLP.

\begin{table}[h]
\centering
\caption{Number of Inducing Points.}
\begin{adjustbox}{max width=1.\textwidth}
\begin{tabular}{rrrrr}
\toprule
 & \textbf{MTGP} & \textbf{M-AMTGP} & \textbf{AMTGP} & \textbf{GP-LVA} (per task)\\
\midrule
Synthetic data  & 250 & 50 & 50 & 25\\
Facial Expressions  & 200 & 200 & 100 & 20\\
Heartbeat Sounds & 200 & 100 & 100 & -\\
Facial Expressions 2 & 250 & 50 & 50 & 20\\
Respiratory Motion Traces & 100 & 100 & 100 & 10\\
\bottomrule
\end{tabular}
\label{table:incucing_points}
\end{adjustbox}
\end{table}

\begin{table}[h]
\centering
\caption{Dataset size.}
\begin{adjustbox}{max width=1.\textwidth}
\begin{tabular}{rrrr}
\toprule
 & \textbf{\# tasks} & \textbf{Sequence length} & \textbf{Total \# of data points}\\
\midrule
Synthetic data  & 10  & 100 & 1000\\
Facial Expressions  & 20  & 65\phantom{0}-\phantom{00}70 & 1350\\
Heartbeat Sounds & 10  & 120\phantom{0}-\phantom{0}150& 1368\\
Facial Expressions 2 & 28 & 56\phantom{0}-\phantom{00}70 & 1704\\
Respiratory Motion Traces & 6 & 100 & 600\\
\bottomrule
\end{tabular}
\label{table:dataset_size}
\end{adjustbox}
\end{table}

\begin{table}[h]
\centering
\caption{Model parameters and training details.}
\begin{adjustbox}{max width=1.\textwidth}
\begin{tabular}{rrrrr}
\toprule
 & \textbf{Temporal kernel} & \textbf{\# warp functions} & \textbf{Warp prior in M-AMTGP} & \textbf{Training iterations}\\
\midrule
Synthetic data  & SE & 10 & SE(0.1, 0.1) & 2000 \\
Facial Expressions  & Matern5/2 & 2 & SE(0.1, 0.1) & 2000 \\
Heartbeat Sounds & Cosine + SE & 10 & SE(0.1, 1.) & 3000\\
Facial Expressions 2 & Matern5/2 & 14 & SE(0.1, 0.1) & 2000\\
Respiratory Motion Traces & Matern5/2 & 6 & SE(0.1, 0.01) & 3000\\
\bottomrule
\end{tabular}
\label{table:model_parameters}
\end{adjustbox}
\end{table}

\begin{table}[h]
\centering
\caption{Approximate computational time, in minutes}
\begin{adjustbox}{max width=1.\textwidth}
\begin{tabular}{rrrrr}
\toprule
\textbf{GP-LVA} & \textbf{MTGP} & \textbf{M-AMTGP} & \textbf{AMTGP}\\
\midrule
Synthetic data & 1 & 5 & 5 & 60  \\
Facial Expressions & 1 & 5 & 5 & 60  \\
Heartbeat Sounds & 1 & 5 & 5 & 60\\
Facial Expressions & 1 & 5 & 5 & 60 \\
Respiratory Motion Traces & 1 & 5 & 5 & 60\\
\bottomrule
\end{tabular}
\label{table:compute_time}
\end{adjustbox}
\end{table}

\begin{table}[h]
\centering
\caption{Analysis of computational time of AMTGP (average time per iteration, sec.)}
\begin{adjustbox}{max width=1.\textwidth}
\begin{tabular}{rrrrr}
\toprule

 \# samples & \# Fourier Features & Pathwise GP sampling & Pathwise monotonic GP sampling & Backprop \\
 \midrule

1   & 1024 & 0.00102 & 0.00606 & 0.03842 \\
1   & 256  & 0.00077 & 0.00442 & 0.02084 \\
10  & 1024 & 0.00120 & 0.05216 & 0.38295 \\
10  & 256  & 0.00080 & 0.01284 & 0.08918 \\
100 & 1024 & 0.00190 & 0.68137 & 3.87087 \\
100 & 256  & 0.00104 & 0.15722 & 0.93967 \\
\bottomrule
\end{tabular}
\label{table:amtgp_compute_time}
\end{adjustbox}
\end{table}

\subsection{Implementation Details}
The model is implemented in Tensorflow. We make use of the \texttt{GPflow} framework~\citep{GPflow2017} and the \texttt{GPflowSampling} path sampling toolkit~\citep{wilson2020pathsampling}.

The code is provided as part of the supplementary materials and will be made publicly available after the review process.

\subsection{Computational time}
Approximate computational time for each experiment in provided in Table \ref{table:compute_time}. The estimate is provided for a MacBook Pro with 2,7 GHz Quad-Core Intel Core i7 and 16 GB memory.

Higher computational time of AMTGP comes from the ODE solver. While extra evaluations of the sampled functions within the solver add to the complexity, most of the computational overhead comes from differentiating through the solver. We illustrate this in Table \ref{table:amtgp_compute_time}.

\end{document}